\documentclass[twoside,11pt]{article}

\usepackage{blindtext}

\usepackage[preprint]{jmlr2e}

\usepackage{amsmath,amsfonts,dsfont,lmodern,mathrsfs}
\usepackage{comment}
\usepackage{xfrac}
\usepackage[dvipsnames]{xcolor}
\usepackage{tikz}
\hypersetup{colorlinks,allcolors=blue}
\usepackage{doi}
\usepackage[shortlabels]{enumitem}
\setlist[itemize,1]{label=--}

\renewenvironment{proof}[1][Proof]{\par\noindent{\bf #1.\ }}{\hfill\BlackBox\\[2mm]}

\newcommand{\rmd}{\mathrm{d}}
\newcommand{\1}{\mathds{1}}
\renewcommand{\P}{\mathrm{P}}

\newcommand{\E}{\mathbb{E}}
\newcommand{\norm}[2][]{#1\lVert #2 #1\rVert}
\newcommand{\abs}[2][]{#1\lvert #2 #1\rvert}

\renewcommand{\epsilon}{\varepsilon}
\renewcommand{\rho}{\varrho}
\renewcommand{\phi}{\varphi}
\renewcommand{\emptyset}{\varnothing}
\renewcommand{\leq}{\leqslant}
\renewcommand{\le}{\leqslant}
\renewcommand{\geq}{\geqslant}
\renewcommand{\ge}{\geqslant}

\usepackage{float}

\usepackage{rotating} 

\usepackage{tikz}
\usepackage{pgfplots}
\pgfplotsset{compat=1.18} 
\usepackage{graphicx}     
\usepackage{booktabs}
\usepackage{multirow}   
\usepackage{siunitx}    
\usepackage{caption}    

\usepgfplotslibrary{groupplots}
\usepgfplotslibrary{fillbetween}
\usepgfplotslibrary{statistics}
\usepackage{pdfpages}
\usetikzlibrary{positioning,calc,arrows,patterns} 
\usepackage{subcaption} 

\ShortHeadings{An RKHS Perspective on Tree Ensembles}{M.~Dagdoug, C.~Dombry and J.-J.~Duchamps} 
\firstpageno{1}

\begin{document}

\title{An RKHS Perspective on Tree Ensembles}
\author{\name Mehdi Dagdoug \email mehdi.dagdoug@mcgill.ca \\
        \addr McGill University, \\ Department of Mathematics and Statistics\\
        Montréal, Canada
        \AND
        \name Clément Dombry \email clement.dombry@umlp.fr \\
       \addr Universit{\'e} Marie et Louis Pasteur,\\ CNRS, LmB (UMR 6623),\\ F-25000 Besan\c{c}on, France
       \AND
       \name Jean-Jil Duchamps \email jean-jil.duchamps@umlp.fr \\
        \addr Universit{\'e} Marie et Louis Pasteur,\\ CNRS, LmB (UMR 6623),\\ F-25000 Besan\c{c}on, France
}
\editor{My editor}

\maketitle

\begin{abstract}
Random Forests and Gradient Boosting are among the most effective algorithms for supervised learning on tabular data. Both belong to the class of tree-based ensemble methods, where predictions are obtained by aggregating many randomized regression trees. In this paper, we develop a theoretical framework for analyzing such methods through \emph{Reproducing Kernel Hilbert Spaces} (RKHSs) constructed on tree ensembles—more precisely, on the random partitions generated by randomized regression trees. We establish fundamental analytical properties of the resulting \emph{Random Forest kernel}, including boundedness, continuity, and universality, and show that a Random Forest predictor can be characterized as the unique minimizer of a penalized empirical risk functional in this RKHS, providing a variational interpretation of ensemble learning. We further extend this perspective to the continuous-time formulation of Gradient Boosting introduced by \citet{DD24,DD24a}, and demonstrate that it corresponds to a gradient flow on a Hilbert manifold induced by the Random Forest RKHS. A key feature of this framework is that both the kernel and the RKHS geometry are data-dependent, offering a theoretical explanation for the strong empirical performance of tree-based ensembles. Finally, we illustrate the practical potential of this approach by introducing a kernel principal component analysis built on the Random Forest kernel, which enhances the interpretability of ensemble models, as well as $\mathrm{GVI}$, a new geometric variable importance criterion.\\
\end{abstract}


\tableofcontents

\section{Introduction}

Tree-based ensemble methods, notably \emph{Random Forests} \citep{breiman2001random} and \emph{Gradient Boosting} \citep{friedman2001greedy}, have become fundamental tools for supervised learning on tabular data. Their recursive partitioning mechanism naturally mitigates the curse of dimensionality by adaptively selecting informative covariates while down-weighting irrelevant or noisy ones \citep{Biau2016}. Modern implementations such as \texttt{ranger} \citep{Wright2017} and \texttt{XGBoost} \citep{CG16} have further enhanced their appeal through computational efficiency and scalability, allowing the training of large-scale models with relative ease. Empirically, Random Forests often perform well with default parameters, while XGBoost typically achieves higher accuracy after moderate tuning \citep{Bentejac2021}. Together, these methods form the backbone of modern predictive modeling for tabular data.

Beyond their empirical success, the theoretical understanding of tree-based ensembles has advanced considerably. For Random Forests, early analyses established asymptotic consistency in simplified settings \citep{Breiman2004, Biau2012}, later extended to more general frameworks with explicit rates of convergence \citep{Scornet2015}. These works highlight how bootstrap aggregation and random feature selection act as implicit regularizers, reducing variance without excessive bias \citep{mentch2020randomization}. Recent developments have also explored their asymptotic properties in high-dimensional regimes \citep{chi2022asymptotic, klusowski2024large}. 

The theoretical study of Gradient Boosting, in contrast, builds on its interpretation as a functional gradient descent in function space \citep{friedman2001greedy}, thereby linking it to additive modeling  \citep{BY03}. A substantial body of work in the statistical literature focuses on consistency, that is, the ability of the procedure to achieve the optimal Bayes error rate as the sample size tends to infinity. Such consistency results have been established for AdaBoost \citep{J04}, for more general boosting procedures \citep{LV04, BLV04, ZY05}, and for infinite-population versions of boosting \citep{B04}. These analyses typically rely on the fact that the Bayes predictor can be approximated by linear combinations of simple functions (such as decision trees), together with a suitable measure of the complexity of this function class (such as VC-dimension). 

Although their analytical perspectives differ—Random Forests as stochastic averages of weak learners and Gradient Boosting as sequential risk minimization—both approaches now benefit from increasing mathematical rigor, gradually narrowing the gap between empirical performance and theoretical understanding.

\subsection{Random Forest weights and kernels}

Consider a regression setting with training data $(X_i, Y_i)_{1 \leq i \leq n}$ in $\mathcal{X} \times \mathbb{R}$. 
The Random Forest predictor can be written as
\[
\hat{F}_n(x) = \sum_{i=1}^n W_{ni}(x) Y_i,
\]
where the weights $(W_{ni}(x))_{1 \leq i \leq n}$ encode the forest structure. 
For a forest of $M$ trees without prediction-level resampling,\footnote{All observations are used for prediction, although resampling may occur during tree construction.} these weights take the form
\[
W_{ni}^M(x) = \frac{1}{M} \sum_{m=1}^M \frac{\1_{\{X_i \in L_m(x)\}}}{|L_m(x)|}, \quad 1 \leq i \leq n,
\]
where $L_m(x)$ denotes the leaf of the $m$-th tree containing $x$, and $|L_m(x)|$ its number of training points.
This representation naturally defines a \emph{Random Forest kernel},
\begin{equation}
\label{eq:RFkernel}
k_n^M(z, z') = \frac{n}{M} \sum_{m=1}^M \sum_{l=1}^{N_m} 
\frac{\1_{\{z \in A_{m,l}\}} \, \1_{\{z' \in A_{m,l}\}}}{|A_{m,l}|},
\qquad z, z' \in \mathcal{X},
\end{equation}
where $A_{m,1}, \dots, A_{m,N_m}$ enumerate the leaves of the $m$-th tree. 
This function is symmetric, positive semi-definite, and satisfies
\[
W_{ni}^M(X_j) = \frac{1}{n}k_n^M(X_i, X_j), \qquad 1 \leq i,j \leq n.
\]
Hence, the Random Forest weight matrix coincides (up to scaling) with the Gram matrix of the kernel evaluated at the training points. 
Although this kernel depends on the training data—since it arises from the fitted Random Forest—it provides a natural bridge between ensemble learning and kernel methods.

\subsection{Contributions}

While earlier works have hinted at connections between Random Forests and kernels \citep{DaviesGhahramani2014, Scornet2016random}, a detailed mathematical analysis of the Random Forest kernel and its associated reproducing kernel Hilbert space (RKHS) is still missing. 
This paper aims to fill this gap. Our main contributions are as follows:

\begin{itemize}
    \item \textbf{Definition and characterization of the Random Forest kernel.}  
We rigorously define the kernel associated with a Random Forest through its random partition structure and establish its fundamental analytical properties—measurability, boundedness, and continuity—under mild assumptions on the partition distribution. We also show that the corresponding RKHS inherits regularity and approximation properties that make it a natural functional space for analyzing ensemble tree predictors. In addition, we analyze the characteristic property of the Random Forest kernel and its relationship to universality, highlighting conditions under which the induced RKHS is dense in the space of continuous functions.

    \item \textbf{RKHS representation of the Random Forest predictor.}  
    We prove that the infinite Random Forest predictor itself belongs to the induced RKHS, and provide an explicit representation of this predictor as a kernel integral. This representation allows us to interpret the forest as a kernel mean embedding of the signed measure $A \mapsto \E[Y\1_A(X)]$, linking tree ensembles to the broader framework of kernel methods.

    \item \textbf{Variational property and gradient structure.} 
    We establish that the infinite Random Forest uniquely minimizes a penalized empirical risk functional in its RKHS, providing a variational characterization of the predictor. This viewpoint reveals that Random Forests implicitly regularize the fitted function through the geometry of their RKHS. Moreover, we show that the infinite Random Forest can be interpreted as the \emph{gradient of the empirical risk}—under the mean squared error—in this space.

    \item \textbf{Connection with gradient boosting and functional flows.}  
    Building on this RKHS framework, we show that the continuous-time limit of gradient boosting—known as \emph{infinitesimal gradient boosting} \citep{DD24,DD24a}—coincides with a \emph{gradient flow} on an infinite-dimensional Riemannian manifold (i.e.\ a Hilbert manifold) naturally induced by the Random Forest RKHS. This result bridges ensemble methods and infinite-dimensional optimization.

    \item \textbf{Empirical illustration via kernel methods.}  
    Finally, we demonstrate the practical potential of the Random Forest kernel by applying kernel PCA to a variety of benchmark classification datasets. The resulting representations exhibit strong class separation, confirming that the adaptive geometry encoded by the Random Forest kernel provides informative and discriminative embeddings of the data. A Geometric Variance Importance criterion, $\mathrm{GVI}$, is proposed and compared empirically with other common measures, such as Mean Decrease Impurity (MDI) and Mean Decrease Accuracy (MDA).
\end{itemize}

\subsection{Related works}

The connection between tree ensembles and kernel methods was made explicit by \citet{DaviesGhahramani2014}, who introduced the notion of a \emph{random partition kernel}, and by \citet{Scornet2016random}, who linked Random Forests to kernel regression. 
However, these works considered a different kernel,
\[
k_n^M(z, z') = \frac{1}{M} \sum_{m=1}^M  \1_{\{z \in L_m(z')\}},
\]
which measures co-occurrence of points in the same leaf (or its infinite forest version). 
Our kernel instead takes the form
\[
k_n^M(z, z') = \frac{n}{M} \sum_{m=1}^M \frac{\1_{\{z \in L_m(z')\}}}{|L_m(z)|},
\]
and coincides (up to scaling) with the Random Forest prediction weights. 
The Random Forest weight matrix has also appeared in contexts such as generalized random forests \citep{Athey2019grf}, distributional forests \citep{meinshausen2006quantile, Cevid2022}, and local linear forests \citep{Friedberg2020}, where it defines adaptive probability measures at each query point. Random forest weights have also been investigated to prove the $L^2$-consistency of the associated random forests via Stone's theorem \citep{stone1977consistent} when the partitions do not depend on $Y$. This has been done, for instance, in \cite{biau2008consistency, Biau2012, scornet2016asymptotics}.

To our knowledge, no prior work has systematically analyzed the mathematical structure of the Random Forest kernel, the geometry of its RKHS, or its integration within kernel-based learning methods.

\subsection{Outline}

Section~\ref{sec:RKHS} introduces the theoretical framework and main results on Random Forest kernels and their RKHSs. 
Section~\ref{sec:IGB} develops the connection between Random Forest kernels and infinitesimal gradient boosting, showing that the latter can be viewed as a gradient flow on the induced Hilbert manifold. 
Section~\ref{sec:illustration} presents empirical illustrations through kernel PCA experiments. Sections~\ref{sec:proofs-1} and~\ref{sec:proofs-2} gather all the proofs, while some technical details are deferred to the Appendix. Notation is summarized at the end of this section.

\subsection{Notation}
 We denote by $\mathcal{P} \left(S\right)$ the set of probability measures on a measurable space $(S,\mathcal{B})$. For a measure $\P \in \mathcal{P}\left(S\right)$ and a measurable function $f : S \to \mathbb{R}$, we write $\P \left[ f \right] = \int f\,\mathrm{d}P$. For $\P \in \mathcal{P}([0,1]^p \times \mathbb{R})$, we shall, with a slight abuse of notation, write $\P(A) := \P(A \times \mathbb{R})$ to denote the probability of a measurable set $A \subseteq [0,1]^p$ with respect to the first marginal of $\P$. We may also write $\P(\rmd x, \rmd y)$ and $\P(\rmd x)$ to denote the joint distribution and its first marginal, respectively.   Given a set of points $Z_i=(X_i,Y_i)$, $1\leq i\leq n$, in $[0,1]^p\times\mathbb{R}$, we denote their empirical measure by $\P_n = n^{-1}\sum_{i=1}^n \delta_{Z_i}$, with $\delta_{Z}$ the Dirac mass at $Z$. Then $\P_n \in \mathcal{P} \left(\left[0,1\right]^p \times \mathbb{R}\right)$ and, due to the above conventions,    $$\P_n\left[f \right] = \dfrac{1}{n}\sum_{i=1}^n f\left(X_i, Y_i\right), \qquad \text{and} \qquad \P_n\left(A\right) = \dfrac{1}{n} \sum_{i=1}^n \1_A\left(X_i\right), $$ for a measurable function $f: \left[0,1\right]^p \times \mathbb{R} \to \mathbb{R}$ and a measurable set $A \subseteq \left[ 0,1\right]^p$. For an integer $p \in \mathbb{N}^*$, we define $[\![1,p]\!] = \left\{1, ..., p \right\}$.

\section{The RKHS associated with a randomized tree}\label{sec:RKHS}

\subsection{A general framework for random partition and tree ensembles}\label{sec:general-framework}

We define a general framework for random partitions of the covariate space $[0,1]^p$ into hyperrectangles, which forms the foundation of our approach. The main models for such random partitions are those induced by randomized decision trees, and our framework is therefore closely related to the random forest algorithm. We first define a random partition and the associated infinite random forest predictor in an abstract setting, the next subsection being devoted to specific models and examples  falling within this framework.

\medskip

\paragraph{Random partitions.} 
We call \emph{hyperrectangle} a subset of $[0,1]^p$ of the form
\[
A=\prod_{j=1}^p [a_j,b_j\rangle,
\]
with $0 \leq a_j < b_j \leq 1$ for $1 \leq j \leq p$, where
\[
[a_j,b_j \rangle =
\begin{cases} 
[a_j,b_j) & \text{if } b_j<1, \\  
[a_j,b_j] & \text{if } b_j=1. 
\end{cases}
\]
We use the shorthand notation $A=[a,b\rangle$ with $a=(a_j)_{1\leq j\leq p}$ and $b=(b_j)_{1\leq j\leq p}$ to denote a hyperrectangle. Let $\mathcal{A}$ denote the set of all such hyperrectangles. We define a metric $\rho$ on $\mathcal{A}$ by
\begin{equation} \label{eq:def-rho}
    \rho([a,b\rangle,[a',b'\rangle)=\max_{1\leq j\leq p}\big(|a_j-a'_j| \vee |b_j-b'_j|\big).
\end{equation}
The metric space $(\mathcal{A},\rho)$ is endowed with the Borel $\sigma$-algebra generated by its open sets.

A \emph{partition} of $[0,1]^p$ is a finite collection of hyperrectangles
\[
\pi=\{A_1,\ldots,A_K\}, \quad K\geq 1,\; A_1,\ldots,A_K\in \mathcal{A},
\]
such that $A_1,\ldots,A_K$ (the components of $\pi$) are pairwise disjoint and cover $[0,1]^p$. The space $\mathfrak{P}$ is equipped with a $\sigma$-algebra that makes the mapping $\pi \mapsto |\pi|$ measurable, where $|\pi|$ denotes the cardinality of $\pi$, as well as the mappings $\pi \mapsto |\pi \cap B|=\sum_{A\in\Pi} \1_{\{A\in B\}}$ for all Borel sets $B \subset \mathcal{A}$.
See Appendix~\ref{app:measurability-partitions} for details on measurability issues in the space of partitions.

\begin{definition}\label{def:random-partition} 
A \emph{random partition} $\Pi$ of $[0,1]^p$ is a measurable mapping $\Pi:\Omega\to \mathfrak{P}$ defined on a probability space $(\Omega,\mathcal{F},\mathbb{P})$. 
\end{definition}

A random partition is generally obtained via the construction of a decision tree, conditionally on a sample $\mathcal{D}_n=(X_1,\ldots,X_n)$, and using some auxiliary randomness -- see Section~\ref{sec:models-and-examples}. We use the notation $\mathbb{P}_\Pi$ for the probability and  $\mathbb{E}_\Pi$ for the expectation with respect to $\Pi$ conditionally on $\mathcal{D}_n$, respectively. The distribution of $\Pi$ (conditionally on $\mathcal{D}_n$) is denoted by $\mu$, so that
\begin{equation}\label{eq:def-mu}
\mathbb{E}_\Pi[ \Phi(\Pi)]=\int_{\mathfrak{P}} \Phi(\pi)\,\mu(\rmd \pi),
\end{equation}
for all integrable functions $\Phi:\mathfrak{P}\to \mathbb{R}$.
We always assume that the size of the partition is integrable, i.e.
\begin{equation}\label{eq:integrability-pi}
\mathbb{E}_\Pi[|\Pi|]=\int_{\mathfrak{P}} |\pi|\,\mu(\rmd \pi)<\infty. \tag{$A_0$}
\end{equation}
This implies that the intensity measure $\nu$ of components of $\Pi$, defined by  
\[
\nu(B)=\mathbb{E}_\Pi\Big[\sum_{A\in\Pi} \1_{\{A\in B\}}\Big], \quad B\subset\mathcal{A}\ \text{Borel},
\]
has a finite total mass equal to $\nu(\mathcal{A})=\mathbb{E}_\Pi[|\Pi|]$. Furthermore, for any integrable function $\Psi:\mathcal{A}\to \mathbb{R}$,
\begin{equation}\label{eq:def-nu}
\mathbb{E}_\Pi\Big[ \sum_{A\in \Pi} \Psi(A)\Big]=\int_{\mathcal{A}}\Psi(A)\,\nu(\rmd A).
\end{equation}
Considering $\Psi(A)=\1_A(z)$ for some fixed $z$, we deduce
\begin{equation}\label{eq:normalisation-nu}
\int_{\mathcal{A}} \1_A(z)\,\nu(\rmd A)=1 , \quad z\in[0,1]^p,
\end{equation}
since for any partition $\pi$, every point $z\in[0,1]^p$ belongs to exactly one component of $\pi$, and hence $\sum_{A\in\Pi}\1_A(z)=1$ almost surely.

\paragraph{Histogram estimation, aggregation, and random forests.}
In a regression framework, let $(X,Y)$ be a random element taking values in $[0,1]^p\times\mathbb{R}$ with distribution $\P^*$. For $n\geq 1$, we consider a sample $(X_i,Y_i)_{1\leq i\leq n}$ of independent copies of $(X,Y)$ with empirical distribution $\P_n$. In what follows, $\P$ will denote either $\P_n$ (finite sample case) or $\P^*$ (infinite sample case).  

Following \citet{nobel1996histogram}, given a partition $\pi\in\mathfrak{P}$, the \emph{histogram estimator} of the regression function of $Y$ given $X=z$ is defined by
\begin{equation}\label{eq:def-T}
T(z;\P,\pi)=\sum_{A\in \pi} \frac{\P[y\1_A(x)]}{\P(A)}\,\1_A(z), \quad z\in[0,1]^p,
\end{equation}
with the convention $0/0=0$. The estimator is constant on the components of the partition with value corresponding to the conditional expectation of $y$ given $x\in A$ with respect to $\P$. Since in practice $\pi$ is obtained from a decision tree, we also refer to $T(\cdot;\P,\pi)$ as a regression tree predictor.

We focus on ensemble models obtained as an aggregation of regression tree predictors obtained by sampling independent random partitions $\Pi_1,\Pi_2,\ldots$ from distribution $\mu$. We call this aggregation the \emph{random forest predictor}. Such ensembles of trees are known to have enhanced predictive power \citep{zhou2012ensemble}.  For $M\geq 1$, we define
\[
\bar T_M(z;\P,\Pi_1,\ldots,\Pi_M)=\frac{1}{M}\sum_{m=1}^M T(z;\P,\Pi_m).
\]
In practice, the number $M$ of ensemble members  is chosen large enough that sampling fluctuations become negligible, leading us to consider the infinite-ensemble limit
\begin{equation}\label{eq:cv-rf}
\bar T(z;\P,\mu):= \lim_{M\to\infty} \frac{1}{M}\sum_{m=1}^M T(z;\P,\Pi_m) 
= \mathbb{E}_\Pi[T(z;\P,\Pi)] \quad \text{a.s.},
\end{equation}
where the almost sure convergence (for each fixed $z$) is a straightforward consequence of the strong law of large numbers. 
 
By the preceding definitions, the limit can be rewritten as
\begin{align*}
\bar T(z;\P,\mu) &= \mathbb{E}_\Pi[T(z;\P,\Pi)]= \mathbb{E}_\Pi\Big[\sum_{A\in\Pi}\frac{\P[y\1_A(x)]}{\P(A)}\,\1_A(z) \Big] \\
&=  \int_{\mathfrak{P}} T(z;\P,\pi)\,\mu(\rmd\pi)=\int_{\mathcal{A}} \frac{\P[y\1_A(x)]}{\P(A)}\,\1_A(z)\,\nu(\rmd A).
\end{align*}
For further reference, we provide the following formal definition. Recall that the measure $\P$ on $[0,1]^p\times\mathbb{R}$ represents the data (finite sample if $\P=\P_n$, or infinite sample if $\P=\P^*$), and that the random partition $\Pi$ has distribution $\mu$ conditionnaly on $\P$.
 
\begin{definition}\label{def:random-forest} The \emph{infinite random forest}, or simply \emph{random forest},
associated with data $\P$ and random partition $\Pi$ is defined by
\begin{equation}\label{eq:def-bar-T}
\bar T(z;\P,\mu)=\int_{\mathcal{A}} \frac{\P[y\1_A(x)]}{\P(A)}\,\1_A(z)\,\nu(\rmd A),\quad z\in[0,1]^p,
\end{equation}
where  $\mu$ and $\nu$ are defined in Equations~\eqref{eq:def-mu} and \eqref{eq:def-nu} respectively.
\end{definition}

\begin{remark}\label{rk:bootstrap} We should emphasize that our setting does not cover Breiman's original random forest \citep{Breiman2004}. Indeed, we do not incorporate in our framework any resampling mechanism such as the bootstrap when forming the prediction $\P[y\1_A(x)]/\P(A)$ on the component $A$. For a finite sample ($\P=\P_n$), we always use the empirical distribution of the original sample, not a bootstrapped version. It is still possible to consider the random partitions obtained from Breiman's algorithm via the bootstrap, but then our construction of the trees and the resulting random forest predictor differ from Breiman's original formulation. 
\end{remark}

In view of the pointwise almost sure convergence $\bar T_M(z)\to\bar T(z)$ stated in Equation~\eqref{eq:cv-rf}, it is natural to ask whether this convergence can be strengthened to hold almost surely for all $z\in[0,1]^p$, or even uniformly on $[0,1]^p$. \citet{scornet2016asymptotics} addresses the former question in his Theorem~3.1, and the latter in Theorem~3.2, where  the standard $\sqrt{M}$ rate of uniform convergence is established under mild assumptions. We propose here an alternative approach, developed in \cite{DD24}, which relies on a Dini-type theorem and yields almost sure uniform convergence without any assumption on the random partition.

\begin{proposition}\label{prop:unif-cv-rf}
Assume that either $\P=\P_n$, or that $\P=\P^*$ and the regression function $x\mapsto \mathbb{E}[Y\mid X=x]$ is bounded on $[0,1]^p$. Then the convergence in~\eqref{eq:cv-rf} holds almost surely uniformly on $[0,1]^p$.
\end{proposition}

\subsection{Specific models and examples}\label{sec:models-and-examples}
We briefly describe some common random forest models from the literature that fit into the framework developed in the preceding section. Recall from Remark~\ref{rk:bootstrap} that, although the bootstrap may be used when constructing the random partition, our framework excludes it when computing the prediction within each partition component. We further comment on this restriction below and compare the two approaches. 

\paragraph{Breiman's random forests \citep{breiman2001random}.} The original random forest algorithm consists of a bootstrap aggregation (bagging) of randomized regression trees. First, bootstrap samples of the data are drawn, indexed by $m=1,\ldots,M$, with corresponding empirical measures $\P_{n}^{(m)}$. 

For each bootstrap sample, a partition is created through successive binary splits based on a greedy criterion: at each step, for a given node $A=\prod_{j=1}^p [a_j, b_j]$, the algorithm chooses a covariate $j$ and a threshold $u$ that maximize the score
\begin{equation} \label{eq:score}
   \Delta_m(j,u)= \frac{\P_{n}^{(m)}[y \1_{A_0}(x)]^2}{\P_{n}^{(m)}(A_0)} \;+\; \frac{\P_{n}^{(m)}[y \1_{A_1}(x)]^2}{\P_{n}^{(m)}(A_1)} ,
\end{equation}
where 
\[
A_0=\{x\in A\colon x_j\leq a_j+u(b_j-a_j)\}\quad\text{and}\quad A_1=\{x\in A\colon x_j> a_j+u(b_j-a_j)\}
\]
denote the subsets resulting from the split defined by $(j,u)$. The splitting procedure is applied recursively until a stopping rule is reached. A node typically stops splitting if it contains fewer than a predefined number of samples (parameter \texttt{min\_samples\_leaf} in \texttt{scikit-learn}'s \texttt{DecisionTreeRegressor}); if the maximum depth $d$ is reached (parameter \texttt{max\_depth}); or if the gain in score falls below a threshold (parameter \texttt{min\_impurity\_decrease}), corresponding to an impurity-based stopping condition. 

The bootstrap induces a randomization of the resulting partition, and an additional randomization is introduced by restricting the search for the best split to a random subset of size \texttt{max\_features}, selected uniformly at random among the $p$ covariates at each binary split. This randomization is key to the diversity of the trees.

Let us stress that, in Breiman's original Random Forest algorithm, the bootstrap is used both to build the partitions and to compute the predictions, whereas our framework covers the case where the bootstrap is used to build the partitions only. More precisely, denoting by $\Pi^{(1)},\ldots,\Pi^{(M)}$ the different partitions, the former random forest predictor can be written as
\[
\widetilde{F}_n(x)=\frac{1}{M}\sum_{m=1}^M \sum_{A\in \Pi^{(m)}} \frac{\P_{n}^{(m)}[y \1_{A}(x)]}{\P_{n}^{(m)}(A)}\1_{A}(x),
\]
whereas the latter is given by
\[
\hat F_n(x)=\frac{1}{M}\sum_{m=1}^M \sum_{A\in \Pi^{(m)}} \frac{\P_{n}[y \1_{A}(x)]}{\P_{n}(A)}\1_{A}(x).
\]

\paragraph{Extra-Trees \citep{geurts2006extremely}.} Extremely randomized trees (Extra-Trees) introduce randomness in a different way than classical random forests. In this case, no bootstrap resampling of the data is performed. The main feature of Extra-Trees is its randomized splitting procedure.  

At each node $A = \prod_{j=1}^p [a_j, b_j]$, a random subset $S \subset [\![1,p]\!]$ of $m$ covariates is chosen (parameter \texttt{max\_features}). For each covariate $j \in S$, a threshold $u_j$ is drawn uniformly at random from the interval $[a_j, b_j]$. Among these $m$ randomly generated candidate splits $(j,u_j)$, the algorithm selects the one that maximizes the score \eqref{eq:score} and uses it to split the node.  

When $m=1$, the splitting variable and threshold are selected uniformly at random, independently of the data. This leads to partitions that do not adapt to the sample and corresponds to the so-called uniform random forests \citep{biau2008consistency}.  

The Extra-Trees algorithm has two main advantages: it considerably speeds up training and introduces an additional regularization effect that helps prevent overfitting. In \texttt{scikit-learn}, the method is implemented under the class \texttt{ExtraTreesRegressor}.  

\paragraph{Softmax Regression Trees \citep{DD24a}.} Softmax regression trees were introduced as an alternative to Extra-Trees, with the aim of providing more tractable mathematical properties while retaining a similar practical behavior. The algorithm does not use resampling and relies on randomized binary splitting.  

At each node, $K$ candidate splits $\zeta_1, \dots, \zeta_K$ are generated and their scores $\Delta_1, \dots, \Delta_K$ are evaluated using \eqref{eq:score}. One of these candidate splits, denoted by $\xi$, is then chosen at random according to the softmax distribution  
\begin{equation}
   \P \bigl( \xi = \zeta_k \bigr) \;=\; \frac{\exp(\beta \Delta_k)}{\sum_{l=1}^K \exp(\beta \Delta_l)}, 
   \qquad 1 \leq k \leq K .
\end{equation}
The parameter $\beta$ is a hyperparameter often referred to as the \emph{temperature}.  When $\beta \to 0$, the selection tends toward a uniform choice among the $K$ candidates, so that softmax trees reduce to uniform random trees. In contrast, when $\beta \to \infty$, the distribution concentrates on the best candidate, recovering the argmax rule and yielding a procedure similar to Extra-Trees.

\subsection{Kernel and RKHS associated with a random partition}
We introduce kernels associated with random partitions and infinite random forests and study some of their basic properties. We also draw connections with the literature on random forest weights and the corresponding kernel.

A \textit{kernel} on a space $D$ refers to a function $k:D\times D\to\mathbb{R}$ which is symmetric and semi-definite positive, in the sense that it satisfies:
\begin{itemize}
    \item $k(z,z')=k(z',z)$, for all $z,z'\in D$;
    \item ${\sum_{1\leq l,l'\leq L}}a_la_{l'}k(z_l,z_{l'})\geq 0$, for all $L\geq 1$, $a_1,\ldots,a_L\in\mathbb{R}$ and $z_1,\ldots,z_L\in D$.
\end{itemize}
We say that the kernel is measurable whenever $D$ is equipped with a $\sigma$-algebra and $k$ is measurable on $D\times D$ with respect to the product $\sigma$-field.

With a random partition $\Pi$ as in Definition~\ref{def:random-partition}, we associate the function 
\begin{align}
k_0(z,z')&=\mathbb{E}_\Pi\Big[\sum_{A\in\Pi}\1_A(z)\1_A(z')\Big],\nonumber \\
&=\int_{\mathcal{A}}\1_A(z)\1_A(z')\nu (\rmd A),\quad z,z'\in [0,1]^p. \label{eq:def-k0}
\end{align}
\cite{scornet2016asymptotics} calls $k_0$ the connection function and provides the  equivalent definition 
\begin{equation}\label{eq:def-k0-bis}
k_0(z,z')=\mathbb{P}_\Pi(z\stackrel{\Pi}{\longleftrightarrow}z'),
\end{equation}
where $z\stackrel{\Pi}{\longleftrightarrow}z'$ means that $z$ and $z'$ lie in the same component of $\Pi$.

With an infinite random forest $\bar T$ as in Definition~\ref{def:random-forest}, we associate the function
\begin{align}
k(z,z')&=\mathbb{E}_\Pi\Big[\sum_{A\in\Pi}\Big(\frac{\1_{\{\P(A)>0\}}}{\P(A)} +\1_{\{\P(A)=0\}} \Big)\1_A(z)\1_A(z')\Big]\nonumber\\
&=\int_{\mathcal{A}}\alpha_\P^2(A)\1_A(z)\1_A(z')\nu (\rmd A),\quad z,z'\in [0,1]^p,\label{eq:def-k}
\end{align}
with $\alpha_\P^2(A)= 1/\P(A)$ if $\P(A)>0$, and $1$ otherwise.

In practice, the random partition often has components with non vanishing $\P$-probability. For instance, in the case of a decision tree, the splitting rules generally avoid creating empty leaves. Then we have the simpler expression 
\[
k(z,z')=\mathbb{E}_\Pi\Big[\sum_{A\in\Pi}\frac{\1_A(z)\1_A(z')}{\P(A)}  \Big].
\]
The expectation defining $k$ is always defined and nonnegative, but the  value $+\infty$ may arise, as in the following simple example.

\begin{example}[Uniform partition in dimension 1.]\label{ex:1} For $p=1$, consider the uniform random partition with depth $d=1$ defined by
\[
\Pi=\{[0,U),(U,1]\} \quad \mbox{with $U$ uniformly distributed on $[0,1]$.}
\]
Since $z\stackrel{\Pi}\longleftrightarrow z'$ if and only $U$ does not lie between $z$ and $z'$, we have
    \[
    k_0(z,z')=1-|z-z'|,\quad 0\leq z,z'\leq 1.
    \]
    Next, we consider two different choices for $\P$ and determine the corresponding function $k$. \\
    For an infinite sample with uniform distribution $\P^*=\mathrm{Unif}([0,1])$, we have
    \begin{align*}
    k^*(z,z')&=\int_{z\vee z'}^1 \frac{1}{u}\,\rmd u+\int_0^{z\wedge z'}\frac{1}{1-u}\,\rmd u\\
    &= -\log(z\vee z')-\log(1-z\wedge z'),
    \end{align*}
    where in particular $k(0,0)=k(1,1)=+\infty$.\\
    For a finite sample with empirical distribution $\P_n=\sum_{i=1}^n\frac{1}{n}\delta_{x_i}$,
    \[
    k_n(z,z')= \int_{z\vee z'}^1 \Big(\frac{\1_{\{\mathrm{F}_n(u)>0\}}}{\mathrm{F}_n(u)}+\1_{\{\mathrm{F}_n(u)=0\}} \Big)\,\rmd u+\int_0^{z\wedge z'}\Big(\frac{\1_{\{\mathrm{F}_n(u)<1\}}}{1-\mathrm{F}_n(u)}+\1_{\{\mathrm{F}_n(u)=0\}} \Big)\,\rmd u,
    \]
    where $\mathrm{F}_n(u)= \P_n([0,u])$ denotes the cumulative distribution of $\P_n$. When $n\to\infty$, the a.s. uniform convergence $\mathrm{F}_n(u)\to u$ on $[0,1]$ implies the a.s.  convergence $k_n(z,z')\to k^*(z,z')$ on $[0,1]^2$, uniformly on compact subsets  that does not contain $(0,0)$ or $(1,1)$. 
\end{example}

\begin{proposition} \label{prop:k_0-k}
Consider the functions $k_0$ and $k$ defined by Equations~\eqref{eq:def-k0} and~\eqref{eq:def-k}, respectively.
\begin{itemize}
    \item The function $k_0$  is a measurable kernel on $[0,1]^p$. It is normalized in the sense that  
    \begin{equation}\label{eq:normalisation-k0}
    k_0(z,z)=1,\qquad z\in [0,1]^p. 
    \end{equation}
    \item The set $D=\{z\in[0,1]^p\colon k(z,z)<\infty\}$ satisfies $\P(D)=1$ and the function $k$ is a measurable kernel on  $D$. It satisfies
    \[
    \int_{[0,1]^p}k(x,x)\,\P(\rmd x)=\int_{\mathcal{A}}\1_{\{\P(A)>0\}}\,\nu(\rmd A) \leq \nu(\mathcal{A})=\mathbb{E}_\Pi[|\Pi|]
    \]
    and the upper bound is finite according to assumption~\eqref{eq:integrability-pi}.
    
    Furthermore, for all $z\in [0,1]^p$, 
    \begin{equation}\label{eq:normalisation-k}
    \int_{[0,1]^p} k(z,x)\,\P(\rmd x)=\int_{\mathcal{A}}\1_{\{\P(A)>0\}}\1_A(z)\,\nu(\rmd A) \leq 1.
    \end{equation}
\end{itemize}
\end{proposition}

\begin{remark} When equality to $1$ holds in Equation~\eqref{eq:normalisation-k}, then the function $x\mapsto k(z,x)$ is a probability density function with respect to $\P$. Let us emphasize two particular cases where this property holds true.
\begin{itemize}
    \item If the components of the random partition $\Pi$ have positive $\P$-probability almost surely,  then $\P(A)>0$ $\nu(\rmd A)$-almost everywhere and Equation~\eqref{eq:normalisation-nu} implies that the integral in \eqref{eq:normalisation-k} is equal to $1$. This occurs in particular for  partitions obtained by recursive binary splitting with a splitting rule avoiding the creation of empty cells.
    \item If $\P(\{z\})>0$, then $\1_{\{\P(A)>0\}}\1_A(z)=\1_A(z)$ and again Equation~\eqref{eq:normalisation-nu} implies that the integral in \eqref{eq:normalisation-k} is equal to $1$. This occurs when $\P=\P_n$ is the empirical distribution of a finite sample and $z$ is a point of this sample. This property is related to the notion of probability weights of a random forest, see Section~\ref{para:weights} below.
\end{itemize}
Note that this property can cause some confusion on the meaning of the term kernel. In this paper we understand the term kernel as a symmetric semi-definite positive function. But we see that, in some cases, $k$ also enjoys the properties of a \emph{probability kernel} (also called \emph{Markov kernel}).
\end{remark}

We next discuss the boundedness of the kernels. The kernel $k_0$ is clearly bounded by $1$, as seen for instance from Equation~\eqref{eq:def-k0-bis}. The kernel $k$, however, may be unbounded, as in Example~\ref{ex:1}; this can occur only in the case where $\P=\P^*$ corresponds to an infinite sample.

\begin{proposition}\label{prop:boundedness}
The kernel $k$ defined in Equation~\eqref{eq:def-k} is bounded as soon as the following condition holds:
\begin{equation}\label{eq:assumption-1}
    \int_{\mathfrak{P}}\max_{A\in\pi} \alpha_\P^2(A) \,\mu(\rmd\pi)<\infty \tag{$A_1$},
\end{equation}
where the integral is an upper bound for $k$.  
In particular, for a finite sample with empirical distribution $\P=\P_n$, the kernel $k$ is bounded by $n$.
\end{proposition}
 Condition~\eqref{eq:assumption-1} is mild. For finite samples, it is automatically satisfied, since $k$ is then bounded by $n$. In the infinite-sample case, it requires only that partition cells with very small $\P$-probability are not too frequent, thereby preventing divergences in the kernel. It also follows from Proposition~\ref{prop:boundedness} that $k$ is bounded by $\int_{\mathcal{A}}\alpha_\P^2(A)\,\nu(\rmd A)$, 
which is an upper bound for the integral in \eqref{eq:assumption-1} expressed in terms of the component intensity measure $\nu$.

\paragraph{Continuity.} 
Continuous kernels on compact domains play a central role in the RKHS literature. For instance, Mercer's theorem applies to such kernels \citep[Theorem 2.10]{ScholkopfSmola2002}. We next consider a condition on the random partition $\Pi$ ensuring continuity of both kernels $k_0$ and $k$.

The boundary of a hyperrectangle $A=[a,b\rangle$ in $[0,1]^p$, denoted by $\partial A$, is the set of points $x\in [a,b]$ with $x_j\in\{a_j,b_j\}\setminus\{0,1\}$ for some $j\in[\![1,p]\!]$. The condition
\begin{equation}\label{eq:assumption-2}
\int_{\mathcal{A}}\1_{\{z\in\partial A\}}\,\nu(\rmd A)= 0,\quad z\in [0,1]^p, \tag{$A_2$}
\end{equation}
is central to our analysis of kernel continuity.

Boundary points $z \in \partial A$ arise when a binary split defining $A$ occurs exactly at the value of one coordinate of $z$, that is, when a coordinate coincides with a splitting threshold. Condition~\eqref{eq:assumption-2} can therefore be ensured by requiring that, in the construction of the randomized partition, the distribution of split positions along each axis is continuous. More formally, for each $j\in[\![1,p]\!]$, we define the measure $\rho_j$ on $(0,1)$ by
\[
\rho_j(B)=\int_{\mathcal{A}} \Big(\delta_{a_j(A)}(B)+\delta_{b_j(A)}(B)\Big)\,\nu(\rmd A),\quad B\subset(0,1)\ \text{Borel},
\]
where $\delta_x$ denotes the Dirac mass at $x$, and for $A=[a,b\rangle=\prod_{j=1}^p[a_j,b_j\rangle$ we set $a_j(A)=a_j$ and $b_j(A)=b_j$. Note that the values $0$ and $1$ are excluded in the definition of $\rho_j$ since they do not correspond to genuine cuts in the partitioning procedure. 

\begin{proposition} \label{prop:continuity}
The following holds:
\begin{itemize}
    \item Under Assumption~\eqref{eq:assumption-2}, $k_0$ is continuous.
    \item Under Assumptions~\eqref{eq:assumption-1} and~\eqref{eq:assumption-2}, $k$ is continuous.
\end{itemize} 
Furthermore, Assumption~\eqref{eq:assumption-2} holds whenever the measures $\rho_j$, $j\in[\![1,p]\!]$, are atomless.
\end{proposition}

\paragraph{RKHS associated with the kernel of a random forest.}
 We briefly recall some basic notions from RKHS theory. More details are to be found, for instance, in \cite{BTA04}, \cite{SC08}, or \cite{HE15}.

An RKHS on a domain $D$  is a subspace $\mathcal{H}$ of the vector space of functions $\{F:D\to\mathbb{R}\}$  endowed with a Hilbert structure such that all evaluation maps $F\mapsto F(x)$ are continuous, for all $x\in D$.
The inner product and norm are denoted by $\langle\cdot,\cdot\rangle_{\mathcal{H}}$ and $\|\cdot\|_{\mathcal{H}}$ respectively.
By Riesz representation theorem, for all $x\in D$, the evaluation map $F\mapsto F(x)$ can be represented by a unique element $k_x\in\mathcal{H}$ such that
\[
  \langle k_x, F\rangle_{\mathcal{H}}=F(x),\quad F\in\mathcal{H}.
\]
Then, the function $k(x,x')=\langle k_x,k_{x'}\rangle_{\mathcal{H}}$ is symmetric and nonnegative definite on $D\times D$, i.e.\ a kernel on $D$. It completely characterizes the RKHS and is called the reproducing kernel of $\mathcal{H}$.
Conversely, Aronszajn's theorem~\citep{A50} states that, for any kernel $k$, one can construct an RKHS that admits $k$ as reproducing kernel; see also \citet[Theorem~2.7.7]{HE15}.

In this paper, we study the RKHSs associated with the kernels $k_0$ and $k$ introduced in the preceding section, and begin by stating some immediate properties. For brevity, we denote by $L^2 = L^2([0,1]^p,\P(\rmd x))$ 
the space of square-integrable functions on $[0,1]^p$ with respect to $\P(\rmd x)$, the first marginal of $\P$. 

\begin{proposition} \label{prop:propRKHS}
    Let $\mathcal{H}_0$ and $\mathcal{H}$ be the RKHSs with reproducing kernels $k_0$ and $k$ defined in Equations~\eqref{eq:def-k0} and~\eqref{eq:def-k}, respectively.  
    \begin{enumerate}[(a)] 

        \item The following properties hold for $\mathcal{H}_0$.
        \begin{enumerate}[i)]
            \item Every $F \in \mathcal{H}_0$ is bounded and measurable on $[0,1]^p$ with $$ \rvert F(x) \rvert \leqslant \rVert F \rVert_{\mathcal{H}_0}, \quad x \in [0,1]^p.$$
            Consequently, $\mathcal{H}_0\subset L^2$ and $$  \rVert F \rVert_{L^2}\leqslant  \rVert F \rVert_{\mathcal{H}_0}.$$
            \item Under Assumption \eqref{eq:assumption-2}, all functions in $\mathcal{H}_0$ are continuous on $[0,1]^p$. 
            \item The constant function belongs to $\mathcal{H}_0.$
        \end{enumerate}
        \item The following properties hold for $\mathcal{H}$.
        \begin{enumerate}[i)]
            \item Every $F \in \mathcal{H}$ is measurable on $D$ and satisfies $$ \rvert F(x) \rvert \leqslant \rVert F \rVert_{\mathcal{H}} \sqrt{k(x,x)}, \quad x \in D.$$
            Consequently, $\mathcal{H}\subset L^2$ and $$  \rVert F \rVert_{L^2} \leqslant m_1  \rVert F \rVert_{\mathcal{H}}, \quad \text{with} \quad m_1 = \bigg( \int_{\mathcal{A}} \1_{\{\P(A)>0\}} \nu (\rmd A)\bigg)^{1/2}.$$
            \item Under Assumption \eqref{eq:assumption-1}, every $ F \in \mathcal{H}$ is bounded in $[0,1]^p$ and $$\|F\|_{\infty} \leq m_2 \|F\|_{\mathcal{H}},  \quad \text{with} \quad m_2 = \bigg(\int_{\mathfrak{P}} \max_{A \in \pi}\Big(\alpha^2_\P(A)\Big) \,\mu(\rmd \pi)\bigg)^{1/2}.$$
            \item Under Assumptions \eqref{eq:assumption-1} and \eqref{eq:assumption-2}, all functions in $\mathcal{H}$ are bounded and continuous on $[0,1]^p$ and bounded subsets of $\mathcal{H}$ are uniformly equicontinuous. 
        \end{enumerate}
        \item Assume that $D = [0,1]^p$. Then, the inclusion $\mathcal{H}_0 \subset \mathcal{H}$ holds with $  \rVert F \rVert_{\mathcal{H}} \leqslant  \rVert F \rVert_{\mathcal{H}_0},$ for $F \in \mathcal{H}_0$.
    \end{enumerate}
\end{proposition}

\color{black}

The following theorem, though simple, is important because it supports the fact that the RKHS associated with the random forest kernel is a very natural space.

\begin{theorem}\label{thm:RKHS-basics}
Consider an infinite random forest $\bar T(\cdot;\P,\mu)$ as in Definition~\ref{def:random-forest}. Let $k$ be the associated kernel  defined in Equation~\eqref{eq:def-k} and $\mathcal{H}$ the RKHS  with reproducing kernel $k$. Then $\bar T(\cdot;\P,\mu)\in\mathcal{H}$ and is given by 
\begin{equation}\label{eq:rf-as-kernel-integral}
\bar T(\cdot;\P,\mu)=\int_{[0,1]^p\times \mathbb{R}}yk(x,\cdot)\, \P(\rmd x,\rmd y),
\end{equation}
with norm $\|\bar T(\cdot;\P,\mu)\|_{\mathcal{H}}^2\leq \P[y^2]$. More precisely, we have the identity
\begin{equation}\label{eq:RKHS-norm-interpretation}
 \P[y^2]=\|\bar T(\cdot;\P,\mu)\|_{\mathcal{H}}^2+\int_{\mathfrak{P}} \P\big[(y-T(x;\P,\pi))^2\big]\,\mu(\rmd\pi).
\end{equation}
\end{theorem}
Equation~\eqref{eq:rf-as-kernel-integral} holds pointwise for all $z\in[0,1]^p$, and also in the stronger sense of equality in the RKHS, where the right-hand side is interpreted as a Bochner integral in the separable Hilbert space $\mathcal{H}$. Indeed, since $\|k(x,\cdot)\|_{\mathcal{H}}=\sqrt{k(x,x)}$, we have
\[
\int_{[0,1]^p}\|y\,k(x,\cdot)\|_{\mathcal{H}}\,\P(\rmd x\,\rmd y)
  \leq \big(\P[y^2]\big)^{1/2} 
             \Big(\int k(x,x)\,\P(\rmd x)\Big)^{1/2}
   <\infty,
\]
which ensures that the Bochner integral is well defined.

Equation~\eqref{eq:RKHS-norm-interpretation} can be interpreted as a decomposition of variance when explaining $Y$ by a randomized tree $T(X;\P,\Pi)$. The RKHS norm is related to a variance explained by the model while the integral term corresponds to a residual variance (mean squared error between the prediction of the randomized tree $T(x; \P, \pi)$ and $y$ averaged with respect to $\P$ and $\mu$). As a consequence, the RKHS squared norm can be seen as the sum of the squared mean $\P[y]^2$ and the mean decrease in impurity across the different splits of the trees of the forest (closely related to the MDI importance criterion).

\begin{remark}[Interpretation in terms of kernel mean embedding]
Equation~\eqref{eq:rf-as-kernel-integral} admits an appealing interpretation in terms of 
\emph{kernel mean embeddings} \citep{Muandet2017}. 
Let $\mathcal{K}$ be the one-dimensional RKHS of linear functions 
$y \mapsto \alpha y$ on $\mathbb{R}$, associated with the linear kernel 
$l(y,y') = yy'$. Consider the tensor product space $\mathcal{H} \otimes \mathcal{K}$, 
which is the RKHS corresponding to the product kernel 
$\kappa((x,y),(x',y')) = k(x,x')\,l(y,y')$. 
The \emph{kernel mean embedding} of the joint distribution 
$\P(\mathrm{d}x,\mathrm{d}y)$ is defined in $\mathcal{H} \otimes \mathcal{K}$ as
\[
\mu_{\P} = \int \kappa((x,y), \cdot)\,\P(\mathrm{d}x, \mathrm{d}y),
\]
and satisfies the reproducing property
\[
\int f(x,y)\,\P(\mathrm{d}x, \mathrm{d}y)
= \langle \mu_{\P}, f \rangle_{\mathcal{H} \otimes \mathcal{K}},
\quad \text{for all } f \in \mathcal{H} \otimes \mathcal{K}.
\]
Because $\mathcal{K}$ is one-dimensional and $\kappa = k \otimes l$, we have
\begin{align*}
\mu_{\P}(x',y') 
&= y' \int y\,k(x,x')\,\P(\mathrm{d}x, \mathrm{d}y)
= \bar{T}(x';\P,\mu)\,y',
\end{align*}
with  $\bar{T}(\cdot;\P,\mu)$ defined by Equation~\eqref{eq:rf-as-kernel-integral}. 
Therefore, for any $g \in \mathcal{H}$ and $f(x,y) = g(x)\,y$, it follows that
\[
\int y\,g(x)\,\P(\mathrm{d}x, \mathrm{d}y)
= \langle \bar{T}(\cdot;\P,\mu), g \rangle_{\mathcal{H}}.
\]
In expectation form, for $(X,Y) \sim \P$,
\[
\mathbb{E}[Yg(X)] 
= \langle \bar{T}(\cdot;\P,\mu), g \rangle_{\mathcal{H}},
\quad \text{for all } g \in \mathcal{H}.
\]
In fact, $\bar{T}(\cdot;\P,\mu) \in \mathcal{H}$ is the \emph{kernel mean embedding} of the signed measure $A\mapsto \mathbb{E}[Y\1_A(X)]$,   $A\subset [0,1]^p$ Borel.
\end{remark}

\smallskip
The proof of Theorem~\ref{thm:RKHS-basics} relies on a general useful representation of the RKHS $\mathcal{H}$.  
For $z\in D$ and $A\in\mathcal{A}$, we set 
\begin{equation}\label{eq:def-g}
g_\P(z,A)= \begin{cases}
    \1_A(z)/\sqrt{\P(A)} & \text{if } \P(A)>0,\\
    \1_A(z) & \text{if } \P(A)=0.
\end{cases}
\end{equation}
so that we have, with this notation, 
\begin{equation}\label{eq:def-k-nu}
k(z,z')=\int_{\mathcal{A}}g_\P(z,A)g_\P(z',A)\,\nu(\rmd A),\quad z,z'\in D.
\end{equation}
The following proposition is a reformulation in our specific framework of Theorem~2.7.7 from \cite{HE15}. We denote by $L^2(\mathcal{A},\nu)$ the Hilbert space of (equivalence classes of) square-integrable functions on $\mathcal{A}$. 

\begin{proposition}\label{prop:RKHS-structure}
The RKHS $\mathcal{H}$  satisfies the following properties:
\begin{enumerate}[i)]
\item For every $G\in L^2(\mathcal{A},\nu)$, the function
\begin{equation}\label{eq:RKHS-representation}
F(z)=\int_{\mathcal{A}} G(A) g_\P(z,A) \,\nu(\rmd A),\quad z\in D,
\end{equation}
belongs to $\mathcal{H}$; conversely, every function $F\in\mathcal{H}$ admits a representation of the form~\eqref{eq:RKHS-representation} for some representative $G\in L^2(\mathcal{A},\nu)$.
\item Every $F\in\mathcal{H}$ admits a unique representative, denoted by $\dot F$, in the subspace 
\begin{equation}\label{eq:def-dot-H}
\dot{\mathcal{H}}= \overline{\mathrm{span}}\left( A\mapsto g_\P(z,A) \;\colon\; z\in D\right)\subset L^2(\mathcal{A},\nu).
\end{equation}
More precisely, $G=\dot F$ is the unique function in $\dot{\mathcal{H}}$ such that equation~\eqref{eq:RKHS-representation}  holds.
\item The mapping $F \mapsto \dot{F}$ defines a one-to-one isometry from $\mathcal{H}$ onto $ \dot{\mathcal{H}}$, and in particular,
\[
\|F\|_{\mathcal{H}}=\|\dot F\|_{L^2(\mathcal{A},\nu)}
\]
\end{enumerate}
\end{proposition}
The representative $G$ of $F$ in Equation~\eqref{eq:RKHS-representation} is not unique, but  it is always related to the unique representative $\dot F$ by the relation
\begin{equation}\label{eq:modulo}
G=\dot F\quad  \text{(mod $\dot{\mathcal{H}}^\perp$)}
\end{equation}
in the sense that $G-\dot F$ is orthogonal to $\dot{\mathcal{H}}$ in $L^2(\mathcal{A},\nu)$.
Equivalently, $\dot F$ is the orthogonal projection of $G$ on the subspace $\dot{\mathcal{H}}$ of $L^2(\mathcal{A},\nu)$.

\subsection{Optimality property of the infinite random forest}\label{sec:fundamental-property}
We provide in this section a characterization of the infinite random forest as the solution of a minimization problem in the RKHS generated by its kernel. 

\begin{theorem}\label{thm:fundamental-property}
The infinite random forest $\bar T(\cdot;\P,\mu )$ from Definition~\ref{def:random-forest} is the unique solution of the penalized optimization problem
\begin{equation}\label{eq:optimization-property}
\texttt{ minimize } -2\P[yF(x)]+\|F\|_\mathcal{H}^2  \texttt{ over } F\in\mathcal{H}. 
\end{equation}
\end{theorem}

The proof of Theorem~\ref{thm:fundamental-property} relies on the following Lemma providing an expression of the mean squared error of any predictor $F\in\mathcal{H}$.

\begin{lemma}\label{lem:identity}
   The following identity holds, for all $F\in\mathcal{H}$:
\begin{align}
&\P[(y-F(x))^2]+\|F\|_{\mathcal{H}}^2-\|F\|_{L^2}^2 \nonumber\\
=&\int_{\mathcal{A}} \P\Big[\Big(y-\frac{\dot{F}(A)}{\sqrt{\P(A)}}\Big)^2\1_A(x) \Big]\,\nu(\rmd A)+ \int_{\mathcal{A}} \dot{F}(A)^2\1_{\{\P(A)=0\}}\,\nu(\rmd A).   \label{eq:identity}
\end{align}
Furthermore, the right hand side is minimized for the function 
\[
\dot F(A)=\frac{\P[y\1_A(x)]}{\sqrt{\P(A)}},\quad A\in\mathcal{A},
\]
which corresponds to the infinite random forest $F=\bar T(\cdot;\P,\mu)$.
\end{lemma}

Several interpretations of this identity can be given. 
In Theorem~\ref{thm:fundamental-property} we highlight the characterization of $\bar T$ as the solution of the optimization problem~\eqref{eq:optimization-property}.  According to Lemma~\ref{lem:identity}, another characterization of $\bar T$ is as the solution of
\begin{equation}\label{eq:optimization-property-bis}
\texttt{minimize } \;\;
\P\big[(y-F(x))^2\big]+\|F\|_{\mathcal{H}}^2-\|F\|_{L^2}^2
\quad \texttt{over } F\in\mathcal{H}.
\end{equation}
This latter optimization problem is reminiscent of ridge regression with penalty $\|F\|_{\mathcal{H}}^2-\|F\|_{L^2}^2$. 
However, it can be shown that the difference of two kernels associated with the quadratic form $\|F\|_{\mathcal{H}}^2-\|F\|_{L^2}^2$ is not positive semi-definite. 
Therefore, although related in spirit, this formulation is not a standard kernel ridge regression problem. 
For this reason, we prefer the formulation~\eqref{eq:optimization-property}, in which the constant terms $P[y^2]$ and  $\P[F(x)^2]-\|F\|_{L^2}^2=0$ have been disregarded.

The penalized objective function $-2\P[yF(x)] + \|F\|_{\mathcal{H}}^2$ from~\eqref{eq:optimization-property}
balances two competing forces. On the one hand, it encourages alignment between the true response $y$ and the prediction $F(x)$ by maximizing the covariance $\P[yF(x)]$. On the other hand, it controls the complexity of $F$ in the RKHS by minimizing the regularization term $\|F\|_{\mathcal{H}}^2$. This trade-off highlights that random forests not only aim to fit the data but also implicitly regularize the predictor through their ensemble structure, thereby mitigating overfitting.

\subsection{Characteristic kernel property}

A fundamental property in kernel methods is that of being \emph{characteristic}. 
A bounded measurable kernel $k$ on a measurable space $\mathcal{X}$ 
is said to be \emph{characteristic} if the kernel mean embedding (KME)
$\Phi \colon \mathcal{P}(\mathcal{X}) \to \mathcal{H}$ defined by
\[
\mathrm{Q} \mapsto \int_{\mathcal{X}} k(\cdot, x)\,\mathrm{Q}(\mathrm{d}x)
\]
is injective on the space $\mathcal{P}(\mathcal{X})$ of probability measures on $\mathcal{X}$. 
In that case, the associated \emph{maximum mean discrepancy} (MMD) between 
$\mathrm{Q}_1, \mathrm{Q}_2 \in \mathcal{P}(\mathcal{X})$,
\[
\mathrm{MMD}(\mathrm{Q}_1, \mathrm{Q}_2)
= \|\Phi(\mathrm{Q}_1) - \Phi(\mathrm{Q}_2)\|_{\mathcal{H}},
\]
defines a distance on $\mathcal{P}(\mathcal{X})$ that admits the equivalent representation
\[
\mathrm{MMD}(\mathrm{Q}_1, \mathrm{Q}_2)
= \sup_{\|F\|_{\mathcal{H}} \le 1}
\Bigg|\int_{\mathcal{X}} F(x)\,\mathrm{Q}_1(\mathrm{d}x)
- \int_{\mathcal{X}} F(x)\,\mathrm{Q}_2(\mathrm{d}x)\Bigg|.
\]

For the kernels introduced in Proposition~\ref{prop:k_0-k}, 
the KME and MMD take simple explicit forms, stated below. 
For clarity, we restrict attention to the case where $k$ is bounded so that its KME is defined on the entire space of probability measures.

\begin{lemma}\label{lem:KME-MMD} ~
\begin{itemize}
    \item The kernel $k_0$ defined by Equation~\eqref{eq:def-k0} admits the KME
    \[
    \Phi(\mathrm{Q})
    = \int_{\mathcal{A}} \mathrm{Q}(A)\,\1_A(\cdot)\,\nu(\mathrm{d}A),
    \qquad \mathrm{Q} \in \mathcal{P}([0,1]^p),
    \]
    and the MMD between $\mathrm{Q}_1, \mathrm{Q}_2 \in \mathcal{P}([0,1]^p)$ satisfies
    \[
    \mathrm{MMD}^2(\mathrm{Q}_1, \mathrm{Q}_2)
    = \int_{\mathcal{A}} 
    \big|\mathrm{Q}_1(A) - \mathrm{Q}_2(A)\big|^2 \,\nu(\mathrm{d}A).
    \]
    \item Suppose that Assumption~\eqref{eq:assumption-1} holds, so that the kernel     $k$ defined by Equation~\eqref{eq:def-k} is bounded.  Then its KME is given by
    \[
    \Phi(\mathrm{Q})
    = \int_{\mathcal{A}}\alpha_{\P}^2(A)\mathrm{Q}(A)\,\1_A(\cdot)\,\nu(\mathrm{d}A),
    \qquad \mathrm{Q} \in \mathcal{P}([0,1]^p),
    \]
    with $\alpha_\P^2(A)$ defined in \eqref{eq:def-k}. The corresponding MMD satisfies
    \[
    \mathrm{MMD}^2(\mathrm{Q}_1, \mathrm{Q}_2)
    = \int_{\mathcal{A}}\alpha_{\P}^2(A)
    \big|
     \mathrm{Q}_1(A) - \mathrm{Q}_2(A)
    \big|^2
    \nu(\mathrm{d}A).
    \]
\end{itemize}
\end{lemma}

As we can see from the preceding expressions, 
the MMD associated with the kernel $k$ 
amplifies the contribution of sets $A$ with small $\P$-probability, 
through the weighting factor $\alpha_\P^2(A)$, 
compared with the MMD associated with $k_0$.

\medskip
We next provide necessary and sufficient conditions for the kernels 
to be characteristic. To this aim, we first introduce the following definitions. 
Recall that $\mathcal{A}$ denotes the set of hypercubes on $[0,1]^p$, 
equipped with the distance $\rho$ defined in~\eqref{eq:def-rho}, and that the block intensity measure 
$\nu$ is a Borel measure on $\mathcal{A}$.

\begin{definition}\label{def:determining-measure} ~
\begin{itemize}
    \item A collection $\mathcal{C}$ of Borel subsets of $[0,1]^p$ 
    is called a \emph{determining class} if, for any two Borel probability measures 
    $\mathrm{Q}_1$ and $\mathrm{Q}_2$ on $[0,1]^p$,
    \[
    \mathrm{Q}_1(A) = \mathrm{Q}_2(A)
    \quad \text{for all } A \in \mathcal{C}
    \quad \Longrightarrow \quad
    \mathrm{Q}_1 = \mathrm{Q}_2.
    \]
    \item A Borel measure $\nu$ on $\mathcal{A}$ is called \emph{determining} 
    if, for any two Borel probability measures $\mathrm{Q}_1$ and $\mathrm{Q}_2$ on $[0,1]^p$,
    \[
    \mathrm{Q}_1(A) = \mathrm{Q}_2(A)
    \quad \text{for $\nu(\mathrm{d}A)$-almost every } A
    \quad \Longrightarrow \quad
    \mathrm{Q}_1 = \mathrm{Q}_2.
    \]
\end{itemize}
\end{definition}

From these definitions, it follows that the support $\mathrm{supp}(\nu)$ 
of a determining measure must itself be a determining class, 
although the converse need not hold. 

\begin{proposition}\label{prop:characteristic-kernel}
The following properties are equivalent:
\begin{enumerate}[i)]
    \item The kernel $k_0$ is characteristic;
    \item The block intensity measure $\nu$ is determining.
\end{enumerate}
If Assumption~\eqref{eq:assumption-1} holds, 
then these properties are also equivalent to:
\begin{enumerate}[i),resume]
    \item The kernel $k$ is characteristic.
\end{enumerate}
\end{proposition}

\begin{example}
It is straightforward to verify that if $\mathrm{supp}(\nu)$ 
contains all hypercubes of the form $[0,x\rangle$ with $x \in [0,1]^p$, 
then $\nu$ is a determining measure. 
As a consequence, random partitions generated by the 
\emph{Extra-Tree} algorithm with depth $d \ge p$, 
or by the \emph{Softmax-Regression-Tree} algorithm with depth $d \ge p$, 
define an intensity measure $\nu$ that is determining and hence a characteristic kernel $k_0$. 
Moreover, replacing the uniform or beta distribution used to sample split locations 
by any distribution with full support on $[0,1]$ does not affect this property. 
We refer to Section~\ref{sec:models-and-examples} for the precise definitions of these algorithms.
\end{example}

The characteristic property of a kernel is closely related to the 
\emph{approximation} or \emph{universality} properties of its RKHS. 
For a detailed discussion on the various notions of universality and their relationships, 
we refer to \citet{Sriperumbudur2011}. 
In particular, \citet{Fukumizu2008} showed that a bounded kernel $k$ 
on a measurable space $\mathcal{X}$ is characteristic 
if and only if the augmented space 
$\mathcal{H} + \mathbb{R} := \{ f + c : f \in \mathcal{H},\, c \in \mathbb{R} \}$ 
is dense in $L^q(\mathcal{X}, \mathrm{Q})$ for all $q \in [1,\infty)$ 
and all $\mathrm{Q} \in \mathcal{P}(\mathcal{X})$.
For continuous kernels on compact spaces $\mathcal{X}$, 
the characteristic property is equivalent to the density of $\mathcal{H}$ 
in the space of continuous functions $C(\mathcal{X}, \mathbb{R})$.

As a consequence of these results, we obtain the following equivalence, 
stated here for reference.

\begin{theorem}\label{thm:characteristic-kernel}
The following properties are equivalent:
\begin{itemize}
    \item[i)] The kernel $k_0$ is characteristic;
    \item[ii)] The block intensity measure $\nu$ is determining;
    \item[iii)] For some (and hence for all) $q \in [1,\infty)$, 
    the RKHS $\mathcal{H}_0$ is dense in $L^q([0,1]^p,\mathrm{Q})$ 
    for every $\mathrm{Q} \in \mathcal{P}([0,1]^p)$.
\end{itemize}
Moreover, if $k_0$ is continuous (e.g., under Assumption~\eqref{eq:assumption-2}), 
these properties are also equivalent to:
\begin{itemize}
    \item[iv)] $\mathcal{H}_0$ is dense in $C([0,1]^p,\mathbb{R})$.
\end{itemize}
\end{theorem}

\medskip
We now discuss situations where the kernel $k_0$ fails to be characteristic, 
in particular when the random partition is generated by a tree of depth $d < p$. 
Such shallow trees are common in gradient boosting methods and correspond to 
restricting the interaction order in the associated additive model. 
If the tree depth is limited to some fixed $d$, then the corresponding partition 
consists of hyperrectangles that cannot involve cuts along all $p$ coordinates, 
but only along a smaller subset $J$ of coordinates. 

More precisely, for a subset $J \subset [\![1,p]\!]$ with cardinality $|J| \le d$, 
let $\mathcal{A}_J$ denote the collection of hyperrectangles obtained by cuts 
on the variables indexed by $J$ only, i.e.\ sets of the form  
\[
A = \prod_{j=1}^p [a_j, b_j\rangle \subset [0,1]^p,
\qquad 
\text{with } a_j = 0,\ b_j = 1 \text{ for all } j \notin J.
\]
Let $\mathcal{A}_d = \bigcup_{|J| \le d} \mathcal{A}_J$ denote the set of hyperrectangles 
with cuts on at most $d$ coordinates. 
Note that both $\mathcal{A}_d$ and the $\mathcal{A}_J$'s are closed subsets of $\mathcal{A}$.

\medskip
We next characterize the expressivity of the RKHS associated with kernels of the form~\eqref{eq:def-k0}. 
For $q \in [1,\infty)$ and a probability measure $\mathrm{Q}$ on $[0,1]^p$, we define
\[
L^q_d([0,1]^p,\mathrm{Q}) 
= \overline{\mathrm{span}}\bigl\{\1_A : A \in \mathcal{A}_d\bigr\}
\subset L^q([0,1]^p,\mathrm{Q}),
\]
the closure in $L^q([0,1]^p,\mathrm{Q})$ of functions involving interactions of order at most $d$. 
Similarly, let $C_d([0,1]^p,\mathbb{R}) \subset C([0,1]^p,\mathbb{R})$ 
denote the subspace spanned by continuous functions depending on at most $d$ coordinates.

\begin{proposition}\label{prop:characteristic-kernel-2}
Assume that the random partition $\Pi$ is generated by a random tree of depth at most $d < p$, 
and consider the kernel $k_0$ defined by Equation~\eqref{eq:def-k0}.  
Then the following properties are equivalent, where item~iii) is considered only 
in the case of a continuous kernel:
\begin{enumerate}[i)]
    \item For all subsets $J \subset [\![1,p]\!]$ with $|J| = d$, 
    the pushforward measure $\nu_J = \nu \circ \tilde{\pi}_J^{-1}$ 
    is determining on the space of hyperrectangles of $[0,1]^J$, where $\tilde{\pi}_J:[a,b\rangle \mapsto [\pi_J(a),\pi_J(b)\rangle$ and $\pi_J:[0,1]^p \to [0,1]^J$ 
    denotes the canonical projection;
    \item For some (and hence for all) $q \in [1,\infty)$, 
    the RKHS $\mathcal{H}_0$ is dense in $L^q_d([0,1]^p,\mathrm{Q})$ 
    for every probability measure $\mathrm{Q}$ on $[0,1]^p$;
    \item $\mathcal{H}_0$ is dense in $C_d([0,1]^p,\mathbb{R})$.
\end{enumerate}
\end{proposition}

\medskip
The proof relies on a decomposition of the form 
$\mathcal{H}_0 =\sum_{|J| = d} \mathcal{H}_0^J$,
where each subspace $\mathcal{H}_0^J$ consists of functions depending only on the coordinates indexed by $J$. 
Theorem~\ref{thm:characteristic-kernel} can then be applied to each of these 
restricted RKHSs, which yields the desired equivalences.

\subsection{Miscellaneous properties}

\paragraph{Comparison of RKHSs.} 
We briefly discuss the inclusion relationships between RKHSs arising from two Random Forest models, where one is a refinement of the other. That is, we say that a partition $\Pi_2$ is a refinement of $\Pi_1$ if for each $A_1 \in \Pi_1$, there exists a partition $D(A_1) \subseteq \Pi_2$ of $A_1$. The partition $D(A_1)$ may contain only $A_1$ if $A_1$ has not been further divided in $\Pi_2$, but can also contain its children. This situation naturally occurs when the two Random Forest algorithms use the same splitting rule but different stopping criteria, with one being less restrictive than the other and thus producing finer partitions. 

\begin{proposition} \label{prop:refinement}
Let $\Pi_1 \sim \mu_1$ and $\Pi_2\sim \mu_2$ be jointly defined random partitions such that $a.s.$, the partition $\Pi_2$ is a refinement of $\Pi_1$. For $A \in \Pi_1$, let $D(A)$ denote the partition of $A$ in $\Pi_2$. The following holds.
\begin{enumerate}[i)]
    \item Denote by $k_0^{(1)}$ and $k_0^{(2)}$ the kernel $k_0$ defined in \eqref{eq:def-k0} associated with $\Pi_1$ and $\Pi_2$ and their respective RKHSs by $\mathcal{H}_0^{(1)}$ and $ \mathcal{H}_0^{(2)}$. Then for all $F \in \mathcal{H}_0^{(1)}$, $$\|F\|_{\mathcal{H}_0^{(2)}} \leq \sqrt{C_0}\|F\|_{\mathcal{H}_0^{(1)}}, \quad \text{with}  \quad C_0= \operatorname*{ess\,sup} \max_{A_1\in \Pi_1} \rvert D(A_1)\rvert,$$
    and so $\mathcal{H}_0^{(1)} \subset \mathcal{H}_0^{(2)}$ whenever $C_0 < \infty$.
    \item Denote by $k^{(1)}$ and $k^{(2)}$ the kernel $k$ defined in \eqref{eq:def-k} associated with $\Pi_1$ and $\Pi_2$ and their respective RKHSs by $\mathcal{H}^{(1)}$ and $ \mathcal{H}^{(2)}$. Assume \eqref{eq:assumption-1} for $\Pi_2$. Then for all $F \in \mathcal{H}^{(1)}$, $$\|F\|_{\mathcal{H}^{(2)}} \leq \sqrt{C_k}\|F\|_{\mathcal{H}^{(1)}}, \quad \text{with}  \quad C_k= \operatorname*{ess\,sup} \max_{A_1\in \Pi_1} \left[  \alpha_\P^2(A_1) \sum_{A_2\in D(A_1)} \dfrac{1}{\alpha_\P^2(A_2) }  \right],$$
    and so $\mathcal{H}^{(1)} \subset \mathcal{H}^{(2)}$ whenever $C_k < \infty$.
    Moreover, for any $z \in [0,1]^p$, we also have $k^{(1)}(z,z) \leq C_k k^{(2)}(z,z)$.
\end{enumerate}
\end{proposition}

\begin{remark}
    The proof of Proposition \ref{prop:refinement} relies on showing that there exists a constant $C>0$ such that $Ck_2-k_1$ is semi-definite positive, which is necessary and sufficient for the inclusion $\mathcal{H}_1 \subset\mathcal{H}_2$, see e.g., Theorem 12 of \cite{BTA04}.
    
In the case where $\mu_2$-\textit{a.e.}, $\P(A)>0$ for all $A \in \Pi_2$, then it holds that $k_2 - k_1$ is semi-definite positive and thus, the inclusion $\mathcal{H}^{(1)} \subset \mathcal{H}^{(2)}$ holds with $\|F\|_{\mathcal{H}_2} \leq \|F\|_{\mathcal{H}_1}$. This follows by noting that, if $\mu_2$-\textit{a.e.}, $\P(A)>0$ for all $A \in \Pi_2$, then for any $A_1 \in \Pi_1$, $$ \alpha_\P^2(A_1) \sum_{A_2\in D(A_1)} \dfrac{1}{\alpha_\P^2(A_2) } = \dfrac{1}{\P(A_1)} \sum_{A_v \in D(A_1)} \P(A_2) = 1$$ by additivity of $\P$ and noting that $D(A_1)$ partitions $A_1$, and so $C_k=1$ in this case. Moreover, for all $z, z' \in [0,1]^p$, the pointwise inequality $k_1(z,z)\leqslant k_2(z,z)$ holds, with strict inequality whenever the event $\{\P (A_2(z)) < \P(A_1(z))\}$ has positive probability, with $A_i(z)$ denoting the node in the partition $\Pi_i$ containing $z$.
\end{remark}

\paragraph{Weight matrix and kernel operator.} \label{para:weights} 
We introduce the \emph{kernel operator} associated with an infinite random forest, which is a natural generalization of the \emph{weight matrix}. For a finite sample $D_n$ with empirical measure $\P_n$, the weight matrix of the infinite random forest is defined as $W = (W_{ni}(X_j))_{1 \leq i,j \leq n}$, with weights 
\begin{equation}\label{eq:def-weights}
    W_{ni}(x) = \E_\Pi \left[ \sum_{A\in \Pi}\frac{\1_{A}(X_i) \1_{A}(x) }{\P_n(A)}\right], \quad 1 \leq i \leq n.
\end{equation}
As discussed in the introduction, the regression forest predictor can be interpreted as a \emph{local averaging predictor}. Local averaging predictors and their analysis is a central concept in nonparametric regression; we explore here their relationships with the random forest kernels; we refer the reader to Chapter 4-6 of \cite{Gyorfi} and Chapter 6 of \cite{bach2024learning} for pedagogical treatments of local averaging methods in nonparametric regression. 

The prediction vector $\hat Y = (\hat F(X_i))_{1 \leq i \leq n}$ at the sample points is related to the response vector $Y = (Y_i)_{1 \leq i \leq n}$ by $\hat Y = W Y$. The apparent linearity is only superficial: $\hat Y$ is linear in $Y$ only when the random partition $\Pi$ is independent of $Y$. If $\Pi$ depends on $Y$, then $W$ is data-dependent and $\hat Y$ is no longer a linear function of $Y$.

\medskip

 From the RKHS point of view, we consider the integral operator $K\colon L^2 \to L^2$ associated with the kernel $k$, defined for $F \in L^2$ by
\[
(KF)(z) = \int_{[0,1]^p} k(x,z) F(x)\, \P(\rmd x), \qquad z\in [0,1]^p.
\]
Recall that we use here the short notation $L^2=L^2(\P(\rmd x))$.
For a finite sample $D_n$ with empirical measure $\P=\P_n$, we can identify a function $F \in L^2$ with the vector $(F(X_i))_{1 \leq i \leq n} \in \mathbb{R}^n$. Then, the matrix associated with the operator $K$ under this identification is precisely the weight matrix $W$. In this sense, the kernel operator $K$ can be seen as a generalization of the weight matrix $W$.

The properties of kernel operators have been studied extensively in functional analysis; we refer, for instance, to \citet[Chapter 4]{HE15} for a general presentation. In the case of kernels associated with an infinite Random Forest, the main results are summarized in the following proposition.

\begin{proposition} \label{prop:integralOP}
Let $k$ be the kernel of the infinite Random Forest defined by Equation~\eqref{eq:def-k} and $K$ the corresponding integral operator.
\begin{enumerate}[i)]
    \item The operator $K$ is self-adjoint, non-negative definite and compact on $L^2$, with image included in $\mathcal{H}$.
    As an operator on $\mathcal{H}$, we can write $K=i^*i$, where $i:\mathcal{H}\to L^2$ is the inclusion operator, and $K\mathcal{H}$ is dense in $\mathcal{H}$.
    \item There exists a sequence of eigenvalues $\lambda_1 \geq \lambda_2 \geq \cdots > 0$ with $\sum_{i\geq 1} \lambda_i^2<\infty$ and an orthonormal (in $L^2$) system of corresponding eigenfunctions $e_1, e_2, \ldots$ in $\mathcal{H}$ such that
    \[
    K = \sum_{i \geq 1} \lambda_i e_i \otimes e_i, \quad \text{i.e., }  KF = \sum_{i \geq 1} \lambda_i \langle F, e_i \rangle_{L^2} e_i \text{ for every } F \in L^2.    \]
    \item Under Assumptions~\eqref{eq:assumption-1}--\eqref{eq:assumption-2}, the eigenvalues satisfy $\sum_{i\geq 1}\lambda_i <\infty$, the $e_i$ can be assumed to be continuous, and the kernel admits the decomposition
    \[
      k(z, z') = \sum_{i \geq 1} \lambda_i e_i(z) e_i(z'), \quad z, z' \in [0,1]^p,
    \]
    in the space $\mathcal{C}([0,1]^p\times [0,1]^p, \mathbb{R})$.
\end{enumerate}
\end{proposition}

\paragraph{Effective sample size.} We highlight interesting connections between the kernel operator and the concepts of \emph{effective sample size}.  
Recall that for a finite sample $D_n$, the vector of predictions can be written $\hat Y=WY$. The local average formula $ \widehat{Y}_i = \sum_{j=1}^n W_{nj}(X_i) Y_j$ gives rise to the notion of  \emph{local effective sample size}, defined as the reciprocal of the sum of squared weights:
\[
N_{\mathrm{eff},i} = \left( \sum_{j=1}^n W_{ni}(X_j)^2 \right)^{-1}. 
\]
The intuition comes from the $k$-nearest neighbors algorithm, where exactly $k$ weights are equal to $1/k$, yielding an effective sample size of $k$. At the sample level, we define the notion of global effective sample size as the harmonic mean:
\begin{equation} \label{def:Neff}
    N_{\mathrm{eff}} = n \left( \sum_{i=1}^n N_{\mathrm{eff},i}^{-1} \right)^{-1}.
\end{equation}
The harmonic mean is chosen since a straightforward computation shows that
\[
N_{\mathrm{eff}} = n \left( \sum_{i=1}^n \sum_{j=1}^n W_{ni}(X_j)^2 \right)^{-1} = \frac{n}{\mathrm{Tr}(W^* W)}.
\]
We recognize in the term $\mathrm{Tr}(W^* W)$ the squared Frobenius norm of the matrix $W$. In the infinite dimensional case, this Frobenius norm of $W$ generalizes into the Hilbert-Schmidt norm of $K$, and we are led to consider
\[
\|K\|_{HS}^2=\mathrm{Tr}(W^* W).
\]

Intuitively, $N_{\mathrm{eff}}$ measures how many training points, on average, each prediction is effectively based on. The following proposition shows that the effective sample size is usually bounded by $1$ and $n$, which is consistent with a notion of sample size. 
\begin{proposition} \label{prop:eff_sampleSize}
Let $\{W_{ni}(x)\}_{i=1}^n$ be local averaging weights such that, for all $j \in [\![1,n]\!]$,
    $$
    \text{(a)} \quad W_{ni}(X_j) \geqslant 0, 
    \qquad 
    \text{(b)} \quad \sum_{i=1}^n W_{ni}(X_j) = 1.
    $$
    Then,
    $$
    1 \leqslant N_{\mathrm{eff}} \leqslant n.
    $$
\end{proposition}
The lower bound $N_{\rm eff}=1$ occurs when, for every in-sample prediction, only one element is used. The upper bound $N_{\rm eff}=n$ happens when the weights of each in-sample prediction are distributed uniformly, thus leading to $\widehat{F}(X_i, D_n) = \bar{X}_n$, the sample mean.

\begin{remark}
    Although we define the effective sample size in the context of Random Forests, the concept naturally extends to any \emph{local averaging predictor}; that is, any regression estimator expressible in the form
$\widehat{F}(x, D_n) = \sum_{i=1}^n W_{ni}(x) Y_i$
for some weights $\{W_{ni}(x)\}_{i=1}^n $. This includes a broad class of methods such as \emph{linear regression}, $k$-nearest neighbors, regression trees, Random Forests, and Nadaraya–Watson kernel estimators, among others. For example, the effective sample size recovers $N_{\mathrm{eff}} = n/p$ in linear regression and $N_{\mathrm{eff}} = k $ in $k$-nearest neighbors. 
\end{remark}

We conclude our discussion on effective sample size with an observation concerning pointwise confidence intervals. 
Let $\widehat{F}_n(x) = \sum_{i=1}^n W_{ni}(x) Y_i$ denote a generic local averaging predictor, where, for simplicity, the weights $(W_{ni}(x))_{i=1}^n$ are assumed to be independent of $Y$ given $X_1, \ldots, X_n$, and the data are homoscedastic with variance $\sigma^2$. 
Conditionally on $X_1, \ldots, X_n$, one then has
\[
\mathrm{Var}\!\left(\widehat{F}_n(X_i, D_n) \,\middle|\, X_1, \ldots, X_n\right)
= \frac{\sigma^2}{N_{\mathrm{eff},i}}, 
\qquad i = 1, \ldots, n.
\]
It is worth noting that this relationship can continue to hold even when the partitions depend on $Y$, for instance, by using either honest forests \citep{wager2018estimation} or cross-fitting \citep{chernozhukov2018double}. 
This simple observation suggests a natural way to construct pointwise confidence intervals by consistently estimating $\sigma^2$. 
A detailed investigation of this direction, however, lies beyond the scope of the present work and is left for future research.

\paragraph{A geometric measure of variable importance.} A central question in the analysis of tree ensembles is to quantify the contribution of each covariate $X_j$ to the fitted model. In practice, the two most widely used variable importance measures are the Mean Decrease in Impurity (MDI) and the Mean Decrease in Accuracy (MDA) \citep{breiman2001random}. MDI summarizes the total decrease in empirical impurity (e.g., variance) attributable to each covariate across all forest splits, and is computed directly from the fitted trees. MDA, on the other hand, measures the loss of predictive performance after permuting the $j$-th variable, and is therefore based on a perturbation of the response. While both measures are popular and effective in many applications, they also suffer from well-known limitations: MDI is sensitive to tree depth, splitting bias, and correlation between covariates, while MDA is computationally expensive and can be unstable; we refer to \cite{scornet2023trees} and the references therein for a recent advancement on the topic.  Both metrics focus solely on predictive relevance: they measure how much a variable matters for predicting $Y$, but not how much it contributes to the geometry of the fitted model. \\

To complement these measures, we introduce a geometric importance score
$\mathrm{GVI}$ based on the forest weight matrix. Let $X_{(j)} \in \mathbb{R}^n$ denote the $j$-th column of the design matrix $X$ and
$H = I_n - n^{-1}\1\1^\top$ be the centering matrix, so
that $X_{(j,c)} = HX_{(j)}$ is the centered version of $X_j$. We define the
geometric variable importance of $X_{(j)}$ as the proportion of its variance
preserved by $W$, that is
\begin{equation} \label{eq:VI}
    \mathrm{GVI}(j)
   =
    \frac{\mathrm{Var}\!\left(W X_{(j,c)}\right)}
         {\mathrm{Var}\!\left(X_{(j,c)}\right)}.
\end{equation}
This quantity measures how much of the variation in $X_{(j)}$ is aligned with the
geometry induced by the Random Forest through its weight matrix $W$. Once the forest is trained, the score $\mathrm{GVI}(j)$ depends only on $W$ and on
$X$. It quantifies how well the variable $X_{(j)}$ is aligned with the 
geometry of the model, that is, how smoothly $X_{(j)}$ varies over the
neighborhoods induced by $W$. In contrast with MDI and MDA, which assess
predictive influence, $\mathrm{GVI}(j)$ captures structural alignment with the
random forest kernel. The two notions are complementary: when they agree, the
role of a variable is unambiguous; when they differ, $\mathrm{GVI}(j)$ might isolate
geometric effects that predictive scores alone may not reveal.

\begin{proposition}\label{prop:VI}
    Let $\rm GVI$ be the variable importance criterion defined in \eqref{eq:VI} associated with either a finite or an infinite Random Forest. The following properties hold.
    \begin{enumerate}[i)]
        \item The $\rm GVI$ criterion is the Rayleigh quotient of $W^2$ and $X_{(j,c)}$, i.e., 
        $$ \mathrm{GVI}(j)
    = \dfrac{X_{(j,c)}^\top W^2X_{(j,c)}}{X_{(j,c)}^\top X_{(j,c)}}, \qquad 1\leqslant j \leqslant p.$$
    \item The $\rm GVI$ criterion is bounded as follows: $$ 0 \leqslant \mathrm{GVI}(j)\leqslant 1, \qquad 1\leqslant j \leqslant p.$$
    \item The $\rm GVI$ criterion is invariant to centering and scaling, that is, for $a\in \mathbb{R}^*$ and $b\in \mathbb{R}^n$, $$\mathrm{GVI}(aX_{(j)}+b) =\mathrm{GVI}(X_{(j)}), \qquad 1\leqslant j \leqslant p.$$
    \end{enumerate}
\end{proposition}

The above proposition shows that $\mathrm{GVI}(j)$ measures how aligned $X_{(j)}$ is with the directions that the forest's geometry considers important. Informally, the weight matrix induces a notion of neighborhood, where the "neighbors" of $X_i$ are the points $X_k$ such that $W_k(X_i)$ is "large" (or, equivalently, those for which $k(X_i, X_k)$ is large). It is close to $1$ when averaging within neighborhoods does not reduce much the variability of $X_{(j)}$, that is, $X_{(j)}$ should be close to constant within each neighborhood, and most of its variations must happen between neighborhoods. On the other hand, it is close to $0$ if the points that the forest treats as "similar" do not have similar values of $X_{(j)}$. Finally, the invariance property makes $\rm GVI$ comparable across features, unlike split-based impurity, which might be biased due to the possibly different scales. 

Note also that $\rm GVI$ could be applied to other local averaging predictors than Random Forests. This is not the case for MDI, which can be computed for trees only. MDA, on the other hand, could technically be computed for any predictive model. However, its computation may be intensive in general; $\rm GVI$ does not suffer from this drawback and could even be used on new collected covariates, initially not included in the model.

\section{Infinite dimensional gradient flows associated with gradient boosting}\label{sec:IGB}
In this section, we build on the results from the previous section concerning Random Forests and their associated RKHS to analyze gradient boosting \citep{friedman2001greedy,F02}, with a particular focus on a continuous variant introduced in \citet{DD24,DD24a}, known as \emph{infinitesimal gradient boosting}. After presenting this algorithm, we introduce a Hilbert manifold associated with the Random Forest RKHS previously defined. We then state the main result of this section, Theorem~\ref{thm:igb-gradient-flow}, which states that infinitesimal gradient boosting corresponds to the gradient flow of the risk on this Hilbert manifold.

\subsection{Gradient properties}\label{sec:gradient-properties}

Our aim here is to show that the Hilbert space structure introduced via the RKHS in the preceding section allows us to relate the infinite Random Forest $\bar{T}(\cdot; \mathrm{P})$ to the gradient of a loss function. We begin by considering regression Random Forests with the mean squared error, and then discuss the case of more general loss functions, with gradient Random Forests closely related to generalized Random Forests \citep{Athey2019grf} and gradient boosting \citep{friedman2001greedy}.

Our first result focuses on least-squares regression, where the risk of a predictor $F \in L^2$ is its mean squared error, defined by
\[
R_{\mathrm{P}}(F) = \frac{1}{2}\,\mathrm{P}\big[(y - F(x))^2\big].
\]
We consider the regression forest $\bar{T}(\cdot; \mathrm{P},\mu)$ introduced in Definition~\ref{def:random-forest} together with the RKHS $\mathcal{H}$ introduced in Proposition~\ref{prop:propRKHS}. Recall that $\mathcal{H}$ is continuously embedded in $L^2$. 

\begin{theorem}\label{thm:rf-as-gradient}
The mean squared error $R_{\mathrm{P}}$ is Fréchet differentiable on $\mathcal{H}$, and $\bar{T}(\cdot; \mathrm{P},\mu) \in \mathcal{H}$ coincides with the negative gradient of $R_{\mathrm{P}}$ at~$0$, that is,
\[
\bar{T}(\cdot; \mathrm{P},\mu) = -\,\nabla R_{\mathrm{P}}(0).
\]
\end{theorem}

Motivated by the analysis of gradient boosting beyond least-squares regression with risk given by the mean squared error, we now introduce the notion of a \emph{gradient Random Forest} and extend the preceding result to this more general framework. 

Consider a general loss function 
$\ell : \mathbb{R} \times \mathbb{R} \to \mathbb{R}$, 
$(z, y) \mapsto \ell(z, y)$, 
assumed convex and twice continuously differentiable 
in its first argument. 
For a predictor $F \in L^2$, 
define the associated risk under $\mathrm{P}$ by
\begin{equation}\label{eq:def-general-risk}
R_{\mathrm{P}}(F) = \mathrm{P}[\ell(F(x), y)].
\end{equation}

To ensure that $R_{\mathrm{P}}(F)$ is finite for all $F \in L^2$, 
we assume that it is finite for $F = 0$ and impose mild regularity conditions on the derivatives of $\ell$. More precisely, we assume that 
\begin{equation}\label{eq:assumptions-on-ell}
\mathrm{P}\big[|\ell(0, y)|\big]<\infty, 
\qquad 
\mathrm{P}\big[|\partial_1 \ell(0, y)|^2\big] < \infty,
\qquad 
\sup_{z,y} |\partial_1^2 \ell(z, y)| < \infty.
\end{equation}
Then the inequality   
\[
|\ell(F(x), y)| 
\le |\ell(0, y)| + |\partial_1\ell(0, y)|\,|F(x)| + \frac{C}{2}|F(x)|^2,
\]
with $C > 0$ the bound on the second order derivative, implies that $R_{\mathrm{P}}(F)$ is finite for all $F \in L^2$. 
These conditions also ensure the square integrability of the pseudo-residuals in the following definition and guarantee that the construction of gradient trees is well-posed.

\begin{definition}\label{def:gradient-tree-and-forest}
Let $\ell$ be a loss function satisfying \eqref{eq:assumptions-on-ell}.
\begin{itemize} 
\item A (randomized) \emph{gradient tree} at an initial predictor $F \in L^2$ is obtained by fitting a (randomized) regression tree to the pseudo-residuals 
$r = -\partial_1 \ell(F(x), y)$. 
More precisely:
\begin{itemize}
\item The partitioning algorithm takes into account the pairs $(x, r)$ of covariates and pseudo-residuals, producing a randomized partition $\Pi$ of $[0,1]^p$. The distribution of $\Pi$ is denoted by $\mu_{\mathrm{P},F}$ and may depend on $\mathrm{P}$ and $F$; the corresponding block intensity measure is denoted by $\nu_{\mathrm{P},F}$.
\item The prediction at a leaf $A \in \mathcal{A}$ is the mean pseudo-residual on that leaf, i.e.
\[
\frac{\mathrm{P}[r\,\1_A(x)]}{\mathrm{P}(A)}
= - \frac{\mathrm{P}[\partial_1 \ell(F(x), y)\,\1_A(x)]}{\mathrm{P}(A)},
\]
with the convention $0/0=0$ for empty leaves.
\item The resulting gradient tree is defined by
\[
T(z; \mathrm{P}, F, \Pi)
= -\sum_{A \in \Pi}
\frac{\mathrm{P}[\partial_1 \ell(F(x), y)\,\1_A(x)]}{\mathrm{P}(A)}\,
\1_A(z).
\]
\end{itemize}
\item The \emph{gradient Random Forest} is the aggregation of (infinitely many) gradient trees, and is defined, analogously to Equation~\eqref{eq:def-bar-T}, by
\[
\bar T(z; \mathrm{P}, F, \mu_{\mathrm{P},F}) 
= -\int_{\mathcal{A}} 
\frac{\mathrm{P}[\partial_1 \ell(F(x), y)\,\1_A(x)]}{\mathrm{P}(A)}\,
\1_A(z)\, \nu_{\mathrm{P},F}(\mathrm{d}A).
\]
\item Analogously to Equation~\eqref{eq:def-k-nu}, the RKHS $\mathcal{H}_{\mathrm{P},F}$ corresponds to the kernel
\[
k_{\mathrm{P},F}(z,z')
= \int_{\mathcal{A}} g_{\mathrm{P}}(z,A)\,g_{\mathrm{P}}(z',A)\,
\nu_{\mathrm{P},F}(\mathrm{d}A). 
\]
\end{itemize}
\end{definition}

The least-squares regression case discussed previously corresponds to 
$\ell(z, y) = \frac{1}{2}(z - y)^2$. 
Then the pseudo-residuals 
\[
r = -\partial_1\ell(F(x), y) = y - F(x)
\]
coincide with the usual notion of residuals in regression. 
Theorem~\ref{thm:rf-as-gradient} naturally extends to this general setting. 
To ensure differentiability of the risk, we additionally assume that 
$(z,y)\mapsto \partial_1^2\ell(z, y)$ is sufficiently regular 
(for instance, Lipschitz in $z$), so that
\begin{equation}\label{eq:regularity-partial-ell}
\sup_{z,y} 
\frac{1}{h} 
\Big|
  \partial_1\ell(z+h, y) - \partial_1\ell(z, y) - h\,\partial_1^2\ell(z, y)
\Big|
\longrightarrow 0 \quad \text{as } h \to 0.
\end{equation}
Equivalently, the Taylor expansion 
$\partial_1\ell(z+h, y) = \partial_1\ell(z, y) + h\,\partial_1^2\ell(z, y) + o(h)$ 
holds uniformly in $(z,y)$ as $h \to 0$.

\begin{theorem}\label{thm:grf-as-gradient}
Assume that the loss function $\ell$ satisfies 
Assumptions~\eqref{eq:assumptions-on-ell}--\eqref{eq:regularity-partial-ell}, 
and consider the risk $R_{\mathrm{P}}$ defined by 
Equation~\eqref{eq:def-general-risk}. 
For $F \in L^2$, let 
$\bar T(\cdot; \mathrm{P}, F, \mu_{\mathrm{P},F})$ 
and $\mathcal{H}_{\mathrm{P},F}$ be defined as in 
Definition~\ref{def:gradient-tree-and-forest}. 

Then the risk functional $R_{\mathrm{P}}$ is Fréchet differentiable on 
$\mathcal{H}_{\mathrm{P},F}$, and 
$\bar T(\cdot; \mathrm{P}, F, \mu_{\mathrm{P},F}) \in \mathcal{H}_{\mathrm{P},F}$ 
coincides with the negative gradient of $R_{\mathrm{P}}$ at $F$, i.e.
\[
\bar T(\cdot; \mathrm{P}, F, \mu_{\mathrm{P},F}) 
= -\,\nabla R_{\mathrm{P}}(F).
\]
\end{theorem}

\subsection{Infinitesimal gradient boosting model}

Gradient boosting is a machine learning technique that builds a strong predictive model by sequentially combining weak learners, typically shallow decision trees. At each iteration, the algorithm fits a new model to the negative gradient (i.e., the residual errors) of the loss function with respect to the current prediction. This process incrementally improves performance by correcting the errors of previous models, effectively minimizing the overall loss. Gradient boosting was originally introduced by Friedman~\citep{friedman2001greedy,F02} and is widely used for both regression and classification tasks due to its flexibility and high predictive accuracy.

\emph{Infinitesimal gradient boosting} (IGB)~\citep{DD24,DD24a} is a continuous-time variant of the standard gradient boosting algorithm, obtained in the limit as the learning rate tends to zero while the number of iterations is rescaled and tends to infinity. To ensure a well-defined limit, the method employs \emph{softmax gradient trees}, in which the selection of split points during tree construction is performed using a \emph{softmax selection} instead of the traditional \emph{argmax selection}. This relaxation is crucial, as the softmax operator is Lipschitz-continuous, whereas the argmax operator is discontinuous.

More precisely, fix the parameters of the softmax gradient tree: depth $d \ge 1$, number of candidate splits $K \ge 1$, and softmax temperature parameter $\beta > 0$ (see Section~\ref{sec:models-and-examples} for details). Using the notation from Section~\ref{sec:gradient-properties}, let $\ell: \mathbb{R} \times \mathbb{R} \to \mathbb{R}$ be a loss function, and let $\mathrm{P} \in \mathcal{P}([0,1]^p \times \mathbb{R})$ denote the data distribution. The softmax gradient tree at a given predictor $F \in L^2$ is defined by
\begin{equation}\label{eq:softmax-gradient-tree}
T(z; \mathrm{P}, F, \Pi)
= -\sum_{A \in \Pi} 
  \frac{\mathrm{P}[\partial_1 \ell(F(x), y)\,\1_A(x)]}{\mathrm{P}(A)}\,\1_A(z),
\end{equation}
where the partition $\Pi$ is obtained by recursive binary splitting of depth $d \ge 1$ using softmax selection (with parameters $K$ and $\beta$) applied to the pseudo-residuals $r=-\partial_1 \ell(F(x), y)$. The dependence on $F$ is essential and is emphasized in the notation. In the following, $\mu_{\P,F}$ and $\nu_{\P,F}$ denote the partition distribution and its block intensity measure for this softmax gradient tree model.

Gradient boosting with learning rate $\lambda > 0$ produces a sequence of predictors $(\hat{F}_k^\lambda)_{k \ge 0}$ defined recursively from an initialization $\hat{F}_0$ by
\[
\hat{F}_{k+1}^\lambda 
= \hat{F}_k^\lambda + \lambda\, T(\cdot; \mathrm{P}, \Pi_{k+1}, \hat{F}_k^\lambda),
\quad k \ge 0,
\]
where $(\Pi_k)_{k \ge 1}$ are independent random partitions with $\Pi_{k+1}\sim \mu_{\P,\hat{F}_{k}^\lambda}$. The initialization is typically chosen as the constant function minimizing the empirical risk:
\begin{equation}\label{eq:igb-initialization}
\hat{F}_0 = \mathop{\mathrm{arg\,min}}_{z \in \mathbb{R}} \mathrm{P}[\ell(z, y)].
\end{equation}

The sequence $(\hat{F}_k^\lambda)_{k \ge 0}$ thus forms a Markov chain. The infinitesimal learning rate regime corresponds to the fluid limit as $\lambda \to 0$ and $k = \lfloor t / \lambda \rfloor \to \infty$. It was shown in~\cite{DD24,DD24a} that the limit
\[
\lim_{\lambda \to 0} 
\hat{F}_{\lfloor t / \lambda \rfloor}^\lambda(x)
=: \hat{F}_t(x), 
\qquad x \in [0,1]^p,\ t \ge 0,
\]
exists and that convergence is uniform on compact sets $[0,1]^p \times [0,T]$ for all $T > 0$. The limit process $(\hat{F}_t)_{t \ge 0}$ is deterministic (i.e., it does not depend on the partition randomness, though it depends on the data distribution $\mathrm{P}$), jointly continuous in $(x,t)$, and is called the \emph{infinitesimal gradient boosting process}. It is characterized as the unique solution to the differential equation
\begin{equation}\label{eq:igb-ODE}
\frac{\mathrm{d} F_t}{\mathrm{d} t}
= \bar{T}(\cdot; \mathrm{P}, F_t,\mu_{\P,F_t}),
\quad t \ge 0,
\end{equation}
with initialization $F_0 = \hat{F}_0$ at time $t = 0$.  
The dynamics are governed by the \emph{softmax gradient forest}
\begin{equation}\label{eq:softmax-gradient-forest}
\bar{T}(z; \mathrm{P}, F,\mu_{\P,F_t})
= \int_{\mathfrak{P}} T(z; \mathrm{P}, F, \pi)\,\mu_{\P,F_t}( \mathrm{d}\pi),
\end{equation}
which is the expectation of the softmax gradient tree~\eqref{eq:softmax-gradient-tree}.

\begin{definition}\label{def:igb}
The \emph{infinitesimal gradient boosting process} $(\hat{F}_t)_{t \ge 0}$ is defined as the unique solution to the ordinary differential equation~\eqref{eq:igb-ODE} in the function space $\mathcal{C}([0,1]^p, \mathbb{R})$, with initialization~\eqref{eq:igb-initialization}, where the vector field $\bar{T}$ is given by the softmax gradient forest defined in~\eqref{eq:softmax-gradient-forest}.
\end{definition}

This result holds both at the sample level (e.g., for the empirical distribution $\mathrm{P} = \mathrm{P}_n$) and at the population level (e.g., for the true distribution $\mathrm{P} = \mathrm{P}^*$). It was shown in~\cite{DD24a} that, as $n \to \infty$, infinitesimal gradient boosting with finite sample size converges uniformly to its population counterpart, and that the limiting process is the solution of the same differential equation~\eqref{eq:igb-ODE} with $\mathrm{P} = \mathrm{P}^*$.

\subsection{A Hilbert manifold}

We now revisit the theory of infinitesimal gradient boosting in light of our results on the RKHS associated with infinite Random Forests.

According to Definition~\ref{def:gradient-tree-and-forest}, we denote by $\mathcal{H}_{\P,F}$ the RKHS associated with the softmax gradient Random Forest at a given predictor $F\in L^2$. This space depends explicitly on $\P$ and $F$, and implicitly on the hyperparameters $K \ge 1$ and $\beta \ge 0$. In the case of a totally Random Forest ($K=1$ or $\beta=0$), the partition distribution $\mu_{\P, F}$ no longer depends on $\P$ or $F$, and consequently $\mathcal{H}_{\P,F}$ becomes independent of $F$. We then denote this reference space by $\mathcal{H}_{\P}^0$.


\begin{lemma}\label{lem:RKHS-igb}
For all $F \in L^2$, the RKHSs $\mathcal{H}_{\P,F}$ and $\mathcal{H}_{\P}^0$ contain the same set of functions (although their inner products may differ), and their norms are equivalent. That is, there exists a constant $c_{\P,F} > 0$ such that
\[
c_{\P,F}^{-1} \, \|G\|_{\mathcal{H}_{\P}^0}
\;\le\;
\|G\|_{\mathcal{H}_{\P,F}}
\;\le\;
c_{\P,F} \, \|G\|_{\mathcal{H}_{\P}^0},
\qquad  G \in \mathcal{H}_{\P}^0.
\]
Moreover, the mapping
\[
F \in L^2
\;\longmapsto\;
\langle \cdot, \cdot \rangle_{\mathcal{H}_{\P,F}}
\in S_2(\mathcal{H}_{\P}^0)
\]
is Fréchet differentiable, where $S_2(\mathcal{H}_{\P}^0)$ denotes the Banach space of symmetric continuous bilinear forms on $\mathcal{H}_{\P}^0$, endowed with the usual operator norm.
\end{lemma}

As a consequence of Lemma~\ref{lem:RKHS-igb}, the space $\mathcal{H}_{\P}^0$ equipped with the family of inner products $\langle \cdot, \cdot \rangle_{\mathcal{H}_{\P,F}}$ satisfies the axioms of a Hilbert manifold --- i.e.\ an infinite-dimensional Riemannian manifold --- in~\citet{Kli95}. Here the manifold structure is particularly simple, since the underlying Hilbert space $\mathcal{H}_{\P}^0$ is fixed, and only the inner product varies smoothly with~$F$.

\begin{definition}\label{def:Riemannian-manifold}
We denote by $\mathbb{H}_{\P}$ the \emph{Hilbert manifold} consisting of the set of functions $\mathcal{H}_{\P}^0$ equipped with the family of inner products
$\langle \cdot, \cdot \rangle_{\mathcal{H}_{\P,F}}$, $F \in \mathcal{H}_{\P}^0$.
\end{definition}

\subsection{Gradient flow associated with infinitesimal gradient boosting}

Using the Hilbert manifold $ \mathbb{H}_\P $ and the results on the gradient of the loss functional from Section~\ref{sec:gradient-properties}, we can now reinterpret infinitesimal gradient boosting as a gradient flow. Indeed, Theorem~\ref{thm:grf-as-gradient} shows that the right-hand side of the differential equation~\eqref{eq:igb-ODE} can be written as
\[
\bar T(\cdot;\P,F, \mu_{\mathrm{P},F}) = -\,\nabla^{\mathcal{H}_{\P,F}} R_{\P}(F),
\]
where the gradient is taken with respect to the inner product in the Hilbert space $ \mathcal{H}_{\P,F} $. More precisely, the Fréchet derivative $ \rmd_F R_{\P} $ at point $ F $ satisfies
\[
(\rmd_F R_{\P})(G)
= \langle \nabla^{\mathcal{H}_{\P,F}} R_{\P}(F),\, G \rangle_{\mathcal{H}_{\P,F}},
\qquad \text{for all } G \in \mathcal{H}_{\P,F}.
\]
By construction of the Hilbert manifold $ \mathbb{H}_\P $, this pointwise gradient coincides with the Riemannian gradient on the manifold:
\[
\nabla^{\mathcal{H}_{\P,F}} R_{\P}(F)
= \nabla^{\mathbb{H}_\P} R_{\P}(F).
\]
Thus, the differential equation~\eqref{eq:igb-ODE} can be rewritten as
\begin{equation}\label{eq:gradient-flow}
\frac{\mathrm{d} F_t}{\mathrm{d} t}
= -\,\nabla^{\mathbb{H}_\P} R_{\P}(F_t),
\qquad t \ge 0,
\end{equation}
which we recognize as the (negative) gradient flow of the functional $ R_{\P} $.

\begin{theorem}\label{thm:igb-gradient-flow}
The infinitesimal gradient boosting process $ (\hat F_t)_{t \ge 0} $, introduced in Definition~\ref{def:igb}, is the unique solution to the gradient flow equation~\eqref{eq:gradient-flow} on the Hilbert manifold $ \mathbb{H}_\P $, with initial condition~\eqref{eq:igb-initialization}.
\end{theorem}

In light of the above discussion, the proof of Theorem~\ref{thm:igb-gradient-flow} reduces to establishing the existence and uniqueness of solutions to the gradient flow equation~\eqref{eq:gradient-flow}. Once this is done, the resulting solution must coincide with the infinitesimal gradient boosting process $ (\hat F_t)_{t \ge 0} $. The following lemma shows that the vector field driving the gradient flow is locally Lipschitz with linear growth, which guarantees global existence and uniqueness of solutions for all time $ t \ge 0 $. 

\begin{lemma}\label{lem:gradient-lipschitz}
The gradient vector field
$F \longmapsto \nabla^{\mathbb{H}_\P} R_{\P}(F)$
is locally Lipschitz on $ \mathcal{H}_{\P}^0 $; that is, for all $F\in \mathcal{H}_{\P}^0$, for all $\delta > 0$, there exists a constant $ C_\delta > 0 $ such that
\[
\big\|
\nabla^{\mathbb{H}_\P} R_{\P}(F_1)
- \nabla^{\mathbb{H}_\P} R_{\P}(F_2)
\big\|_{\mathcal{H}_{\P}^0}
\le
C_\delta\, \|F_1 - F_2\|_{\mathcal{H}_{\P}^0},
\]
for all $F_1, F_2 \in \mathcal{H}_{\P}^0$ with $\|F_i-F\|_{\mathcal{H}_{\P}^0} \le \delta$. 
Moreover, the vector field satisfies a linear growth condition: there exists $D > 0$ such that
\[
\big\|\nabla^{\mathbb{H}_\P} R_{\P}(F)\big\|_{\mathcal{H}_{\P}^0}
\le D\,(\|F\|_{\mathcal{H}_{\P}^0} + 1),
\qquad  F \in \mathcal{H}_{\P}^0.
\]
\end{lemma}

By applying the Cauchy--Lipschitz (Picard--Lindelöf) theorem in Hilbert spaces 
(see, e.g., \citet[Chapter~9]{Brezis2011}), 
the local Lipschitz continuity of the gradient vector field ensures that the gradient flow equation~\eqref{eq:gradient-flow} admits a unique maximal solution defined on $[0, T_{\max})$. 
The linear growth condition guarantees that no finite-time blow-up occurs, so that $ T_{\max} = +\infty $. 
In other words, the solution is globally defined for all $ t \ge 0 $.

\smallskip
To conclude this section, we briefly discuss the large-sample behavior of infinitesimal gradient boosting. 
Consider an i.i.d.\ sequence of observations $ Z_i = (X_i, Y_i) $, $ i \ge 1 $, drawn from a distribution $ \P^* $. 
The empirical distribution $ \P_n $, constructed from a sample of size $ n $, converges almost surely (in distribution) to $ \P^* $. 
As a result, one may expect the infinitesimal gradient boosting process driven by the gradient flow of the empirical risk $ R_{\P_n} $ to converge almost surely to the process driven by the gradient flow of the population risk $ R_{\P^*} $. 
Such convergence has been established in \citet{DD24a}, in a weaker topology, for processes taking values in $ \mathcal{C}^0([0,1]^p, \mathbb{R}) $. 

In view of the results established in this paper, it is plausible to strengthen this convergence in a suitable reproducing kernel Hilbert space---even under norms stronger than that of $ \mathcal{H}_{\P^*}^0 $. 
However, these refined results pertain to a forthcoming project devoted to the study of the fluctuations of infinitesimal gradient boosting, in the sense of a functional central limit theorem. 
We therefore do not pursue this direction further in the present work.

\section{Illustration and discussion}\label{sec:illustration}

Random Forest kernels could be leveraged in various applications, such as support vector machines, kernel ridge regression, principal components analysis, to name just a few. In this section, we illustrate the application of Random Forest kernels in practical settings through kernel Principal Component Analysis (PCA) and investigate briefly the empirical behavior of the Geometric Variable Importance (GVI) introduced in \eqref{eq:VI}.

\subsection{Random Forest Kernel PCA}

We investigated kernel Principal Component Analysis (kPCA) \citep{scholkopf1998nonlinear} on several real classical benchmark datasets and compared the performance of different kernel functions. As baselines, we included classical (linear) PCA and the standard Gaussian radial basis function (RBF) kernel, defined by
\begin{equation*}
k_{\mathrm{rbf}}(x, x') = \exp(-\gamma \|x - x'\|^2).
\end{equation*}
In our simulations, we used the simple choice $\gamma = 1/p$, which reduces to \texttt{scikit-learn}'s default parameter in our settings. In addition, we considered the two Random Forest based kernel functions studied in this article, namely $k_0$ and $k$,  defined respectively in \eqref{eq:def-k0} and \eqref{eq:def-k}. Given that we shall compare $k$ and $k_0$, in this section, $k$ will be denoted by $k_\P$ for better readability. These kernels were computed using four Random Forest algorithms: Breiman's Random Forest with bootstrap resampling and no restriction on tree depth (\texttt{BRF}); Extra-Trees with maximum depth $d=5$ (\texttt{ET5}); Extra-Trees with unrestricted depth (\texttt{ET}); and the Uniform Random Forest (\texttt{Unif}). For the Uniform Random Forest algorithm, we used the Extra-Trees implementation with a single randomly chosen feature per split. This yields highly randomized trees whose partitions closely approximate those of a uniform random forest. The only distinction is that Extra-Trees may reject a random split if it provides no impurity decrease, making the method almost, but not strictly, uniform. \\ This simulation design allowed us to assess not only the differences between $k_0$ and $k$, but also how their behavior varied across algorithms.

In all simulations, we set the hyperparameters of both the Random Forest models and the RBF kernel to standard default values. We wanted to evaluate how well the kernels perform when hyperparameters are not carefully tuned, reflecting the performance that users can typically expect without extensive hyperparameter selection.\\

Our goal is to study the behavior of kernel PCA in settings where a potentially high-dimensional set of covariates is reduced to a low-dimensional representation that remains informative about an outcome of interest. This is therefore not a completely unsupervised setting, although we note that random forest kernels can also be used in the unsupervised setting by creating random partitions independent of the outcome, say with uniform forests. We considered both binary and continuous outcomes, using about ten publicly available benchmark datasets for each task.

All datasets were pre-processed consistently across methods: (i) categorical variables were one-hot encoded; (ii) missing values were imputed by the mean for numeric variables and the mode for categorical variables; and (iii) each feature was centered and scaled. Each dataset was then randomly split into a training set (70\%) and a test set (30\%). Although a more thorough imputation procedure could have been employed, the simulation was not designed to address this aspect in detail.
Dimensionality reduction, via PCA or kernel PCA, was performed on the training set only, and all evaluation metrics were computed on the test set to assess generalization performance.

\subsubsection{Random Forest Kernel PCA with binary outcome}
Experiments were conducted on eleven publicly available binary classification datasets. A complete list, along with a description of the main features of each dataset, is provided in Appendix~\ref{appendix:simu}. Several datasets were originally multiclass and were converted into binary classification problems.

\begin{figure}[htbp]
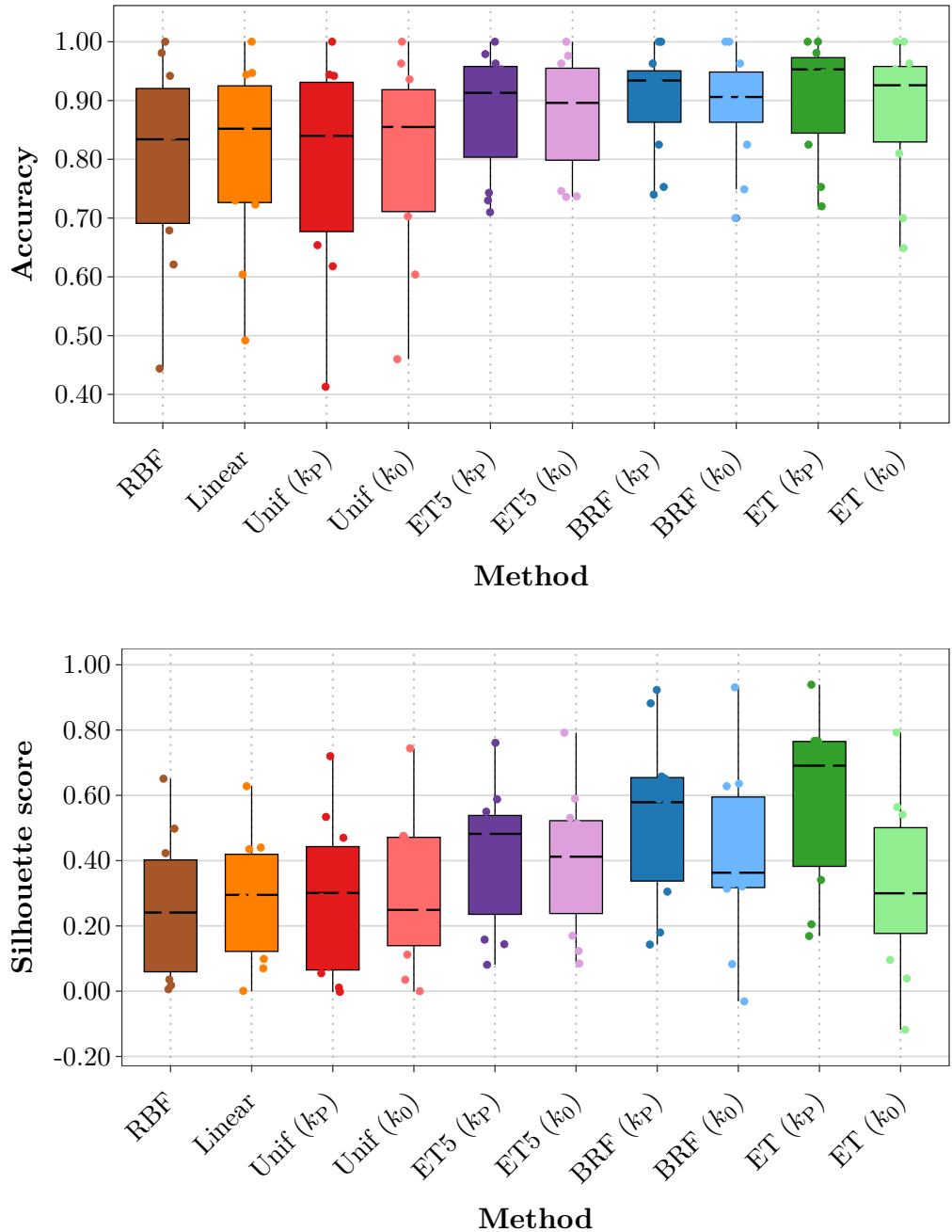

  \centering
  \begin{subfigure}{1\linewidth}
    \centering
    \resizebox{0.9\linewidth}{!}{\input{accuracy_boxplot.tex}}
    \label{fig:accuracy}
  \end{subfigure}
\vspace{2mm}

  \begin{subfigure}{1\linewidth}
    \centering
    \resizebox{0.9\linewidth}{!}{\input{silhouette_boxplot.tex}}    \label{fig:silhouette}
  \end{subfigure}
  \caption{Summary of the simulation results for the PCA application. Subfigure (a) reports the classification accuracy, while subfigure (b) reports the silhouette scores.} \label{graph_classif}
\end{figure}

We compared the six dimensionality reduction methods using three criteria: (i) the quality of two-dimensional visualizations obtained by projecting the test data onto the first two components; (ii) the average silhouette score \citep{rousseeuw1987silhouettes} computed on the test data; and (iii) the classification accuracy of a linear logistic regression model trained on the projected training data and evaluated on the test data. The results are summarized in Figure~\ref{graph_classif}, where each boxplot shows the distribution of results for a given kernel method.\\

\begin{sidewaysfigure}[htbp]
  \centering
  \rotatebox{180}{
    \begin{minipage}{\textwidth}
      \centering
      \resizebox{\textwidth}{!}{%
        \includegraphics[page=1]{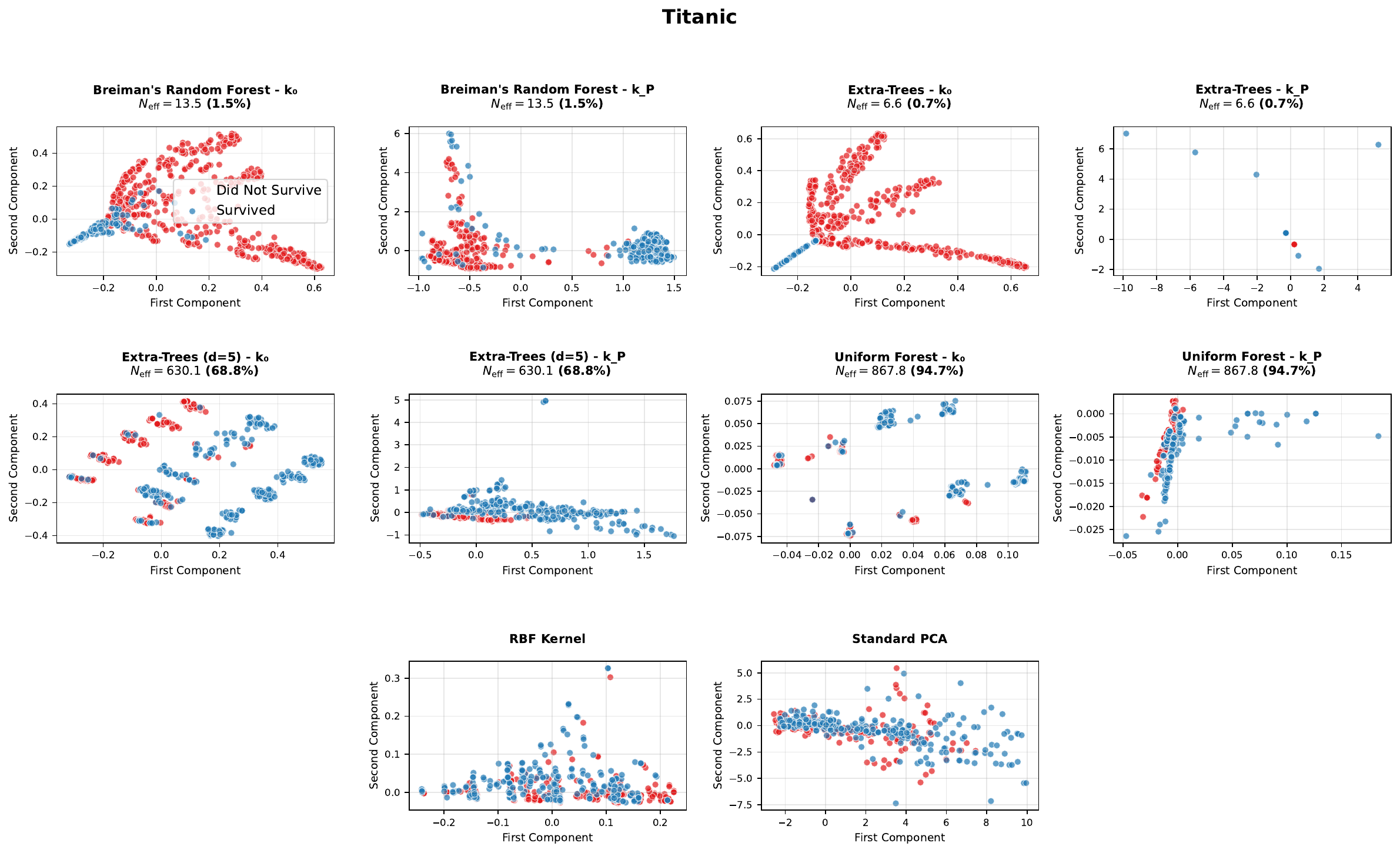}%
      }
      \caption{Visualization of the data projected onto the first two principal components of the KPCA for the Titanic dataset.}
      \label{fig:titanic}
    \end{minipage}%
  }
\end{sidewaysfigure}

Across the eleven benchmark datasets, the Random Forest kernels based on Breiman’s Random Forest (BRF), Extra–Trees (ET), and Extra–Trees with limited depth (ET5) achieved the highest average classification accuracy and silhouette scores.  
These three kernels consistently outperformed standard PCA, the Gaussian RBF kernel, and the uniform forest kernels.  
Within each forest family, the weighted kernel $k_{P}$ generally gave the best results: for BRF and ET, $k_{P}$ provided both higher mean accuracy and significantly higher silhouette scores than the corresponding unweighted kernel $k_{0}$.  
The advantage of $k_{P}$ was most pronounced for the fully data adaptive forests (BRF and ET), whereas for the uniform forest, whose partitions are independent of the data, the difference between $k_{P}$ and $k_{0}$ was negligible or slightly in favour of $k_{0}$.  
Restricting tree depth (ET5) led to a moderate decrease in both accuracy and silhouette compared with unrestricted Extra–Trees, showing that limiting partition complexity reduces the expressiveness of the kernel.  
At the same time, for certain datasets, the Extra–Trees embeddings occasionally produced almost degenerate two-dimensional representations, where nearly all points were projected to an extremely small region of the plane (as illustrated by Figure \ref{fig:titanic}), even though the downstream classification accuracy remained high.

Across datasets, the effective sample size $N_{\mathrm{eff}}$ (reported as a percentage of the training size) varied significantly by construction: Uniform forest had median $38.4\%$ (range $19.8\%$–$94.7\%$), Extra–Trees had median $5.2\%$ ($0.3\%$–$44.2\%$), Extra–Trees with depth $5$ had median $27.8\%$ ($16.2\%$–$68.8\%$), and Breiman’s Random Forest had median $9.6\%$ ($1.5\%$–$49.1\%$). These results indicate no universal target level for $N_{\mathrm{eff}}$: performance tended to be best when the effective number of regions balanced the richness of fine partitions with the stability of coarser ones, in a manner that depended on both the forest construction and the dataset. 

\subsubsection{Random Forest Kernel PCA with continuous outcome}
Experiments were conducted on eight publicly available regression datasets. A complete list, together with a description of the main features of each dataset, is provided in Appendix~\ref{appendix:simu}. 

\begin{figure}[htbp]
 \centering
    \resizebox{0.9\linewidth}{!}{\input{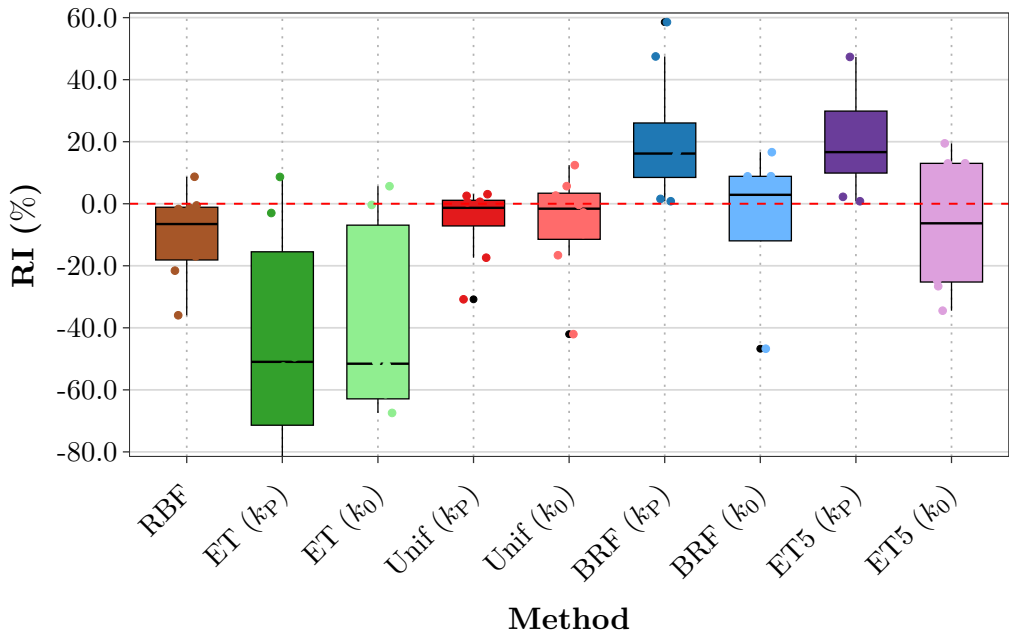}}
    \caption{Relative improvement of MSE versus standard PCA.}
    \label{fig:pca_reg}
\end{figure}

For continuous outcomes, we used two evaluation criteria: (i) the quality of two-dimensional visualizations obtained by projecting the test data onto the first two components; and (ii) the mean squared error (MSE) of a linear regression model fitted on the first two components of KPCA and evaluated on the test data. Because MSE values were on different scales across datasets, we adopted a relative measure. Taking Principal Component Regression (PCR) as the baseline, we computed the relative improvement (RI) of each kernel $k$ as 
\[
\text{RI}(k) = 100 \times \frac{\text{MSE}(k) - \text{MSE}(k_{\mathrm{pcr}})}{\text{MSE}(k_{\mathrm{pcr}})}.
\] 
This measure represents the percentage improvement (or degradation) of kernel $k$ relative to PCR. The results are summarized in Figure~\ref{fig:pca_reg}, with one boxplot per method.

Across the eight regression datasets, the Random Forest kernels based on Breiman’s Random Forest (BRF) and Extra Trees with limited depth (ET5) achieved the best overall performance.  
In particular, $k_{P}$ for BRF attained the lowest MSE on four datasets, and $k_{P}$ for ET5 was best on three; the uniform forest was best on \textit{abalone}.  
Both BRF $k_{P}$ and ET5 $k_{P}$ outperformed the Gaussian RBF kernel and standard PCA on all eight datasets, whereas the unrestricted Extra–Trees variants performed worst on average under the default settings; within constructions, $k_{P}$ improved upon $k_{0}$ for BRF and ET5 on all datasets, was mixed for the uniform forest, and was generally inferior to $k_{0}$ for unrestricted Extra–Trees.

\begin{sidewaysfigure}[htbp]
  \centering
  \rotatebox{180}{
    \begin{minipage}{\textwidth}
      \centering
      \resizebox{\textwidth}{!}{%
        \includegraphics[page=1]{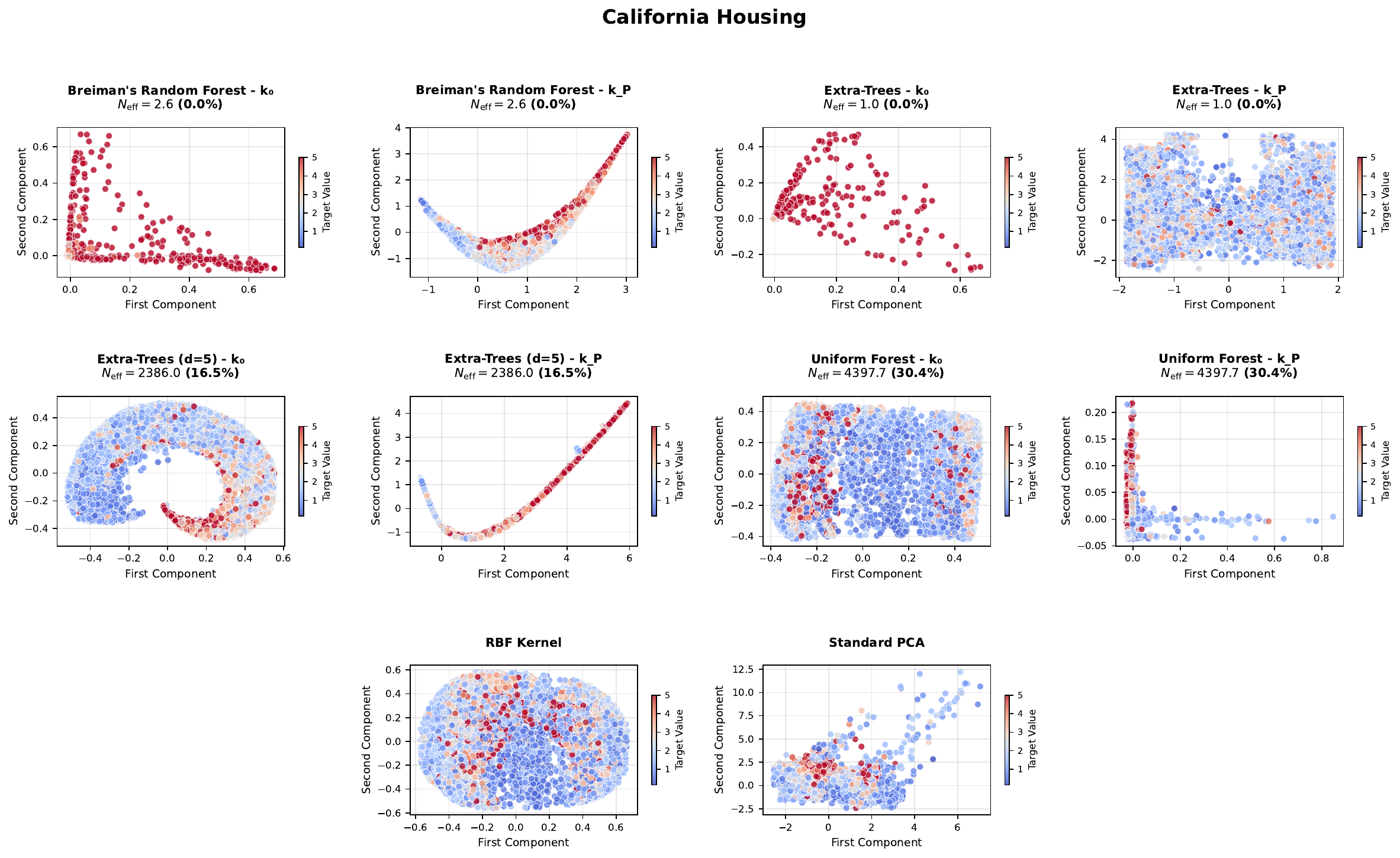}%
      }
      \caption{Visualization of the data projected onto the first two principal components of the KPCA for the California Housing dataset.}
      \label{fig:calif}
    \end{minipage}%
  }
\end{sidewaysfigure}

This pattern contrasts with classification, where unrestricted Extra–Trees performed strongly, and is consistent with the view that regression typically requires stronger regularization: bootstrap sampling in BRF acts as a natural regularizer that stabilizes the induced kernels. To further investigate this phenomenon, we reran the simulations and added bootstrap to Extra–Trees (with all other parameters unchanged): this led to marked improvements in both error and visualization. Additionally, the comparatively weak performance of the untuned RBF kernel highlights its sensitivity to $\gamma$, which we deliberately set to a default value to replicate the no–tuning setup used for all methods. Figure \ref{fig:calif} shows the visualization for the California Housing dataset.

\subsection{An investigation of the GVI criterion for Random Forest interpretability}

We compared the proposed GVI criterion against two standard variable importance measures: Mean Decrease in Impurity (MDI), which is the default importance computed by tree-based methods, and Mean Decrease in Accuracy (MDA), also known as permutation importance. Our simulation study aims to assess whether GVI provides complementary information to these established metrics, and to identify settings where each approach excels or struggles. We focus particularly on scenarios involving correlated features, redundancy, and weak signals, which are common challenges in practice.

\paragraph{Data-generating scenarios.}
We consider ten scenarios with sample size $n=500$ and dimensionality $p=20$ (except one higher-dimensional case with $p=50$). Each scenario defines a signal set $S \subset \{1,\dots,p\}$ of features that truly affect the outcome. The scenarios span a range of realistic challenges: additive models with various correlation structures (S1--S2), pure interaction effects without main effects (S3), categorical noise variables (S4), localized effects (S5), redundant features (S6), high-dimensional sparse signals (S7), weak signals with low signal-to-noise ratio (S8), correlated noise acting as "confounding correlations" (S9), and context-dependent effects (S10). A detailed description of each scenario is given in Appendix \ref{appendix:simu}.

\paragraph{Simulation protocol.}
For each scenario, we generate $R=100$ independent replicates. At each replicate, we fitted an Extra-Trees ensemble with $M = 500$ trees, $\texttt{max\_features}=\sqrt{p}$, and $\texttt{min\_samples\_leaf}=5$. We then compute three importance vectors:
\begin{itemize}
  \item $\mathrm{GVI}$: For each feature $j$, we compute the geometric variable importance as defined in \eqref{eq:VI}. This ratio measures the proportion of the variance in $X_j$ that is preserved by the kernel-smoothing operator $W$.
  
  \item MDI: The mean decrease in impurity, extracted directly from the fitted ExtraTrees model via \texttt{feature\_importances\_}. This is the sum of weighted impurity decreases for all splits involving feature $j$.
  
  \item MDA: The mean decrease in accuracy, computed via permutation importance with 5 random permutations of each feature on the training data. This measures the drop in model performance when feature $j$ is randomly shuffled.
\end{itemize}

\paragraph{Evaluation criteria.}
For each method and each replicate, we computed three metrics to assess how well the importance scores distinguish signal from noise:
\begin{itemize}
  \item Precision\_K: Let $K = |S|$ be the number of true signals. We rank features by importance and compute the fraction of the top-$K$ features that belong to $S$. This directly measures whether the method successfully identifies the right features.
  
  \item Separation: We compute the difference between the mean importance score of signal features and the mean importance score of noise features:
  \[
  \text{Separation} = \frac{1}{|S|}\sum_{j \in S} s_j - \frac{1}{p - |S|}\sum_{j \notin S} s_j,
  \]
  where $s_j$ denotes the importance score of feature $j$. Larger values indicate stronger discrimination between signal and noise.
  
  \item Spearman vs.\ MDA: We compute the Spearman rank correlation between the importance scores from the method under evaluation and those from MDA. This measures the agreement between the method's ranking and the ranking based on predictive performance. For MDA itself, this correlation is trivially 1; for GVI and MDI, it reveals how closely their geometric or split-based rankings align with predictive importance.
\end{itemize}
We report the mean of each metric across the $R=100$ replicates, along with the mean computation time and the average standard deviation of importance scores across features (a measure of between-replicate variability). All reported values are rounded to two decimal places. The results are presented in Table \ref{Table1}.

\begin{table}[ht!]
\centering
\caption{Comparison of variable-importance methods across simulation scenarios ($n=500$). 
Each entry reports the average over 50 replicates. Bolded values indicate the best performance per scenario 
for Precision\_K and Separation.}
\resizebox{\textwidth}{!}{
\begin{tabular}{llccccr}
\toprule
\textbf{Scenario} & \textbf{Method} & \textbf{Precision\_K} & \textbf{Separation} & \textbf{Spearman vs MDA} & \textbf{Time (s)} \\
\midrule
\multirow{3}{*}{S1: Additive} 
& GVI & \textbf{1.00} & 0.20 & 0.85 & 0.01 \\
& MDI & \textbf{1.00} & \textbf{0.26} & 0.92 & \textbf{0.00} \\
& MDA & \textbf{1.00} & 0.22 & 1.00 & 2.11 \\
\midrule
\multirow{3}{*}{S2: Collinear signals} 
& GVI & \textbf{1.00} & \textbf{0.43} & 0.92 & 0.03 \\
& MDI & \textbf{1.00} & 0.41 & 0.88 & \textbf{0.00} \\
& MDA & \textbf{1.00} & 0.33 & 1.00 & 2.10 \\
\midrule
\multirow{3}{*}{S3: XOR} 
& GVI & 0.24 & 0.02 & 0.93 & 0.02 \\
& MDI & \textbf{1.00} & 0.06 & 0.92 & \textbf{0.00} \\
& MDA & 0.96 & \textbf{0.07} & 1.00 & 2.15 \\
\midrule
\multirow{3}{*}{S4: Categorical distractor} 
& GVI & \textbf{1.00} & 0.15 & 0.95 & 0.02 \\
& MDI & \textbf{1.00} & \textbf{0.33} & 0.91 & \textbf{0.00} \\
& MDA & \textbf{1.00} & 0.28 & 1.00 & 2.14 \\
\midrule
\multirow{3}{*}{S5: Local relevance} 
& GVI & \textbf{1.00} & 0.25 & 0.94 & 0.02 \\
& MDI & \textbf{1.00} & 0.62 & 0.90 & \textbf{0.00} \\
& MDA & \textbf{1.00} & \textbf{0.63} & 1.00 & 2.18 \\
\midrule
\multirow{3}{*}{S6: Redundant noises} 
& GVI & 0.50 & \textbf{0.28} & 0.90 & 0.03 \\
& MDI & 0.50 & 0.19 & 0.89 & \textbf{0.00} \\
& MDA & \textbf{0.56} & 0.15 & 1.00 & 2.13 \\
\midrule
\multirow{3}{*}{S7: High-dim ($p=50$)} 
& GVI & 0.60 & \textbf{0.16} & 0.93 & 0.03 \\
& MDI & 0.61 & 0.06 & 0.94 & \textbf{0.00} \\
& MDA & \textbf{0.65} & 0.07 & 1.00 & 5.29 \\
\midrule
\multirow{3}{*}{S8: Weak signal} 
& GVI & 0.90 & \textbf{0.07} & 0.95 & 0.01 \\
& MDI & \textbf{0.91} & 0.06 & 0.92 & \textbf{0.00} \\
& MDA & \textbf{0.91} & 0.05 & 1.00 & 2.05 \\
\midrule
\multirow{3}{*}{S9: Correlated noise} 
& GVI & 0.37 & \textbf{0.29} & 0.93 & 0.01 \\
& MDI & \textbf{0.38} & 0.08 & 0.95 & \textbf{0.00} \\
& MDA & 0.52 & 0.07 & 1.00 & 2.02 \\
\midrule
\multirow{3}{*}{S10: Context-dependent} 
& GVI & 0.98 & 0.18 & 0.96 & 0.01 \\
& MDI & \textbf{1.00} & 0.19 & 0.91 & \textbf{0.00} \\
& MDA & \textbf{1.00} & \textbf{0.21} & 1.00 & 2.03 \\
\bottomrule
\end{tabular}
}
\label{Table1}
\end{table}

Across all scenarios, the proposed $\mathrm{GVI}$ measure performed closely to MDA while being orders of magnitude faster to compute. In most cases, $\mathrm{GVI}$ was more correlated to MDA than MDI was. It consistently identified the relevant features (high Precision\_K) in smooth or correlated settings and achieved separations comparable to or often higher than both MDI and MDA, particularly when signals are correlated or partially redundant. These results indicate that $\mathrm{GVI}$ captures stable and meaningful variations in feature relevance, with small variability across replicates. In more complex or highly nonlinear situations, such as the XOR and local relevance scenarios, $\mathrm{GVI}$ showed slightly lower sensitivity to interaction effects or sharply localized signals, which reflects its reliance on the smooth geometric structure induced by the ensemble. Thus, what is not fully captured by the ensemble will not typically perform high in terms of $\mathrm{GVI}$. Overall, $\mathrm{GVI}$ seems to provide a reliable and efficient complement to MDI and MDA, reproducing their main feature-ranking patterns at almost no additional computational cost.

\section{Proofs related to Section~\ref{sec:RKHS}}\label{sec:proofs-1}

\begin{proof}[Proof of Proposition~\ref{prop:unif-cv-rf}] 
The proof uses material from \cite[Section~3]{DD24} and the introduction of the space of tree functions that is not used in the remainder of this paper. For this reason and for the sake of brevity,  we propose only a sketch of proof here and refer to \cite{DD24} for more details. The space $\mathbb{T}$ of tree functions on $[0,1]^p$ is defined as the Banach space of functions
\[
T(z)=\rho([0,z]),\quad z\in[0,1]^p, \quad \text{for some  finite signed Borel measure  $\rho$ on $[0,1]^p$,}
\]
endowed with the norm $\|T\|=\|\rho\|_{\mathrm{TV}}$, where  $\|\cdot\|_{\mathrm{TV}}$ denotes the total variation norm. Similarly as in \citet[Proposition~3.2]{DD24}, one can prove that for all $M\geq 1$, the finite forest
\[
\bar T_M(z;\P,\Pi_1,\ldots,\Pi_M)=\frac{1}{M}\sum_{m=1}^M T(z;\P,\Pi_m).
\]
is a tree function, whose associated measure $\bar \rho_M=\frac{1}{M}\sum_{m=1}^M \rho_{m}$
is a sum of independent and identically distributed terms  with total variation bounded by $2^p M|\Pi_m|$, $M$ denoting the absolute bound for the $Y_i$'s or for the regression function. By Assumption~\eqref{eq:integrability-pi}, $|\Pi|$ is integrable and this implies that the positive part $\bar \rho^+_M$ and negative part $\bar\rho^-_M$ of $\bar\rho_M$ converge  weakly to some limit $\bar\rho^+$ and $\bar\rho^-$ as $M\to\infty$ almost surely . An argument similar to the Glivenko-Cantelli theorem then implies the almost sure convergence of $\bar\rho_M^\pm(z)\to \bar\rho^\pm(z)$ uniformly on $[0,1]^p$. The uniform convergence of $\bar T_M(z)=\bar\rho_M^+(z)-\bar\rho_M^-(z)$ on $[0,1]^p$ follows with limit $\bar T(z)$  identified by the pointwise convergence stated in Equation~\eqref{eq:cv-rf}.
\end{proof}

\begin{proof}[Proof of Proposition~\ref{prop:k_0-k}]
    The symmetry of $k_0$ is clear from the definition~\eqref{eq:def-k0}  and semi-definite positivity follows from
    \[
    \sum_{1\leq l,l'\leq L} a_la_{l'}k_0(z_l,z_{l'})=\int_\mathcal{A} \Bigg( \sum_{l=1}^La_l\1_A(z_l)\Bigg)^2\,\nu(\rmd A)\geq 0.
    \]
    The normalization $k(z,z)=1$ is a consequence of Equation~\eqref{eq:normalisation-nu}, or can equivalently be seen from Equation~\eqref{eq:def-k0-bis}, since $z\stackrel{\Pi}\longleftrightarrow z$ is always true.

    Next we prove that $D$ has $\P$-probability $1$ and that $k$ is a kernel on $D$. The symmetry of $k$ is clear from the definition~\eqref{eq:def-k0}.
    Then, 
    \[ k(z,z')=\int_{\mathcal{A}}\alpha_\P^2(A)\1_A(z)\1_A(z')\,\nu(\rmd A).
    \]
    An application of the Fubini--Tonelli theorem yields
    \begin{align*}
    \int_{[0,1]^p} k(z,z)\,\P(\rmd z) &= \int_{[0,1]^p}\int_{\mathcal{A}} \alpha_\P^2(A)\1_A(z)\,\nu(\rmd A)\P(\rmd z)\\
    &= \int_{\mathcal{A}} \alpha_\P^2(A)\P(A)\,\nu(\rmd A)\\
    &=  \int_{\mathcal{A}} \1_{\{\P(A)>0\}}\,\nu(\rmd A)\\
    &\leq \nu(\mathcal{A}).   
    \end{align*}
    The announced upper bound on the integral then follows from the relation $\nu(\mathcal{A})=\mathbb{E}_\Pi[|\Pi|]$, which is hence finite by assumption~\eqref{eq:integrability-pi}. As a consequence, $k(z,z)$ is finite $\P(\rmd z)$-almost everywhere and $\P(D)=1$. The semi-definite positivity of $k$ on $D$ follows from
    \[
    \sum_{1\leq l,l'\leq L} a_la_{l'}k_0(z_l,z_{l'})=\int_{\mathcal{A}}\Bigg(\sum_{l=1}^La_l\alpha_{\P}(A)\1_A(z_l)\Bigg)^2\,\nu(\rmd A)\geq 0.
    \]
    Finally, for $z\in[0,1]^p$, we have
    \begin{align*}
    \int_{[0,1]^p}k(z,x)\,\P(\rmd x)&= \int_{[0,1]^p} \int_{\mathcal{A}}\alpha_\P(A)\1_A(z)\1_A(x)\, \P(\rmd x)\nu(\rmd A)\\
    &= \int_{\mathcal{A}}\alpha_\P(A)\P(A)\1_A(z)\, \nu(\rmd A) \\
    &= \int_{\mathcal{A}}\1_{\{\P(A)>0\}}\1_A(z) \nu(\rmd A)\\
    &\leq \int_{\mathcal{A}}\1_A(z) \nu(\rmd A)=1,        
    \end{align*}
    where the last equality was stated in~\eqref{eq:normalisation-nu}.\end{proof}

\begin{proof}[Proof of Proposition~\ref{prop:boundedness}]
It suffices to check that $k$ is bounded on the diagonal $\{(z,z): z\in[0,1]^p\}$. Indeed, if $D=[0,1]^p$, then the Cauchy–Schwarz inequality gives
\[
|k(z,z')|\leq \sqrt{k(z,z)}\sqrt{k(z',z')} \leq \sup_{z\in[0,1]^p} k(z,z),
\]
which implies that $k$ is bounded on $[0,1]^p\times[0,1]^p$.
By definition \eqref{eq:def-k},
\[
\sup_{z\in[0,1]^p} k(z,z) \leq \mathbb{E}_\Pi\Bigg[\sup_{z\in[0,1]^p} \sum_{A\in\Pi}\alpha_\P^2(A)\1_A(z)\Bigg] 
= \mathbb{E}_\Pi\Big[\max_{A\in\Pi} \Big(\alpha_\P^2(A)\Big)\Big].
\]
Since $\mu$ is the distribution of $\pi$, the right-hand side is equal to
\[
\int_{\mathfrak{P}}\max_{A\in\pi} \Big(\alpha_\P^2(A)\Big)\,\mu(\rmd\pi),
\]
which serves as an upper bound for $k$.  
For a finite sample with empirical distribution $\P=\P_n$, any nonempty component $A$ of $\Pi$ contains at least one sample point, so $\P_n(A)\geq 1/n$. Hence
\[
\max_{A\in\Pi} \Big(\alpha_{\P_n}^2(A)\Big) \leq n \quad \text{a.s.}
\]
Taking expectations shows that $n$ is an upper bound for $k$.  
Finally, note that
\[
\mathbb{E}_\Pi\Big[\max_{A\in\Pi} \Big(\alpha_\P^2(A)\Big)\Big]
\leq \mathbb{E}_\Pi\Big[\sum_{A\in\Pi} \alpha_\P^2(A)\Big],
\]
which yields
\[
\int_{\mathfrak{P}}\max_{A\in\pi} \Big(\alpha_\P^2(A)\Big)\,\mu(\rmd\pi)
\leq \int_{\mathcal{A}}\alpha_\P^2(A)\,\nu(\rmd A),
\]
as claimed.
\end{proof}

\begin{proof}[Proof of Proposition~\ref{prop:continuity}]
Fix $z,z'\in [0,1]^p$.  
Assumption~\eqref{eq:assumption-2} implies that for $\nu$-almost every $A$, we have $z,z'\notin \partial A$, so that $\1_A(\cdot)$ is continuous at both $z$ and $z'$.  
Therefore, for any sequence $(z_n,z'_n)\to (z,z')$, we obtain
\[
   \1_{A}(z_n)\1_{A}(z'_n) \;\longrightarrow\;  \1_{A}(z)\1_{A}(z') \qquad \nu\text{-a.e.}
\]
Since these terms are bounded by $1$ and $\nu$ is a finite measure, an application of the dominated convergence theorem yields the continuity of $k_0$.  

The reasoning for $k$ is similar:  
\[
  \sum_{A\in\pi}\alpha_\P^2(A)\,\1_{A}(z_n)\1_{A}(z'_n) 
  \;\longrightarrow\;     
  \sum_{A\in\pi}\alpha_\P^2(A)\,\1_{A}(z)\1_{A}(z') 
  \qquad \mu\text{-a.e.},
\]
where $\alpha_\P(A)$ is defined as in the proof of Proposition~\ref{prop:boundedness}.  
These terms are bounded by $\max_{A\in \pi} \bigl(\tfrac{1}{\P(A)}\wedge 1\bigr)$, which is $\pi$-integrable by Assumption~\eqref{eq:assumption-1}.  
Another application of the dominated convergence theorem then gives the continuity of $k$.  

Finally, by the definition of $\partial A$ and of the measures $\rho_j$, we have
\[
\int_{\mathcal{A}}\1_{\{z\in\partial A\}}\,\nu(\rmd A)\;\leq\; \sum_{j=1}^p \rho_j(\{z_j\}).
\]
This shows that Assumption~\eqref{eq:assumption-2} holds whenever the measures $\rho_j$ are atomless.
\end{proof}
\begin{proof}[Proof of Proposition~\ref{prop:propRKHS}]
    \emph{Proof of (a).} We start with (i).  Using the reproducing property of $k$, for all $z\in D$ and $F\in \mathcal{H}_0$,
 \[
  \abs{F(z)} = \abs{\langle F, k_z\rangle_{\mathcal{H}_0}} \leq \norm{F}_{\mathcal{H}_0}\norm{k_z}_{\mathcal{H}_0},
 \]
 and note that $ \norm{k_z}^2_{\mathcal{H}_0}= \langle k_z, k_z\rangle_{\mathcal{H}_0} = k_0(z,z) =1$ where the last equality follows Proposition \ref{prop:k_0-k}. For (ii), under \eqref{eq:assumption-2}, $k_0$ is continuous which implies continuity of the functions in $\mathcal{H}_0$. For (iii), to see that the constant function belongs to both RKHSs, note  that by  \citet[Theorem~2.7.7]{HE15}, $\mathcal{H}_0$ consists of the functions admitting a representation of the form
\[
F(z)\;=\;\mathbb{E}_\Pi\Big[\sum_{A\in\Pi} G(A)\,\1_A(z)\Big],\quad z\in[0,1]^p,
\]
where $\mathbb{E}_\Pi[\sum_{A\in\Pi} G(A)^2]<\infty$. Choosing $G(A)\equiv 1$ yields $F(z)\equiv1$ because $\Pi$ is a partition of $[0,1]^p$.

\medskip

   \emph{Proof of (b).} The proof is similar to that of (a), only the constants change. For (i), as above, for any $F \in \mathcal{H}$, we have
 \[
  \abs{F(z)} \leqslant \norm{F}_{\mathcal{H}} \sqrt{k(z,z)}.\]
It follows that 
 \[
 \left(\int_{[0,1]^p} \abs{F(z)}^2\P(\rmd z)\right)^{1/2} \leqslant \norm{F}_{\mathcal{H}} \left(  \int_{[0,1]^p} k(z,z) \P(\rmd z) \right)^{1/2} =  \norm{F}_{\mathcal{H}} \int_{\mathcal{A}}\1_{\{\P(A)>0\}}\,\nu(\rmd A), \] where the equality follows by Proposition \ref{prop:k_0-k}. For (ii), a similar reasoning applies using Proposition \ref{prop:boundedness}. Finally, for (iii), continuity of functions in $\mathcal{H}$ follows from continuity of $k$ under our assumptions by Proposition \ref{prop:continuity}. It only remains to show that bounded sets in $\mathcal{H}$ are equicontinuous. Using the reproducing property of $k$, for any $F\in \mathcal{H}$ and $z,z'\in D$, we have
 \begin{align*}
  \abs{F(z)-F(z')} &= \abs{\langle F, k_z - k_{z'}\rangle_{\mathcal{H}}} \\
  &\leq \norm{F}_{\mathcal{H}} \norm{k_z - k_{z'}}_{\mathcal{H}} \\
  & = \norm{F}_{\mathcal{H}} \Big(k(z,z)+k(z',z')-2k(z,z')\Big)^{1/2}.
 \end{align*}
 Under Assumptions~\eqref{eq:assumption-1}-\eqref{eq:assumption-2}, by Proposition~\ref{prop:continuity}, the kernel $k$ is continuous on the compact space $[0,1]^p\times [0,1]^p$ and hence uniformly continuous.
 Therefore,  for $M,\delta>0$ and any $F\in \mathcal{H}$ with $\norm{F}_{\mathcal{H}}\leq M$, we have
 \[
   \sup_{\abs{z-z'}<\delta}\abs{F(z)-F(z')} \leq M \sup_{\abs{z-z'}<\delta}\Big(k(z,z)+k(z',z')-2k(z,z')\Big)^{1/2},
 \]
 which tends to $0$ as $\delta\to 0$.
 
\medskip

   \emph{Proof of (c).} By Theorem \citet[Theorem~2.7.11]{HE15}, the inclusion $\mathcal{H}_0 \subset \mathcal{H}$ and the norm inequality will follow from showing that $k-k_0$ is positive semi-definite. Since we assume $D=[0,1]^p$, the difference is finite and, from Equations~\eqref{eq:def-k0} and~\eqref{eq:def-k}, it can be written as
\[
(k-k_0)(z,z')
   = \mathbb{E}_\Pi\Bigg[\sum_{A\in \Pi}\beta_\P(A)\,\1_A(z)\1_A(z')\Bigg],
\]
where $\beta_\P(A)=1/\P(A)-1$ if $\P(A)>0$ and $\beta_\P(A)=0$ if $\P(A)=0$. Positive semi-definiteness then follows from the fact that all coefficients $\beta_\P(A)$ are nonnegative (by the same argument as in the proof of Proposition~\ref{prop:k_0-k}).
 \end{proof}

\color{black}

\begin{proof}[Proof of Proposition~\ref{prop:RKHS-structure}]
The proposition is a straightforward reformulation in our setting of \citet[Theorem~2.7.7]{HE15}. We only need to check that $A\mapsto g_\P(z,A)$ is an element of $L^2(\mathcal{A},\nu)$ for all $z\in D$. This is a direct consequence of the definition $D=\{z\in[0,1]^p\colon k(z,z)<\infty\}$ since
\[
\int_{\mathcal{A}}g_\P(z,A)^2 \nu(\rmd A)=k(z,z). 
\]
Hence $z\in D$ is equivalent to $g_\P(z,\cdot)\in L^2(\mathcal{A},\nu)$.
\end{proof}

\begin{proof}[Proof of Theorem~\ref{thm:RKHS-basics}] 
We first check that Equation~\eqref{eq:rf-as-kernel-integral} holds and compute, for all $z\in D$,
\begin{align*}
&\int_{[0,1]^p}yk(x,z)\, \P(\rmd x,\rmd y)\\
&=\int_{[0,1]^p\times\mathcal{A}}y \Big(\frac{\1_{\{\P(A)>0\}}}{\P(A)}+ \1_{\{\P(A)=0\}}\Big)\1_{A}(x)\1_A(z)\, \P(\rmd x, \rmd y)\nu(\rmd A) \\
&= \int_{\mathcal{A}}\frac{\P[y\1_A(x)]}{\P(A)}\1_A(z)\,\nu(\rmd A) \\
&= \bar T(z;\P,\mu).
\end{align*}
We have used successively Equation \eqref{eq:def-k} defining $k$, Fubini-Tonelli's theorem and Equation~\eqref{eq:def-bar-T} defining $\bar T$. 

By Equation~\eqref{eq:def-g},
\[
\bar T(z;\P,\mu)=\int_{\mathcal{A}}  \frac{\P[y\1_A(x)]}{\sqrt{\P(A)}}g_\P(A) \,\nu(\rmd A).
\]
It follows that a representation of the form~\eqref{eq:RKHS-representation} holds with 
\[
G(A)=\frac{\P[y\1_A(x)]}{\sqrt{\P(A)}},
\]
where $G\in L^2(\mathcal{A},\nu)$ because
\begin{align*}
\int_{\mathcal{A}} G(A)^2\,\nu(\rmd A)&= \int_{\mathcal{A}} \frac{\P[y\1_A(x)]^2}{\P(A)}\,\nu(\rmd A)\\
&\leq \int_{\mathcal{A}} \P[y^2\1_A(x)]\,\nu(\rmd A) \\
&=\int_{\mathfrak{P}}\P\Big[y^2\sum_{A\in\pi}\1_A(x)\Big] \,\mu(\rmd \pi)\\
&= \P[y^2]<\infty.
\end{align*}
The second line uses Cauchy-Schwarz inequality
\[
|\P[y\1_A(x)]|\leq \sqrt{\P[y^2\1_A(x)]}\sqrt{\P(A)}. 
\]
It follows that $\bar T\in\mathcal{H}$ with RKHS norm 
\[
\|\bar T\|_{\mathcal{H}}^2\leq \| G\|_{L^2(\nu,\mathcal{A})}^2\leq \P[y^2].
\]
One can be more precise and prove that the first inequality is in fact an equality. Observe that $G\in\dot{\mathcal{H}}$ because $G(A)=\P[yg_\P(A)]$, where each function $A\mapsto yg_\P(A)$ is in $\dot{\mathcal{H}}$ by Equation~\eqref{eq:def-dot-H}. As a consequence, $G$ is the unique representative of $\bar T(\cdot;\P,\mu)$ in $\dot{\mathcal{H}}$ and 
\[
\|\bar T\|_{\mathcal{H}}^2= \| G\|_{L^2(\nu,\mathcal{A})}^2.
\]
Then, we compute
\begin{align*}
\|\bar T(\cdot;\P,\mu)\|_{\mathcal{H}}^2&=  \int_{\mathcal{A}} \frac{\P[y\1_A(x)]^2}{\P(A)}\,\nu(\rmd A)\\
&=  \P[y^2] - \int_{\mathfrak{P}}\sum_{A\in\pi} \P\Bigg[ \Bigg(y-\frac{\P[y\1_A(x)]}{\P(A)}\Bigg)^2\1_A(x)\Bigg]\,\mu(\rmd\pi) \\
&= \P[y^2] - \int_{\mathfrak{P}} \P[(y-T(x;\P,\theta))^2]\,\mu(\rmd\pi),
\end{align*}
where the second equality follows from the decomposition of variance formula and the third one  from the definition of $T(x;\P,\theta)$. For the reader's convenience, we recall the decomposition of variance formula in our notation: for every partition $\pi$ of $[0,1]^p$ and every probability $\P$ on $[0,1]^p\times\mathbb{R}$ satisfying $\P[y^2]<\infty$,  
\[ 
\P[y^2]=\sum_{A\in\pi}\P(A)\bar y(A)^2+\sum_{A\in\pi}\P[(y-\bar y(A))^2\1_A(x)],
\]
with $\bar y(A)=\P[y\1_A(x)]/\P(A)$ the mean of $y$ on the group defined by the condition $x\in A$.
\end{proof}

\begin{proof}[Proof of Lemma~\ref{lem:identity}] By Proposition~\ref{prop:RKHS-structure}, $F(x)$ is equal to
\[
F(x)=\int_{\mathfrak{P}}\sum_{A\in\pi}\alpha_\P(A) \dot F(A)\1_A(x)\,\mu(\rmd\pi),
\]    
and, because $\pi$ is a partition of $[0,1]^p$, it holds
\[
y=\int_{\mathfrak{P}}\sum_{A\in\pi}y\1_A(x)\,\mu(\rmd\pi).
\]
We deduce
\[
\P[(y-F(x))^2]=\P\Big[\Big(\int_{\mathfrak{P}}\sum_{A\in\pi}\Big(y- \alpha_\P(A)\dot F(A)\Big)\1_A(x)\,\mu(\rmd\pi)\Big)^2 \Big].
\]
Note that the sum over $A$ must rather be seen as a case disjunction : only one term does not vanish according to which component of the partition $x$ belongs to. 

Using Huygens formula 
\[
\mathbb{E}[X]^2=\mathbb{E}[X^2]-\mathbb{E}[(X-\mathbb{E}(X))^2],
\]
the squared integral is equal to
\begin{align*}
&\Big(\int_{\mathfrak{P}}\sum_{A\in\pi}\Big(y- \alpha_\P(A) \dot F(A)\Big)\1_A(x)\,\mu(\rmd\pi)\Big)^2\\
=&\int_{\mathfrak{P}}\sum_{A\in\pi}\Big(y- \alpha_\P(A) \dot F(A)\Big)^2\1_A(x)\,\mu(\rmd\pi)-\int_{\mathfrak{P}}\Big(\sum_{A\in\pi}\alpha_\P(A) \dot F(A)\1_A(x)-F(x)\Big)^2\,\mu(\rmd\pi).
\end{align*}
Using Huygens formula again for the last term seen as the variance of a random variable in $\pi$ with expectation $F(x)$, we get
\[
\int_{\mathfrak{P}}\Big(\sum_{A\in\pi}\alpha_\P(A) \dot F(A)\1_A(x)-F(x)\Big)^2\,\mu(\rmd\pi)=\int_{\mathfrak{P}}\sum_{A\in\pi}\alpha_\P(A)^2 \dot F(A)^2\1_A(x)\,\mu(\rmd\pi)-F(x)^2.
\]
Collecting the three terms and integrating with respect to $\P$, we obtain the identity
\begin{align*}
   & \P[(y-F(x))^2]\\
    =& \int_{\mathfrak{P}}\sum_{A\in\pi}\P\Big[\Big(y- \alpha_\P(A) \dot F(A)\Big)^2\1_A(x)\Big]\,\mu(\rmd\pi)- \int_{\mathfrak{P}}\sum_{A\in\pi}\dot F(A)^2\1_{\{\P(A)>0\}}\,\mu(\rmd\pi)+ \P[F(x)^2]\\
    =& \int_{\mathcal{A}}\P\Big[\Big(y- \alpha_\P(A) \dot F(A)\Big)^2\1_A(x)\Big]\,\nu(\rmd A)- \int_{\mathcal{A}}\dot F(A)^2\1_{\{\P(A)>0\}}\,\nu(\rmd A)+ \P[F(x)^2].
\end{align*}
Identity~\eqref{eq:identity} then follows readily since $\P[F(x)^2]=\|F\|_{L^2}^2$ and
\[
\|F\|_{\mathcal{H}}^2=\int_{\mathcal{A}}\dot F(A)^2\,\nu(\rmd A)
=\int_{\mathcal{A}}\dot F(A)^2\1_{\{\P(A)>0\}}\,\nu(\rmd A)+\int_{\mathcal{A}}\dot F(A)^2\1_{\{\P(A)=0\}}\,\nu(\rmd A).
\]
Next, we discuss the minimization of the right-hand side of Equation~\eqref{eq:identity}, namely
\[
\int_{\mathcal{A}}\P\Big[\Big(y- \frac{\dot F(A)}{\sqrt{\P(A)}}\Big)^2 \1_A(x)\Big]\,\nu(\rmd A)
\;+\;\int_{\mathcal{A}}\dot F(A)^2 \1_{\{\P(A)=0\}}\,\nu(\rmd A).
\]
The first integral depends only on the values $\dot F(A)$ for sets $A$ with $\P(A)>0$, 
while the second integral depends only on the values $\dot F(A)$ for sets $A$ with $\P(A)=0$. 
Hence the two terms can be minimized separately.

Clearly, the second integral is minimized by setting $\dot F(A)=0$ whenever $\P(A)=0$. 
For the first integral, fix $A\in\mathcal{A}$ with $\P(A)>0$. Then, for any $c\in\mathbb{R}$,
\[
\P\big[(y-c)^2 \1_A(x)\big]
   = \P(A)\,\P\big[(y-c)^2 \mid x\in A\big],
\]
which is uniquely minimized at
\[
c = \P[y \mid x\in A] 
   = \frac{\P[y\1_A(x)]}{\P(A)}.
\]
We can therefore deduce that any minimizer $F\in\mathcal{H}$ of the first integral must satisfy, 
for $\nu$-almost every $A$ such that $\P(A)>0$,
\[
\frac{\dot F(A)}{\sqrt{\P(A)}} 
   = \frac{\P[y\1_A(x)]}{\P(A)} \qquad \nu\text{-a.e.},
\]
or equivalently,
\[
\dot F(A) = \frac{\P[y\1_A(x)]}{\sqrt{\P(A)}} \qquad \nu\text{-a.e.}.
\]
Note that the function $\dot F$ defined this way lies in $L^2(\mathcal{A},\nu)$, and the corresponding element $F\in\mathcal{H}$ is given by
\begin{align*}
F(z)
&= \int_{\mathcal{A}} \dot F(A)\,\frac{\1_A(z)}{\sqrt{\P(A)}} \,\nu(\rmd A) \\
&= \int_{\mathcal{A}} \frac{\P[y\1_A(x)]}{\P(A)} \,\1_A(z) \,\nu(\rmd A) \\
&= \bar T(z;\P,\mu), \quad z\in[0,1]^p,
\end{align*}
which concludes the proof.
\end{proof}

\begin{proof}[Proof of Theorem~\ref{thm:fundamental-property}] Expanding the square $(y-F(x))^2$, the left-hand side of Equation~\eqref{eq:identity} becomes
\[
\P\big[(y-F(x))^2\big]+\|F\|_{\mathcal{H}}^2-\|F\|_{L^2}^2
   \;=\; -2\,\P[yF(x)] \;+\; \|F\|_{\mathcal{H}}^2 \;+\; \P[y^2].
\]
Since the term $\P[y^2]$ does not depend on $F$, the optimization problems~\eqref{eq:optimization-property} and~\eqref{eq:optimization-property-bis} are equivalent. 
The solution to these equivalent problems is given in Lemma~\ref{lem:identity}.
\end{proof}

\begin{proof}[Proof of Lemma~\ref{lem:KME-MMD}]
We detail the proof for the kernel $k_0$ only, as the argument for $k$ is entirely analogous. The expression of the KME follows from the direct computation, using Equation~\eqref{eq:def-k0} defining the kernel:
\begin{align*}
\Phi(\mathrm{Q})
&= \int_{[0,1]^p} k_0(x, \cdot)\,\mathrm{Q}(\mathrm{d}x) \\
&= \int_{[0,1]^p} \int_{\mathcal{A}} \1_A(x)\,\1_A(\cdot)\,\nu(\mathrm{d}A)\,\mathrm{Q}(\mathrm{d}x) \\
&= \int_{\mathcal{A}} \mathrm{Q}(A)\,\1_A(\cdot)\,\nu(\mathrm{d}A).
\end{align*}
It follows from Proposition~\ref{prop:RKHS-structure} 
(or an analogous result adapted to the kernel $k_0$) 
that $\Phi(\mathrm{Q})$ admits the representative $A \mapsto \mathrm{Q}(A)$ in $\dot{\mathcal{H}}_0$. 
Hence,
\[
\|\Phi(\mathrm{Q}_1) - \Phi(\mathrm{Q}_2)\|_{\mathcal{H}_0}^2
= \|\mathrm{Q}_1(A) - \mathrm{Q}_2(A)\|_{L^2(\nu)}^2
= \int_{\mathcal{A}} \big|\mathrm{Q}_1(A) - \mathrm{Q}_2(A)\big|^2\,\nu(\mathrm{d}A),
\]
from which the claimed expression for the MMD immediately follows.
\end{proof}

\begin{proof}[Proof of Proposition~\ref{prop:characteristic-kernel}]
The characteristic property in item~(i) is equivalent to the statement that, 
for any two measures $\mathrm{Q}_1, \mathrm{Q}_2 \in \mathcal{P}([0,1]^p)$,
\[
\|\Phi(\mathrm{Q}_1) - \Phi(\mathrm{Q}_2)\|_{\mathcal{H}_0} = 0 
\quad \implies \quad 
\mathrm{Q}_1 = \mathrm{Q}_2.
\]
From Lemma~\ref{lem:KME-MMD}, we have
\[
\|\Phi(\mathrm{Q}_1) - \Phi(\mathrm{Q}_2)\|_{\mathcal{H}_0}^2
= \int_{\mathcal{A}} 
  \big|\mathrm{Q}_1(A) - \mathrm{Q}_2(A)\big|^2 \,
  \nu(\mathrm{d}A).
\]
Hence,
\[
\|\Phi(\mathrm{Q}_1) - \Phi(\mathrm{Q}_2)\|_{\mathcal{H}_0} = 0
\quad \Longleftrightarrow \quad
\mathrm{Q}_1(A) = \mathrm{Q}_2(A)
\ \text{for $\nu(\mathrm{d}A)$-almost every $A$.}
\]
As a consequence, the kernel $k_0$ is characteristic 
if and only if, for any $\mathrm{Q}_1, \mathrm{Q}_2 \in \mathcal{P}([0,1]^p)$,
\[
\mathrm{Q}_1(A) = \mathrm{Q}_2(A)
\ \text{for $\nu(\mathrm{d}A)$-a.e.}
\quad \implies \quad 
\mathrm{Q}_1 = \mathrm{Q}_2,
\]
which is precisely the definition of a determining measure 
(Definition~\ref{def:determining-measure}). 
This proves the equivalence $i) \Leftrightarrow ii)$.

\medskip
Next, assume that Assumption~\eqref{eq:assumption-1} holds. 
Define the measure 
\[
\nu_{\P}(\mathrm{d}A) 
= \alpha_{\P}(A)\,\nu(\mathrm{d}A),
\]
with weight $\alpha_\P(A)$ defined in Lemma~\ref{lem:KME-MMD}.
By an argument identical to the one used for $k_0$, 
one can show that the kernel $k$ is characteristic 
if and only if $\nu_{\P}$ is determining. 
Since $\nu$ and $\nu_{\P}$ are equivalent measures,  $\nu$ is determining if and only if $\nu_{\P}$ is determining. 
This establishes the equivalence $ii) \Leftrightarrow iii)$.
\end{proof}

\begin{proof}[Proof of Proposition~\ref{prop:characteristic-kernel-2}]
Since the random partition $\Pi$ is generated by a randomized tree with depth at most $d < p$,
the support of the block intensity measure $\nu$ defined by~\eqref{eq:def-nu} 
is contained in $\mathcal{A}_d$.  

For each subset $J \subset [\![1,p]\!]$, we have $\mathcal{A}_J \subset \mathcal{A}_d$, 
and we introduce the kernels
\[
k_0^J(x,x')
= \int_{\mathcal{A}_J} \1_A(x)\,\1_A(x')\,\nu(\mathrm{d}A),
\qquad 
k_0^{-J}(x,x')
= \int_{\mathcal{A}_d \setminus \mathcal{A}_J} 
  \1_A(x)\,\1_A(x')\,\nu(\mathrm{d}A),
\]
together with their associated RKHSs $\mathcal{H}_0^J$ and $\mathcal{H}_0^{-J}$.  
The relation $k_0 = k_0^J + k_0^{-J}$ implies 
$\mathcal{H}_0 = \mathcal{H}_0^J + \mathcal{H}_0^{-J}$ at the RKHS level 
(see \citet[Proposition~3.22]{HE15}). 
Hence $\mathcal{H}_0^J \subset \mathcal{H}_0$ for all $J$, 
and therefore 
\[
\sum_{|J|=d} \mathcal{H}_0^J \subset \mathcal{H}_0.
\]

On the other hand, since $\mathcal{A}_d$ admits the cover
$\mathcal{A}_d = \bigcup_{|J|=d} \mathcal{A}_J$, it follows that
 $\sum_{|J|=d} k_0^J - k_0$ is a positive definite kernel. 
Consequently, $\mathcal{H}_0 \subset \sum_{|J|=d} \mathcal{H}_0^J$. 
Combining both inclusions yields the equality
\[
\mathcal{H}_0 = \sum_{|J|=d} \mathcal{H}_0^J.
\]

\medskip
\noindent
\textbf{Step 1: $i) \Rightarrow ii)$.}  
Assume that $i)$ holds, and let $q > 1$ and $\mathrm{Q} \in \mathcal{P}([0,1]^p)$.  
For any fixed subset $J$ such that $|J| = d$, 
let $L^q_J(\mathrm{Q}) \subset L^q([0,1]^p,\mathrm{Q})$ 
denote the subspace spanned by indicator functions of hyperrectangles 
$A \in \mathcal{A}_J$ 
(that is, functions depending only on the coordinates indexed by $J$).  
Since $\nu_J$ is assumed to be determining on $[0,1]^J$, 
an application of Theorem~\ref{thm:characteristic-kernel} 
implies that $\mathcal{H}_0^J$ is dense in $L^q_J(\mathrm{Q})$.  
Therefore, in $L^q([0,1]^p,\mathrm{Q})$,
\[
\overline{\mathcal{H}_0}
= \overline{\sum_{|J|=d} \mathcal{H}_0^J}
\supset \sum_{|J|=d} \overline{\mathcal{H}_0^J}
= \sum_{|J|=d} L^q_J(\mathrm{Q})
= L^q_d([0,1]^p,\mathrm{Q}).
\]
The reverse inclusion follows since $\mathcal{H}_0 \subset L^q_d([0,1]^p,\mathrm{Q})$, 
which is closed. 
Hence, $\overline{\mathcal{H}_0} = L^q_d([0,1]^p,\mathrm{Q})$, 
proving item~$ii)$.

\medskip
\noindent
\textbf{Step 2: $ii) \Rightarrow i)$.}  
We prove the converse.  
Assume that $i)$ does not hold, so that 
$\nu_{J_0}$ is not determining on $[0,1]^{J_0}$ for some subset $J_0$ with $|J_0| = d$.  
Then, by Theorem~\ref{thm:characteristic-kernel}, 
the kernel $k_0^{J_0}$ is not characteristic, 
and there exists a probability measure $\mathrm{Q}_0$ on $[0,1]^{J_0}$ 
and some $q > 1$ such that $\mathcal{H}_0^{J_0}$ 
is not dense in $L^q([0,1]^{J_0},\mathrm{Q}_0)$.  
Lift $\mathrm{Q}_0$ to a probability measure $\mathrm{Q}$ on $[0,1]^p$ 
by tensorization with a Dirac mass at~$0$ on the remaining $p-d$ coordinates.  
It is then straightforward to verify that 
$\mathcal{H}_0$ is not dense in $L^q([0,1]^p,\mathrm{Q})$, 
which shows that $ii)$ does not hold. 
\end{proof}

\begin{proof}[Proof of Proposition \ref{prop:refinement}] \emph{Proof of i).} Given that $k_0$ is always bounded, both $k_0^{(1)}$ and $k_0^{(2)}$ share the same domain. By Theorem 12 of \cite{BTA04}, we need to show that there exists a constant $C\geq 0$ such that $Ck_0^{(2)}-k_0^{(1)}$ is semi-definite positive. For $m\geq 1$, $z_1,\ldots,z_m\in[0,1]^p$ and $a_1,\ldots,a_p$, recalling that $\P (A)>0$ for all $A \in \Pi_2$ and thus for all $A \in \Pi_1$, we have
\begin{align*}
    &\sum_{1\leq i,j\leq m}a_ia_j(Ck^{(2)}-k^{(1)})(z_i,z_j)\\
    =&\E_\Pi \left[ \sum_{1\leq i,j\leq m}a_ia_j \Big(C\sum_{A_2\in\Pi_2}\1_{A_2}(z_i)\1_{A_2}(z_j) -\sum_{A_1\in\Pi_1}\1_{A_1}(z_i)\1_{A_1}(z_j)\Big)\right] \\
    =& \E_\Pi \left[ C\sum_{A_2\in\Pi_2}\psi(A_2)^2 -\sum_{A_1\in\Pi_1}\psi(A_1)^2\right],
\end{align*}
where we use the notation $\psi(A) = \sum_{i=1}^m a_i\1_{A}(z_i) $ for arbitrary rectangles.

Given that $\Pi_2$ is a refinement of $\Pi_1$, for each $A_1 \in \Pi_1$, there exists a finite partition $D(A_1) = \{D_1, ..., D_K\} \subseteq \Pi_2$ of $A_1$. It follows that $$\sum_{A_2\in \Pi_2} \psi(A_2)^2 = \sum_{A_1\in \Pi_1}\sum_{A_2\in D(A_1)} \psi(A_2)^2$$ so that 
$$\sum_{1\leq i,j\leq m}a_ia_j(Ck^{(2)}-k^{(1)})(z_i,z_j) = \E_\Pi \left[\sum_{A_1\in\Pi_1} \bigg(C\sum_{A_2\in D(A_1)}\psi(A_2)^2 -\psi(A_1)^2\bigg)\right].$$
Now, notice that, for any $A_1\in \Pi_1$, $$\psi(A_1) =\sum_{i=1}^m a_i\1_{A_1}(z_i) =\sum_{i=1}^m a_i\sum_{D \in D(A_1)} \1_{D}(z_i) = \sum_{D \in D(A_1)} \psi(D). $$
An application of Cauchy-Schwarz inequality gives, for any $A_1 \in \Pi_1$, 
\begin{equation*}
    \psi(A_1)^2 = \bigg(\sum_{D \in D(A_1)} \psi(D)\bigg)^2 \leqslant |D(A_1)|  \sum_{D \in D(A_1)} \psi(D)^2 \leqslant C_0 \sum_{D \in D(A_1)} \psi(D)^2,
\end{equation*}
with $C_0= \operatorname*{ess\,sup} \max_{A_1\in \Pi_1} \rvert D(A_1)\rvert$.
 This shows that, $a.s.$, for all $A_1\in \Pi_1,$ $$C_0\sum_{D\in D(A_1)}\psi(D)^2-\psi(A_1)^2\geqslant 0.$$ The result follows by summing over $\Pi_1$ and integrating with respect to $\mu$.

 \emph{Proof of ii).} Assuming that $C_k$ is finite, to show the inclusion $\mathcal{H}^{(1)}\subset \mathcal{H}^{(2)}$, we follow the same strategy as above, i.e.\ we show that $C_k k^{(2)} - k^{(1)}$ is a semi-definite positive kernel on $[0,1]^p$. Note that this also implies the bound $k^{(1)}(z,z) \leq C_k k^{(2)}(z,z)$ since semi-definite positive kernels take nonnegative values on the diagonal.
 This amounts to showing that $$ \E_\Pi \left[\sum_{A_1\in\Pi_1} \bigg(C_k\sum_{A_2\in D(A_1)}\psi(A_2)^2\alpha^2_\P(A_2) -\psi(A_1)^2\alpha^2_\P(A_1)\bigg)\right]\geqslant0.$$
The Cauchy-Schwarz inequality shows that
\begin{align*}
\alpha_\P^2(A_1)\,\psi^2(A_1)
&\leq \alpha_\P^2(A_1)
\left(\sum_{D \in D(A_1)} \frac{1}{\alpha_\P^2(D)}\right)\left(\sum_{D \in D(A_1)} \psi^2(D)\,\alpha_\P^2(D)\right)\\
&\leq C_k \sum_{D \in D(A_1)} \psi^2(D)\,\alpha_\P^2(D), 
\end{align*}
with $C_k$ as defined in the statement, concluding the proof.
\end{proof}

\begin{proof}[Proof of Proposition \ref{prop:integralOP}]
  The arguments are taken from \citet[Section 4.6]{HE15}, but only continuous kernels are considered there, so we still develop the proof.
  By defining $k_x:=k(x,\cdot)\in \mathcal{H}$, the reproducing property of $k$ implies that for any $x,y\in [0,1]^p$,
  \[
    k(x,y) = \langle k_x, k_y\rangle_{\mathcal{H}} \leq \norm{k_x}_{\mathcal{H}}\norm{k_y}_{\mathcal{H}} = \big(k(x,x) k(y,y) \big)^{\frac12}.
  \]
  As a result, we have
  \[
    \int k(x,y)^2\,\P(\rmd x)\P(\rmd y) \leq \bigg(\int k(x,x) \,\P(\rmd x)\bigg)^2,
  \]
  which is finite by Proposition~\ref{prop:k_0-k} under the assumption that $\Pi$ has an integrable size, see Assumption~\eqref{eq:integrability-pi}.
  This ensures that $K$ is an integral Hilbert--Schmidt operator on $L^2$, with a Hilbert--Schmidt norm bounded by $\int k(x,x) \,\P(\rmd x)$.
  Therefore it is a compact operator and the fact that $k$ is symmetric ensures that it is self-adjoint.
  To show that it is non-negative definite, we compute, for $F\in L^2$,
  \begin{align*}
    \langle KF, F\rangle_{L^2} &= \int F(x) \E_\Pi \bigg[ \sum_{A\in \Pi} \alpha^2_\P(A)\1_{A}(x)\1_{A}(x')F(x')  \bigg]\,\P(\rmd x)\P(\rmd x')\\
    &= \E_\Pi \bigg[ \sum_{A\in \Pi} \alpha^2_\P(A) \bigg(\int \1_{A}(x)F(x) \,\P(\rmd x) \bigg)^2 \bigg] \; \geq 0.
  \end{align*}
  Let us now show that the image of $K$ is in $\mathcal{H}$. For $F\in L^2$ and $z\in [0,1]^p$, we have
  \begin{align*}
    KF(z) &= \int_{[0,1]^p} k(x,z) F(x)\, \P(\rmd x)\\
    &= \int_{[0,1]^p} \E_\Pi \bigg[  \sum_{A\in \Pi}\alpha_\P^2(A)\1_{A}(x)\1_{A}(z) F(x) \bigg]\, \P(\rmd x)\\
    &= \E_\Pi \bigg[  \sum_{A\in \Pi}\alpha_\P(A)\1_{A}(z) \bigg(\int_{[0,1]^p} \alpha_\P(A)\1_{A}(x)F(x) \, \P(\rmd x)\bigg) \bigg],
  \end{align*}
  where the interchange of integrals is justified by the same computation by taking the absolute value of $F$ instead: the final integral is finite for $\P(\rmd x)$-almost every $z$ since $KF$ is well-defined in $L^2(\P(\rmd x))$.
  This computation shows that $KF\in \mathcal{H}$, with representative in $\dot{\mathcal{H}}$ given by
  \[
    A \mapsto \int_{[0,1]^p} \frac{\1_{\P(A)>0}}{\sqrt{\P(A)}}\1_{A}(x)F(x) \, \P(\rmd x).
  \]
  We can now study $K$ as a self-adjoint operator on $\mathcal{H}$.
  The key observation is that for $F,G\in \mathcal{H}$, we have
  \begin{align*}
    \langle KF, G\rangle_{\mathcal{H}} &= \E_\Pi \left[ \sum_{A\in \Pi} \int_{[0,1]^p} \alpha_\P(A)\1_{A}(x)F(x)\dot{G}(A) \, \P(\rmd x) \right]\\
    &=  \int_{[0,1]^p} \bigg(\E_\Pi \left[ \sum_{A\in \Pi} \alpha_\P(A)\1_{A}(x)\dot{G}(A)\right] \bigg) \, F(x) \, \P(\rmd x) \\
    &= \int_{[0,1]^p} G(x) F(x) \, \P(\rmd x) \;=\; \langle F, G\rangle_{L^2}.
  \end{align*}
  This means we have $K=i^*i$, where $i:\mathcal{H}\to L^2$ is the inclusion operator.
  We also deduce from this that $\mathrm{ker}(K)=\{0\}$, which implies that its range is dense (see e.g.~\citealt[Theorem~3.3.7]{HE15}), concluding the proof of point (i).\\

  Point (ii) is then deduced from the spectral theorem for compact operators, and the fact that the Hilbert--Schmidt norm is the sum of the squared eigenvalues.\\
  
  For point (iii), we know from Propositions~\ref{prop:boundedness} and~\ref{prop:continuity} that under Assumptions~\eqref{eq:assumption-1}--\eqref{eq:assumption-2}, the kernel $k$ is a bounded and continuous function on $D=[0,1]^p$ which is a compact metric space, therefore we can directly apply Mercer's theorem \citep[Theorems~4.6.5 and~4.6.7]{HE15} to get the result.
\end{proof}

\begin{proof}[Proof of Proposition \ref{prop:eff_sampleSize}]
    For all $i,j\in [\![1,n]\!]$, we have $ W_{ni}(X_j)\in  [0,1]$ so that $W_{ni}(X_j) \geq W_{ni}(X_j)^2$ and we deduce 
    \[
    n=\sum_{i=1}^n \sum_{j=1}^n W_{ni}(X_j) \geq \sum_{i=1}^n \sum_{j=1}^nW_{ni}(X_j)^2.
    \]
    Inverting and multiplying by $n$ leads to the lower bound $N_{\mathrm{eff}}\geq 1$. For the upper bound, an application of Cauchy-Schwarz inequality shows that 
    \[
    1 = \left(\sum_{i=1}^nW_{ni}(X_j) \right)^2\leqslant n \sum_{i=1}^nW_{ni}(X_j)^2.
    \]
    Summing over $j\in [\![1,n]\!]$, inverting and multiplying by $n$ gives the upper bound $N_{\mathrm{eff}}\leq n$. 
\end{proof}

\begin{proof}[Proof of Proposition \ref{prop:VI}]
  Let $X_{(j)}$ be an arbitrary column of the design matrix. Proofs of i) and iii) are straightforward and thus omitted. For ii), let $\P_\mu$ denote either $\mu$ in the case of infinite forests or the empirical measure with point masses at each of the $B$ partitions $\{\Pi_b\}_{b=1}^B.$ We write $\E_{\P_\mu}$ to denote the expectation with respect to $\P_\mu$. It follows that we can decompose $$ W = \E_{\P_\mu} \left[ W_b\right]$$
  where $W_b$ is the matrix weight for tree $b$. It can be shown that $W_b$ is an orthogonal projection matrix (thus with eigenvalues $0$ and $1$) onto the subspace of $\mathbb{R}^n$ whereby vectors are constant within the nodes of the $b$-th tree. Thus, an application of Jensen's inequality gives $$ \rVert W \rVert_2 =   \big\rVert \E_{\P_\mu} \left[ W_b \right]\big\rVert_2 \leqslant \E_{\P_\mu}  \left[ \rVert  W_b \rVert_2\right] = 1.$$
  This shows that $W$ has eigenvalues between $0$ and $1$ and thus, since $\mathrm{GVI}$ is a Rayleigh quotient with $W^2$, its range is bounded in $[0,1]$.
\end{proof}

\section{Proofs related to Section~\ref{sec:IGB}}\label{sec:proofs-2}
\begin{proof}[Proof of Theorem~\ref{thm:rf-as-gradient}]
Clearly, the squared error satisfies, for $F, G \in \mathcal{H}$ and $x \in [0,1]^p$,
\[
\frac{1}{2} (y - (F+G)(x))^2 
= \frac{1}{2} (y - F(x))^2 - (y - F(x))\,G(x) + \frac{1}{2} G(x)^2.
\]
Integrating with respect to $\P$ and using 
$\lvert G(x) \rvert \le \|G\|_{\mathcal{H}}\,k(x,x)^{1/2}$, we deduce
\[
\Big| R_{\P}(F+G) - R_{\P}(F) 
   + \P[G(x)(y - F(x))] \Big|
   \le \|G\|_{\mathcal{H}}^2 \int_{[0,1]^p} k(x,x)\,\P(\mathrm{d}x)<\infty.
\]
It follows that the Fréchet derivative of $R_{\P}$ at $F$ is the linear map
\[
\mathrm{d}_F R_{\P} : 
G \longmapsto -\,\P\big[G(x)\,(y - F(x))\big].
\]

By Proposition~\ref{prop:RKHS-structure}, any $G \in \mathcal{H}$ can be written as
\[
G(x) = \int_{\mathcal{A}} \dot{G}(A)\,g_{\P}(x,A)\,\nu(\mathrm{d}A),
\]
with $\dot{G} \in \dot{\mathcal{H}} \subset L^2(\mathcal{A},\nu)$.  
We can therefore rewrite
\begin{equation}\label{eq:proof_gradient_L}
\mathrm{d}_F R_{\P}(G)
= -\,\P\bigg[
  \int_{\mathcal{A}} 
  \dot{G}(A)\,g_{\P}(x,A)\,(y - F(x))\,\nu(\mathrm{d}A)
\bigg].
\end{equation}

To justify the exchange of integration, we use the Cauchy--Schwarz inequality:
\begin{align*}
  &\P\bigg[
    \int_{\mathcal{A}} 
    \big|\dot{G}(A)\,g_{\P}(x,A)\,(y - F(x))\big|
    \,\nu(\mathrm{d}A)
  \bigg] \\
  &\le
  \Bigg(\int_{\mathcal{A}} \dot{G}(A)^2\,\nu(\mathrm{d}A)\Bigg)^{1/2}
  \P\Bigg[
    \Bigg( \int_{\mathcal{A}} (y - F(x))^2\,g_{\P}(x,A)^2\,\nu(\mathrm{d}A)
    \Bigg)^{1/2}
  \Bigg].
\end{align*}
The first factor is finite since $\dot{G} \in L^2(\nu)$, 
and the second factor can be rewritten as
\[
\P\Big[ 
  |y - F(x)| 
  \Big( \int_{\mathcal{A}} g_{\P}(x,A)^2\,\nu(\mathrm{d}A) \Big)^{1/2}
\Big]
= \P\big[ |y - F(x)|\,k(x,x)^{1/2} \big],
\]
which is finite because $y$ has a finite second moment and 
both $F$ and $x \mapsto k(x,x)^{1/2}$ belong to $L^2(\P(\mathrm{d}x))$.

Hence, \eqref{eq:proof_gradient_L} can be rewritten as
\[
\mathrm{d}_F R_{\P}(G)
= -\int_{\mathcal{A}} 
   \dot{G}(A)\,\P[(y - F(x))\,g_{\P}(x,A)]\,\nu(\mathrm{d}A)
= -\,\langle \Gamma_F, G \rangle_{\mathcal{H}},
\]
where $\Gamma_F$ is defined by
\[
\Gamma_F(z)
= \int_{\mathcal{A}} 
   \frac{\P[(y - F(x))\,\1_A(x)]}{\P(A)} 
   \1_A(z)\,\nu(\mathrm{d}A),
\]
and satisfies
\[
\dot{\Gamma}_F(A)
= \frac{\P[(y - F(x))\,\1_A(x)]}{\sqrt{\P(A)}}
\quad \text{(mod $\dot{\mathcal{H}}^\perp$)},
\]
see Proposition~\ref{prop:RKHS-structure} and Equation~\eqref{eq:modulo}.  
Note that the term $\1_A(x)\,\1_{\{\P(A)=0\}}$ 
in the definition of $g_{\P}(x,A)$ 
has no contribution since it vanishes $\P(\mathrm{d}x)$-a.e.  
This proves that $\nabla R_{\P}(F) = -\,\Gamma_F$.  
Setting $F = 0$, we obtain
\[
\nabla R_{\P}(0) = -\,\bar{T}(\cdot; \P),
\]
as claimed.
\end{proof}

\begin{proof}[Proof of Theorem~\ref{thm:grf-as-gradient}]
The proof is similar to that of Theorem~\ref{thm:rf-as-gradient}, and we provide only the main arguments.
By a Taylor expansion and using that $\ell$ has a bounded second-order derivative with respect to its first variable (with bound $C$), we have
\[
\big|\ell(F(x)+G(x), y) - \ell(F(x), y) - G(x)\,\partial_1\ell(F(x), y)\big|
\le \frac{C}{2}\,G(x)^2 
\le \frac{C}{2}\,\|G\|_{\mathcal{H}_{\mathrm{P},F}}^2\,k_{\mathrm{P},F}(x,x).
\]
Integrating with respect to $\mathrm{P}$ yields
\[
R_{\mathrm{P}}(F+G) 
= R_{\mathrm{P}}(F) + \mathrm{P}\big[G(x)\,\partial_1 \ell(F(x),y)\big] + O(\|G\|_{\mathcal{H}_{\mathrm{P},F}}^2),
\]
showing that the Fréchet derivative of $R_{\mathrm{P}}$ at $F$ is the linear map
\[
\mathrm{d}_F R_{\mathrm{P}} : 
G \longmapsto \mathrm{P}\big[G(x)\,\partial_1 \ell(F(x),y)\big].
\]
Similarly to Equation~\eqref{eq:proof_gradient_L}, 
Proposition~\ref{prop:RKHS-structure} implies that
\[
\mathrm{d}_F R_{\mathrm{P}}(G)
= \mathrm{P}\bigg[\int_{\mathcal{A}} 
  \dot{G}(A)\,\partial_1 \ell(F(x),y)\,g_{\mathrm{P}}(x,A)\,
  \nu_{\mathrm{P},F}(\mathrm{d}A)\bigg].
\]
We may exchange the order of integration under the integrability condition
\begin{equation*}\label{eq:proof_gradient_L_bis}
\mathrm{P}\bigg[
\Big(
  \int_{\mathcal{A}} 
  (\partial_1 \ell(F(x),y))^2\,g_{\mathrm{P}}(x,A)^2\,
  \nu_{\mathrm{P},F}(\mathrm{d}A)
\Big)^{1/2}
\bigg] < \infty,
\end{equation*}
which holds as in the proof of Theorem~\ref{thm:rf-as-gradient}, 
since $|\partial_1 \ell(F(x),y)| \le |\partial_1\ell(0,y)| + C|F(x)|$.  
Exchanging the order of integration then gives
\[
\mathrm{d}_F R_{\mathrm{P}}(G)
= \int_{\mathcal{A}} 
  \dot{G}(A)\,\mathrm{P}[\partial_1 \ell(F(x),y)\,g_{\mathrm{P}}(x,A)]\,
  \nu_{\mathrm{P},F}(\mathrm{d}A),
\]
so that 
\[
\mathrm{d}_F R_{\mathrm{P}}(G)
= \langle \Gamma_F, G \rangle_{\mathcal{H}_{\mathrm{P},F}},
\]
where
\[
\Gamma_F(z)
= \int_{\mathcal{A}}
  \frac{\mathrm{P}[\partial_1 \ell(F(x),y)\,\1_A(x)]}{\mathrm{P}(A)}\,
  \1_A(z)\,
  \nu_{\mathrm{P},F}(\mathrm{d}A).
\]
We can observe that $\Gamma_F(z) = -\,\bar{T}(z; \mathrm{P}, F, \mu_{\mathrm{P},F})$, 
and therefore conclude that 
$\nabla R_{\mathrm{P}}(F) = -\,\bar{T}(\cdot; \mathrm{P}, F, \mu_{\mathrm{P},F})$.
\end{proof}

\begin{proof}[Proof of Lemma~\ref{lem:RKHS-igb}]
The proof is divided into two parts: first, we compare the reproducing kernel Hilbert spaces, and second, we establish the regularity of the inner product.

\medskip
\noindent\textbf{Encoding via splitting schemes.}
As recalled in Appendix~\ref{app:splitting-schemes}, the partition generated by a softmax gradient tree of depth $d \geq 1$ can be represented by a \emph{splitting scheme} 
$\xi = (j_v, u_v)_{v \in \mathcal{T}_{d-1}}$,
which records, at each internal node $v$ of the binary tree $\mathcal{T}_d$, the variable index $j_v \in [\![1,p]\!]$ and the threshold $u_v \in (0,1)$ used to perform the split. 
We denote by $Q_{\mathrm{P},F}$ the distribution of the splitting scheme of the softmax gradient tree at $F\in L^2$, the parameters $d\geq 1$, $K\geq 1$ and $\beta\geq 0$ remaining implicit. In the totally random case ($K = 1$ or $\beta = 0$), the corresponding reference distribution is denoted by $Q_0$. 
A key ingredient in the proof is that the Radon--Nikodym derivative 
$\mathrm{d}Q_{\mathrm{P},F}/\mathrm{d}Q_0$
exists and admits an explicit expression, given in Proposition~\ref{prop:app-RN-splitting-scheme}.

\medskip
\noindent\textbf{First point: comparison properties.}
The fact that the RKHSs coincide as sets and have equivalent norms follows from the uniform bounds
\begin{equation}\label{eq:RN-bounded}
  c \;\le\; \frac{\mathrm{d}Q_{\mathrm{P},F}}{\mathrm{d}Q_0}(\xi)
  \;\le\; c^{-1},
  \qquad \text{for some } c>0,
\end{equation}
which are a direct consequence of Propositions~\ref{prop:app-RN-splitting-scheme} and~\ref{prop:RN-diff}.

Let $(A_v^\xi)_{v\in\{0,1\}^d}$ denote the leaves of the partition of $[0,1]^p$ induced by the splitting scheme $\xi$.  
The reproducing kernel $k_{\mathrm{P},F}$ can then be expressed as
\begin{equation}\label{eq:kernel-xi}
k_{\mathrm{P},F}(x,x')
= \int \sum_{v\in\{0,1\}^d}
   g_\P(x,A_v^\xi)g_\P(x',A_v^\xi)\,
   Q_{\mathrm{P},F}(\mathrm{d}\xi).
\end{equation}
An analogous representation holds for $k_{\mathrm{P}}^0$, with $Q_{\mathrm{P},F}$ replaced by $Q_0$.  
Introducing the Radon--Nikodym derivative, we can write
\[
\bigl(k_{\mathrm{P},F}-c\,k_{\mathrm{P}}^0\bigr)(x,x')
= \int \sum_{v\in\{0,1\}^d}
   g_\P(x,A_v^\xi)g_\P(x',A_v^\xi)
   \Bigl(\frac{\mathrm{d}Q_{\mathrm{P},F}}{\mathrm{d}Q_0}(\xi)-c\Bigr)\,
   Q_0(\mathrm{d}\xi),
\]
which defines a nonnegative definite kernel by virtue of~\eqref{eq:RN-bounded}.  
By \citet[Theorem~2.7.11]{HE15}, this implies the inclusion
\[
\mathcal{H}_{\mathrm{P}}^0 \subset \mathcal{H}_{\mathrm{P},F}
\quad \text{and} \quad
\|G\|_{\mathcal{H}_{\mathrm{P},F}}
\;\le\; c^{-1/2}\,\|G\|_{\mathcal{H}_{\mathrm{P}}^0},
\quad  G\in \mathcal{H}_{\mathrm{P}}^0,
\]
where we used that scaling a kernel by $c$ multiplies the RKHS norm by $c^{-1/2}$.  
By symmetry, interchanging $Q_{\mathrm{P},F}$ and $Q_0$ yields the reverse inclusion
\[
\mathcal{H}_{\mathrm{P},F} \subset \mathcal{H}_{\mathrm{P}}^0
\quad \text{and} \quad
\|G\|_{\mathcal{H}_{\mathrm{P}}^0}
\;\le\; c^{-1/2}\,\|G\|_{\mathcal{H}_{\mathrm{P},F}},
\quad  G\in \mathcal{H}_{\mathrm{P},F}.
\]
Hence, the two spaces coincide as sets and their norms are equivalent.

\medskip
\noindent\textbf{Second point: regularity of the inner product.}
From \eqref{eq:kernel-xi} and \citet[Theorem~2.7.7]{HE15}, we obtain a representation of $\mathcal{H}_{\mathrm{P},F}$ analogous to that in Proposition~\ref{prop:RKHS-structure}, expressed in terms of $(\xi,v)$. Namely, every $G \in \mathcal{H}_{\mathrm{P},F}$ admits the representation
\[
  G(z)
  = \int \sum_{v\in\{0,1\}^d}
    \dot{G}(\xi,v)g_\P(z,A_v^\xi)\;
    Q_{\mathrm{P},F}(\mathrm{d}\xi),
\]
with $\int \sum_{v} \dot{G}(\xi,v)^2\,Q_{\mathrm{P},F}(\mathrm{d}\xi)<\infty$.
For $G\in \mathcal{H}_{\mathrm{P}}^0$ with representative $\dot{G}$ we also have
\begin{equation}\label{eq:G-representatives}
\begin{aligned}
G(z)
&= \int \sum_{v} \dot{G}(\xi,v)\,
    g_\P(z,A_v^\xi)\;
    Q_0^{\mathrm{P}}(\mathrm{d}\xi) \\
&= \int \sum_{v} \dot{G}(\xi,v)\,
   \Big(\tfrac{\mathrm{d}Q_{\mathrm{P},F}}{\mathrm{d}Q_0}(\xi)\Big)^{-1}
   g_\P(z,A_v^\xi)\;
   Q_{\mathrm{P},F}(\mathrm{d}\xi),
\end{aligned}
\end{equation}
so that $G$, viewed as an element of $\mathcal{H}_{\mathrm{P},F}$, has representative
\[
(\xi,v)\longmapsto 
\dot{G}(\xi,v)
\Big(\frac{\mathrm{d}Q_{\mathrm{P},F}}{\mathrm{d}Q_0}(\xi)\Big)^{-1}.
\]
Let $\mu_0^{\mathrm{P}}(\mathrm{d}\xi,\mathrm{d}v)=Q_0^{\mathrm{P}}(\mathrm{d}\xi)\otimes \#(\mathrm{d}v)$ and $\mu_{\mathrm{P},F}(\mathrm{d}\xi,\mathrm{d}v)=Q_{\mathrm{P},F}(\mathrm{d}\xi)\otimes \#(\mathrm{d}v)$ on $\{0,1\}^d$.  
By \eqref{eq:RN-bounded}, $L^2(\mu_{\mathrm{P},F})=L^2(\mu_0^{\mathrm{P}})$ as sets, with equivalent norms. 
Write $\langle\cdot,\cdot\rangle_0$ and $\langle\cdot,\cdot\rangle_F$ for the inner products on $L^2(\mu_0^{\mathrm{P}})$ and $L^2(\mu_{\mathrm{P},F})$, respectively, and define the invertible self-adjoint operator
\begin{equation}\label{eq:def-A_F}
A_F: L^2(\mu_0^{\mathrm{P}})\to L^2(\mu_0^{\mathrm{P}}),
\qquad
g \longmapsto g \cdot \frac{\mathrm{d}Q_{\mathrm{P},F}}{\mathrm{d}Q_0^{\mathrm{P}}}.
\end{equation}
Then $\langle g_1,g_2\rangle_F = \langle A_F g_1, g_2\rangle_0$ for all $g_1,g_2$.  
Let $\dot{\mathcal{H}}$ be the closure in $L^2(\mu_{\mathrm{P},F})$ of
\[
(\xi,v)\longmapsto 
g_\P(z,A_v^\xi),
\qquad z\in D\subset[0,1]^p,
\]
and denote by $\phi_{\mathrm{P},F}:\mathcal{H}_{\mathrm{P},F}\to\dot{\mathcal{H}}$ and
$\phi_0^{\mathrm{P}}:\mathcal{H}_0^{\mathrm{P}}\to\dot{\mathcal{H}}$ the corresponding isometries.  
Since the norms on $L^2(\mu_{\mathrm{P},F})$ are equivalent across $F$, the set $\dot{\mathcal{H}}$ itself does not depend on $F$.
With this notation, \eqref{eq:G-representatives} implies that if $\dot{G}=\phi_0^{\mathrm{P}}(G)$, then $(A_F^{-1}\dot{G}-\phi_{\mathrm{P},F}(G))\perp \dot{\mathcal{H}}$ with respect to $\langle\cdot,\cdot\rangle_F$, hence
\[
  \phi_{\mathrm{P},F}(G)=P_F A_F^{-1}\dot{G},
\]
where $P_F$ (resp.\ $P_0$) denotes the orthogonal projector onto $\dot{\mathcal{H}}$ in $\langle\cdot,\cdot\rangle_F$ (resp.\ $\langle\cdot,\cdot\rangle_0$).  
Consequently, for fixed $G_1,G_2\in\mathcal{H}_{\mathrm{P}}^0$ with $\dot{G}_i=\phi_0^{\mathrm{P}}(G_i)$,
\begin{align*}
\langle G_1,G_2\rangle_{\mathcal{H}_{\mathrm{P},F}}
&= \langle P_F A_F^{-1}\dot{G}_1,\,P_F A_F^{-1}\dot{G}_2\rangle_F \\
&= \langle P_F A_F^{-1}\dot{G}_1,\,A_F^{-1}\dot{G}_2\rangle_F \\
&= \langle P_F A_F^{-1}\dot{G}_1,\,\dot{G}_2\rangle_0,
\end{align*}
where we used the projector property of $P_F$ and the relation $\langle u,v\rangle_F=\langle u,A_F v\rangle_0$.

It remains to show that the mapping
\[
\psi:\; F\in L^2
\longmapsto P_F A_F^{-1}\in \mathcal{L}\!\big(L^2(\mu_0^{\mathrm{P}})\big)
\]
is continuously Fréchet differentiable, where $\mathcal{L}$ denotes the space of bounded linear operators endowed with the operator norm.  
Then, uniformly over $\|G_i\|_{\mathcal{H}_{\mathrm{P}}^0}\le 1$,
\[
\langle G_1,G_2\rangle_{\mathcal{H}_{\mathrm{P},F+H}}
= \langle G_1,G_2\rangle_{\mathcal{H}_{\mathrm{P},F}}
  + \big\langle (\mathrm{d}_F\psi)[H](\dot{G}_1),\,\dot{G}_2\big\rangle_0
  + o\big(\|H\|_{L^2}\big),
\]
which establishes the Fréchet differentiability of 
$F\mapsto \langle\cdot,\cdot\rangle_{\mathcal{H}_{\mathrm{P},F}}$.

By Proposition~\ref{prop:RN-diff}, the map 
$F\mapsto \tfrac{\mathrm{d}Q_{\mathrm{P},F}}{\mathrm{d}Q_0^{\mathrm{P}}}(\xi)$ 
is uniformly Fréchet differentiable in $\xi$, hence 
$F\mapsto A_F\in\mathcal{L}(L^2(\mu_0^{\mathrm{P}}))$ 
is Fréchet differentiable.  
The differentiability of $\psi$ then follows by composing $F\mapsto A_F$ with the continuously Fréchet differentiable map described in Corollary~\ref{cor:linear-algebra} (applied with $H=L^2(\mu_0^{\mathrm{P}})$ and $K=\dot{\mathcal{H}}$).  
This completes the proof.
\end{proof}

\begin{proof}[Proof of Lemma~\ref{lem:gradient-lipschitz}]
By Proposition~\ref{prop:RKHS-structure} (or its variant using splitting schemes as above), the representative of $\bar T(\cdot;\mathrm{P},F)$ in $\mathcal{H}_{\mathrm{P},F}$ is
\[
(\xi,v)\longmapsto
\mathrm{P}\!\big[\partial_1\ell(F(x),y)\,g_\P(x,A_v^\xi)\big].
\]
Hence, viewed in $\mathcal{H}_{\mathrm{P}}^0$, a representative is
\[
(\xi,v)\longmapsto
\mathrm{P}\!\big[\partial_1\ell(F(x),y)\,g_\P(x,A_v^\xi)\big]
\;\frac{\mathrm{d}Q_{\mathrm{P},F}}{\mathrm{d}Q_0}(\xi),
\]
and, by \eqref{eq:RN-bounded}, the corresponding $\mathcal{H}_{\mathrm{P}}^0$-norms are equivalent to the $L^2(\mu_0^{\mathrm{P}})$-norms of these representatives, up to universal constants.

\medskip
\noindent\textbf{Local Lipschitz property.}
Fix $\delta>0$ and restrict attention to the ball 
$\{F:\|F\|_{\mathcal{H}_{\mathrm{P}}^0}\le \delta\}$. 
For $F_1,F_2$ in this ball, write
\[
\bar T(\cdot;\mathrm{P},F_1, \mu_{\mathrm{P},F_1})-\bar T(\cdot;\mathrm{P},F_2, \mu_{\mathrm{P},F_2})  = \mathrm{I} + \mathrm{II},
\]
where
\begin{align*}
\mathrm{I} &:=
\int\sum_{v}
\mathrm{P}\!\big[(\partial_1\ell(F_1(x),y)-\partial_1\ell(F_2(x),y))\,g_\P(x,A_v^\xi)\big]
\;\frac{\mathrm{d}Q_{\mathrm{P},F_1}}{\mathrm{d}Q_0}(\xi)\,
\1_{A_v^\xi}(\cdot)\,Q_0(\mathrm{d}\xi),\\
\mathrm{II} &:=
\int\sum_{v}
\mathrm{P}\!\big[\partial_1\ell(F_2(x),y)\,g_\P(x,A_v^\xi)\big]
\left(\frac{\mathrm{d}Q_{\mathrm{P},F_1}}{\mathrm{d}Q_0}
      -\frac{\mathrm{d}Q_{\mathrm{P},F_2}}{\mathrm{d}Q_0}\right)(\xi)\,
\1_{A_v^\xi}(\cdot)\,Q_0(\mathrm{d}\xi).
\end{align*}
Using the bound 
\[
|\partial_1\ell(F_1(x),y)-\partial_1\ell(F_2(x),y)|
\le \Big(\sup_{z,y}|\partial_1^2\ell(z,y)|\Big)\,|F_1(x)-F_2(x)|
\]
and applying Cauchy--Schwarz, together with the norm comparison
$\|\cdot\|_{L^2} \leq C_1 \|\cdot\|_{\mathcal{H}_{\mathrm{P}}^0}$ (from Proposition~\ref{prop:propRKHS}), we obtain
\[
\|\mathrm{I}\|_{\mathcal{H}_{\mathrm{P}}^0}
\;\le\; C_1\,\|F_1-F_2\|_{\mathcal{H}_{\mathrm{P}}^0}.
\]
For term $\mathrm{II}$, Proposition~\ref{prop:RN-diff} gives (on bounded sets) a Lipschitz estimate
\[
\Big\|
\frac{\mathrm{d}Q_{\mathrm{P},F_1}}{\mathrm{d}Q_0}
-\frac{\mathrm{d}Q_{\mathrm{P},F_2}}{\mathrm{d}Q_0}
\Big\|_{L^\infty}
\;\le\; C_2(\delta)\,\|F_1-F_2\|_{L^2}
\;\le\; C_2'(\delta)\,\|F_1-F_2\|_{\mathcal{H}_{\mathrm{P}}^0},
\]
and, using that $\mathrm{P}[\partial_1\ell(F_2(x),y)^2]$ is uniformly bounded on the ball by assumption on $\ell$, together with the boundedness of $\tfrac{\mathrm{d}Q_{\mathrm{P},F_2}}{\mathrm{d}Q_0}$ on the ball, we conclude
\[
\|\mathrm{II}\|_{\mathcal{H}_{\mathrm{P}}^0}
\;\le\; C_3(\delta)\,\|F_1-F_2\|_{\mathcal{H}_{\mathrm{P}}^0}.
\]
Combining the two bounds yields the local Lipschitz estimate with constant 
$C_\delta:=C_1(\delta)+C_3(\delta)$.

\medskip
\noindent\textbf{Linear growth.}
Similarly,
\begin{align*}
\|\bar T(\cdot;\mathrm{P},F, \mu_{\mathrm{P},F})\|_{\mathcal{H}_{\mathrm{P}}^0}^2
&\;\lesssim\;
\int \sum_{v\in\{0,1\}^d}
\mathrm{P}\!\big[\partial_1\ell(F(x),y)^2\,g_\P(x,A_v^\xi)^2\big]\;
Q_{\mathrm{P},F}(\mathrm{d}\xi) \\
&\;\le\; C_{\text{tree}}\,
\mathrm{P}\!\big[\partial_1\ell(F(x),y)^2\big],
\end{align*}
for some constant $C_{\text{tree}}>0$ depending only on the tree parameters 
$(d,K,\beta)$ and the bounds in \eqref{eq:RN-bounded}.
Since
\[
|\partial_1\ell(F(x),y)|
\le |\partial_1\ell(0,y)| + C\,|F(x)|,
\]
we get 
$\mathrm{P}[\partial_1\ell(F(x),y)^2]\le C\,(1+\|F\|_{L^2}^2)
\lesssim 1+\|F\|_{\mathcal{H}_{\mathrm{P}}^0}^2$, and hence
\[
\|\bar T(\cdot;\mathrm{P},F, \mu_{\mathrm{P},F})\|_{\mathcal{H}_{\mathrm{P}}^0}
\;\le\; D\,(1+\|F\|_{\mathcal{H}_{\mathrm{P}}^0})
\]
for some $D>0$.
Finally, by Theorem~\ref{thm:grf-as-gradient},
\[
\nabla^{\mathbb{H}_{\mathrm{P}}} R_{\mathrm{P}}(F_1)
- \nabla^{\mathbb{H}_{\mathrm{P}}} R_{\mathrm{P}}(F_2)
= \bar T(\cdot;\mathrm{P},F_2, \mu_{\mathrm{P},F_2})-\bar T(\cdot;\mathrm{P},F_1, \mu_{\mathrm{P},F_1}),
\]
so the claimed Lipschitz and growth properties hold for the gradient vector field.
\end{proof}

\begin{proof}[Proof of Theorem~\ref{thm:igb-gradient-flow}]
Lemma~\ref{lem:gradient-lipschitz} establishes that the vector field
$F\mapsto \nabla^{\mathbb{H}_{\mathrm{P}}} R_{\mathrm{P}}(F)$
is locally Lipschitz and satisfies a linear growth condition on $\mathcal{H}_{\mathrm{P}}^0$.
By the Cauchy--Lipschitz (Picard--Lindelöf) theorem in Hilbert spaces
(see, e.g., \citet[Chapter~9]{Brezis2011}),
the gradient flow equation~\eqref{eq:gradient-flow} with initial condition
\eqref{eq:igb-initialization} admits a unique global solution
defined for all $t\ge 0$.
Since $\nabla^{\mathbb{H}_{\mathrm{P}}} R_{\mathrm{P}}(F)=-\,\bar T(\cdot;\mathrm{P},F, \mu_{\mathrm{P},F})$,
this solution also satisfies~\eqref{eq:igb-ODE} and thus coincides with
the infinitesimal gradient boosting process $(\hat F_t)_{t\ge 0}$.
\end{proof}

\appendix

\section{Theoretical complements}
\subsection{Measurability on the space of partitions}\label{app:measurability-partitions}

Recall from Section~\ref{sec:general-framework} that $\mathcal{A}$ denotes the space of hyperrectangles endowed with the metric $\rho$, and $\mathfrak{P}$ the space of finite partitions of $[0,1]^p$ into hyperrectangles. We present here some technical details on the $\sigma$-algebra considered on $\mathfrak{P}$, together with a few useful properties regarding the measurability of related functionals.

We consider on $\mathfrak{P}$ the $\sigma$-algebra generated by the family of mappings
\[
  A_x: 
  \begin{cases}
    \mathfrak{P} \to \mathcal{A},\\
    \pi \mapsto A_x(\pi), \text{ with } x \in A_x(\pi) \in \pi,
  \end{cases}
  \qquad x \in [0,1]^p.
\]
In words, $A_x(\pi)$ denotes the hyperrectangle of the partition $\pi$ that contains the point $x$.

As a result, it is not difficult to check the measurability of the functionals mentioned in the paper:
\begin{itemize}
  \item The mapping $(A,x) \mapsto \1_{A}(x)$ is measurable, since the set 
  \[
    \{(a,b,x)\in ([0,1]^p)^3 : x\in [a,b\rangle\}
  \]
  is clearly Borel and, by definition of the distance $\rho$ on $\mathcal{A}$, the mapping 
  \[
    [a,b\rangle \in \mathcal{A} \mapsto (a,b) \in ([0,1]^p)^2
  \]
  is continuous, and therefore measurable.
  
 \item If $f : (A, Z) \in \mathcal{A}\times\mathcal{Z} \mapsto \mathbb{R}$ is a measurable function, where $(\mathcal{Z}, \mathscr{Z})$ is a measurable space, then the mapping
\[
  (\pi, Z) \mapsto \sum_{A\in \pi} f(A, Z)
\]
is measurable. Indeed, assuming without loss of generality that $f$ is non-negative, we can write
\[
  \sum_{A\in \pi} f(A, Z)
  = \sup_{n\in \mathbb{N}} 
    \Bigg( 
    \sup_{x_1,\dots,x_n \in ([0,1]\cap \mathbb{Q})^p} 
    \Big( 
      \1_{\{\forall\, 1\leq i < j \leq n,\, A_{x_i}(\pi) \neq A_{x_j}(\pi)\}}
      \sum_{i=1}^n f(A_{x_i}(\pi), Z)
    \Big)
    \Bigg),
\]
which is readily seen to be measurable by definition of the $\sigma$-algebra on $\mathfrak{P}$. Consequently, the mappings
\begin{align*}
  \pi &\mapsto |\pi| = \sum_{A\in \pi} 1,\\
  \pi &\mapsto |\pi \cap B| = \sum_{A\in \pi} \1_B(A),
\end{align*}
are measurable on $\mathfrak{P}$.

\item As another consequence, the definition of the block intensity measure 
\[
  \nu(B) = \int_{\mathfrak{P}} \sum_{A\in\pi} \1_{A\in B} \, \mu(\rmd\pi),
  \qquad B\subset\mathcal{A} \text{ Borel},
\]
is well posed. Moreover, the function
\[
  T(z; \P, \pi)
  = \sum_{A\in\pi} \frac{\P[y\,\1_A(x)]}{\P(A)}\, \1_A(z)
\]
is jointly measurable in $(z,\pi) \in [0,1]^p \times \mathfrak{P}$.

\end{itemize}

\subsection{Softmax gradient trees and splitting schemes} \label{app:splitting-schemes}
We introduce some material from \citet[Section~2.2]{DD24a} that is useful for the analysis of softmax gradient trees. In particular, we introduce the notion of splitting scheme that encodes the splits performed during the construction of the binary tree and explicit its distribution in the softmax gradient tree model with parameters $d\geq 1, K\geq 1, \beta\geq 0$ --- see Section~\ref{sec:models-and-examples} and~\ref{sec:gradient-properties}.

The binary rooted tree with depth $d\geq 1$ (from graph theory) is defined on the vertex set $\mathscr{T}_d=\cup_{l=0}^d \{0,1\}^l$. A vertex $v\in \{0,1\}^l$ is   seen as a word of size $l$ in the letters $0$ and $1$. The empty word  $v=\emptyset$ corresponds to the tree root. The vertex set is divided into the internal nodes $v\in \mathscr{T}_{d-1}$ and the terminal nodes $v\in\{0,1\}^d$, also called leaves. Each internal node $v$ has two children denoted by $v0$ and $v1$ (concatenation of words) while the leaves have no offspring.

A gradient tree is encoded by a \textit{splitting scheme}
\[
\xi=(j_v,u_v)_{v\in\mathscr{T}_{d-1}}\in ([\![1,p]\!]\times (0,1))^{\mathscr{T}_{d-1}}
\]
giving the splits at each internal node. The splitting scheme $\xi$ allows us to associate to each vertex $v\in\mathscr{T}_{d}$ a hyperrectangle $A_v=A_v^\xi$ defined recursively by $A_\emptyset=[0,1]^p$ and, for $v\in\mathscr{T}_{d-1}$,
\begin{equation} \label{eq:split-A0-A1}
  \begin{aligned}
    &A_{v0} = A_v \cap \{x\in[0,1]^p\ :\ x^{j_{v}} < a_v+u_v(b_v-a_v)\},\\
 &A_{v1} = A_v \cap \{x\in[0,1]^p\ :\ x^{j_{v}} \geq a_v+u_v(b_v-a_v)\},
  \end{aligned}
\end{equation}
with $a_v=\inf_{x\in A_v} x^{j_v}$ and $b_v=\sup_{x\in A_v} x^{j_v}$. Note that $A_v$ depends only on the splits attached to the ancestors of $v$. For each level $l=0,\ldots,d$,  $(A_v)_{v\in \{0,1\}^l}$ is a partition   of $[0,1]^p$ into $2^l$ hyperrectangles. 

Considering a loss function $\ell$ satisfying~\eqref{eq:assumptions-on-ell} and~\eqref{eq:regularity-partial-ell}, we consider a gradient tree according to Definition~\ref{def:gradient-tree-and-forest}, built from $\P$ and with an initial predictor $F\in L^2$. We denote by $Q_{\P,F}$ the distribution of the splitting scheme $\xi$ under this procedure, and by $Q_{0}$ its distribution under the specific case $\beta=0$ (the totally random case where the splitting scheme distribution does not depend on $\P$ and $F$).
We provide the expression for the Radon-Nikodym derivative $\rmd Q_{\P,F}/\rmd Q_0$ that was derived in \citet[Proposition~2.1]{DD24a}. The notation $\Delta_{\P,F}(j,u;A)$ stands for the score of the binary split  $A=A_0\cup A_1$ resulting from the choice of covariate $j$ and threshold $u$ --- see Equation~\eqref{eq:split-A0-A1} --- and defined by  
\begin{equation}~\label{eq:def_delta_n}
  \Delta_{\P,F}(j,u;A)= \frac{\P[\partial_1\ell(F(x),y)\1_{A_0}(x)]^2}{\P(A_0)}+\frac{\P[\partial_1\ell(F(x),y)\1_{A_1}(x)]^2}{\P(A_1)},
\end{equation}
which is equivalent to Equation~\eqref{eq:score}.

\begin{proposition}\label{prop:app-RN-splitting-scheme}
For all $F\in L^2$, the splitting scheme distribution $Q_{\P,F}$ is absolutely continuous with respect to $Q_0 $ with Radon-Nikodym derivative
\begin{equation}\label{eq:RN}
\frac{\rmd Q_{\P,F}}{\rmd Q_0}(\xi)= 
\int \prod_{v\in \mathscr{T}_{d-1}}  \frac{\exp(\beta \Delta_{\P,F}(j_v^1,u_v^1; A_v^\xi))}{K^{-1}\sum_{k=1}^K \exp(\beta \Delta_{\P,F}(j_v^{k},u_v^{k}; A_v^\xi))} Q_0(\rmd \xi^2)\cdots Q_0(\rmd \xi^K),
\end{equation}
with $\xi^k=(j_v^k,u_v^k)_{v\in\mathscr{T}_{d-1}}$, $2\leq k\leq K$, and for $k=1$, we take $\xi^1=(j_v^1,u_v^1)_{v\in\mathscr{T}_{d-1}}=\xi$.
Note that the Radon-Nikodym derivative~\eqref{eq:RN} is bounded from above by $K^{2^{d-1}}$.
\end{proposition}
The next result goes further in the study of the regularity of $\rmd Q_{\P,F}/\rmd Q_0$ as a function of $F$ and is required in the proof of Lemma~\ref{lem:RKHS-igb}.

\begin{proposition} \label{prop:RN-diff}
  We assume that $\ell$ satisfies~\eqref{eq:assumptions-on-ell} and~\eqref{eq:regularity-partial-ell}.
  For any splitting scheme $\xi$, the mapping
  \[
    q^{(\xi)}:\begin{cases}
      L^2 &\to \mathbb{R}\\
      F &\mapsto \frac{\rmd Q_{\P,F}}{\rmd Q_0}(\xi)
    \end{cases}
  \]
  is:
  \begin{enumerate}[i)]
    \item bounded away from zero on bounded sets of $L^2$;
    \item Fréchet differentiable;
  \end{enumerate}
  with both these properties being uniform in $\xi$, the uniform differentiability meaning that for any $F\in L^2$,
  \[
    \sup_{\xi} \abs[\Big]{q^{(\xi)}(F+G) - q^{(\xi)}(F) - \rmd_F q^{(\xi)} (G) } = o(\norm{G}_{L^2}).
  \]
\end{proposition}

\begin{proof}[Proof of Proposition~\ref{prop:app-RN-splitting-scheme}]
  The proof relies on proving analogous uniform statements for the scores $\Delta_{\P,F}(j,u; A)$ defined in~\eqref{eq:def_delta_n}.
  First let us consider $M\geq 0$ and show that these scores are bounded, uniformly in $\norm{F}_{L^2} \leq M$ and in $(j,u,A)$.
  Indeed, both terms in~\eqref{eq:def_delta_n} can be rewritten
  \begin{align*}
    \frac{\P[\partial_1\ell(F(x),y)\1_{A_0}(x)]^2}{\P(A_0)} &= \P(A_0) \P[\partial_1\ell(F(x),y) \mid x\in A_0]^2\\
    &\leq \P(A_0) \P[\partial_1\ell(F(x),y)^2 \mid x\in A_0]\\
    & = \P[\partial_1\ell(F(x),y)^2 \1_{A_0}(x)].
  \end{align*}
  Taking the sum, we simply get the bound
  \[
    \Delta_{\P,F}(j,u;A) \leq \P[\partial_1\ell(F(x),y)^2\1_{A}(x)] \leq C(1+M^2),
  \]
  for some constant $C$, where the last inequality is straightforward from the regularity assumptions~\eqref{eq:assumptions-on-ell} on the function $\ell$.
  As a result, plugging this bound in~\eqref{eq:RN} yields the bound
  \[
    q^{(\xi)}(F) \geq \bigg(\frac{1}{\exp(\beta C(1+M^2))}\bigg){2^{d-1}} > 0,
  \]
  and this bound is uniform in $\xi$.
  This proves point~\textit{(i)} of the proposition, and point~\textit{(ii)} will be proved similarly.
  By assumption~\eqref{eq:regularity-partial-ell}, we have for any $F,G\in L^2$,
  \[
    \partial_1\ell(F(x)+G(x),y) = \partial_1\ell(F(x),y) + \partial_1^2\ell(F(x),y)G(x) + o(G(x)),
  \]
  the $o(G(x))$ term being uniform in $y$.
  Therefore, similar computations as above show that
  \begin{align*}
      \frac{\P[\partial_1\ell(F(x)+G(x),y)\1_{A_0}(x)]}{\P(A_0)^{1/2}} & = \frac{\P[\partial_1\ell(F(x),y)\1_{A_0}(x)]}{\P(A_0)^{1/2}} \\
      &\quad + \frac{\P[\partial_1^2\ell(F(x),y)G(x)\1_{A_0}(x)]}{\P(A_0)^{1/2}} + o(\norm{G}_{L^2}),
  \end{align*}
  where the $o(\norm{G}_{L^2})$ term is uniform in $A_0$.
  Taking the square and summing over $A_0$ and $A_1$, we get differentiability of $F\mapsto \Delta_{\P,F}(j,u;A)$, uniformly in $(j,u,A)$.
  
  The softmax function $(x_1,\dots,x_K)\mapsto e^{x_1} / (\sum_i e^{x_i})$ being very regular (it is not hard to see that it is infinitely differentiable with bounded derivatives), plugging the differentiability of the scores into~\eqref{eq:RN}, we readily get the differentiability of $q^{(\xi)}$ at any point $F$, uniformly in $\xi$.
\end{proof}

\section{Auxiliary Lemmas in Hilbert space theory}

We gather here some results on Hilbert space linear algebra that are used in the proof of Lemma~\ref{lem:RKHS-igb}.

\begin{lemma} \label{lem:linear-algebra}
Let $K$ be a closed subspace of a real Hilbert space $(H,\langle\cdot,\cdot\rangle)$, and let $A$ be an invertible, positive definite, bounded, self-adjoint operator.
Denote by $P$ (resp.\ $P_A$) the orthogonal projection onto $K$ (resp.\ the orthogonal projection with respect to the inner product $\langle \cdot, \cdot\rangle_{A} := \langle A\cdot, \cdot\rangle$).
Then:
\begin{enumerate}[i)]
  \item The operator $PA_{|K}\colon K \to K$ is invertible, with
  \[
    (PA_{|K})^{-1} = A^{-1/2}\,\tilde{P}_A\,A^{-1/2},
  \]
  where $\tilde{P}_A$ denotes the orthogonal projection onto $A^{1/2}K$.
  \item We can write
  \[
    P_A = (PA_{|K})^{-1}PA.
  \]
\end{enumerate}
\end{lemma}

\begin{proof}
To show \textit{(i)}, first note that $A^{-1/2}\tilde{P}_A A^{-1/2}$ has image in $K$, since the range of $\tilde{P}_A$ is $A^{1/2}K$.
For elements $x,y \in K$, the following statements are equivalent:
\begin{align*}
  PAx = y
  &\iff \forall z\in K,\; \langle Ax, z\rangle = \langle y, z\rangle, \\
  &\iff \forall z\in K,\; \langle A^{1/2}x, A^{1/2}z\rangle = \langle A^{-1/2}y, A^{1/2}z\rangle, \\
  &\iff \forall z\in A^{1/2}K,\; \langle A^{1/2}x, z\rangle = \langle A^{-1/2}y, z\rangle, \\
  &\iff A^{1/2}x = \tilde{P}_A A^{-1/2}y,
\end{align*}
where each equivalence is straightforward.
This shows that $PA_{|K}\colon K\to K$ is a bijection, with inverse given by $A^{-1/2}\tilde{P}_A A^{-1/2}$.

We now prove \textit{(ii)} by showing that $PAP_AA^{-1} = P$.
This implies $PAP_A = PA$, and since both $P$ and $P_A$ have image $K$ by definition, composing with the inverse of $PA_{|K}$ yields the desired expression for $P_A$.
Indeed, for any $x\in H$ and $y\in K$, we have
\[
  \langle PAP_AA^{-1}x, y\rangle
  = \langle AP_AA^{-1}x, y\rangle
  = \langle P_AA^{-1}x, y\rangle_A
  = \langle A^{-1}x, y\rangle_A
  = \langle x, y\rangle,
\]
which characterizes $PAP_AA^{-1} = P$, concluding the proof.
\end{proof}

\begin{corollary} \label{cor:linear-algebra}
With the notation of Lemma~\ref{lem:linear-algebra}, let
\[
  \mathcal{D} = \{A \in \mathcal{L}(H) : PA_{|K} \text{ has an inverse in } \mathcal{L}(K)\}.
\]
Then the mapping
\[
  A \in \mathcal{D} \longmapsto (PA_{|K})^{-1}PA \in \mathcal{L}(H)
\]
is continuously Fréchet differentiable, and it coincides with $A \mapsto P_A$ when $A$ is an invertible, self-adjoint, positive definite operator.
\end{corollary}

\begin{proof}
The mapping $A \in \mathcal{L}(H) \mapsto PA_{|K} \in \mathcal{L}(K)$ is linear and bounded, and it is standard that the inverse map on $\mathcal{L}(K)$ is continuously Fréchet differentiable on its domain.
The result follows, since the full mapping can be written as a composition of continuously differentiable maps.
\end{proof}

\section{Details on the simulation studies}\label{appendix:simu}

\subsection{Details on the KPCA simulation study}
The exact datasets and their main features are described in Table \ref{tab:classification_datasets} and Table \ref{tab:regression_datasets} for binary and continuous outcomes, respectively. In the "Features" columns, a tilde is used to indicate the number of features present prior to one-hot encoding.

\begin{table}[h]
\centering
\begin{tabular}{@{}llrl@{}}
\toprule
\textbf{Dataset} & \textbf{Task Description} & \textbf{Samples} & \textbf{Features} \\
\midrule
Breast Cancer & Cancer diagnosis (malignant vs. benign) & 569 & 30 \\
Mushroom & Edibility classification & 8,124 & $\sim$20 \\
German Credit & Credit approval prediction & 1,000 & $\sim$60 \\
Heart Disease & Cardiovascular diagnosis & 270 & $\sim$13 \\
Ionosphere & Radar signal classification & 351 & 34 \\
Sonar & Signal classification (mine vs. rock) & 208 & 60 \\
Diabetes & Disease diagnosis & 768 & $\sim$8 \\
Spam & Email spam detection & 4,601 & 57 \\
Titanic & Survival prediction & 1,309 & $\sim$10 \\
Wine & Cultivar classification & 178 & 13 \\
Iris & Species classification & 150 & 4 \\
\bottomrule
\end{tabular}
\caption{Classification datasets used in the empirical evaluation. All multi-class datasets were converted to binary classification problems.}
\label{tab:classification_datasets}
\end{table}

\begin{table}[h]
\centering
\begin{tabular}{@{}llrl@{}}
\toprule
\textbf{Dataset} & \textbf{Task Description} & \textbf{Samples} & \textbf{Features} \\
\midrule
Diabetes & Disease progression prediction & 442 & 10 \\
California Housing & Housing price prediction & 20,640 & 8 \\
Wine Quality (Red) & Red wine quality scoring & 1,599 & 11 \\
Wine Quality (White) & White wine quality scoring & 4,898 & 11 \\
Auto MPG & Fuel efficiency prediction & 392 & $\sim$7 \\
Concrete & Compressive strength prediction & 1,030 & 8 \\
Abalone & Age prediction & 4,177 & $\sim$8 \\
Power Plant & Energy output prediction & 9,568 & 4 \\
\bottomrule
\end{tabular}
\caption{Regression datasets used in the empirical evaluation.}
\label{tab:regression_datasets}
\end{table}

\begin{figure}[htbp]
  \centering
    \resizebox{0.7\linewidth}{!}{
    \input{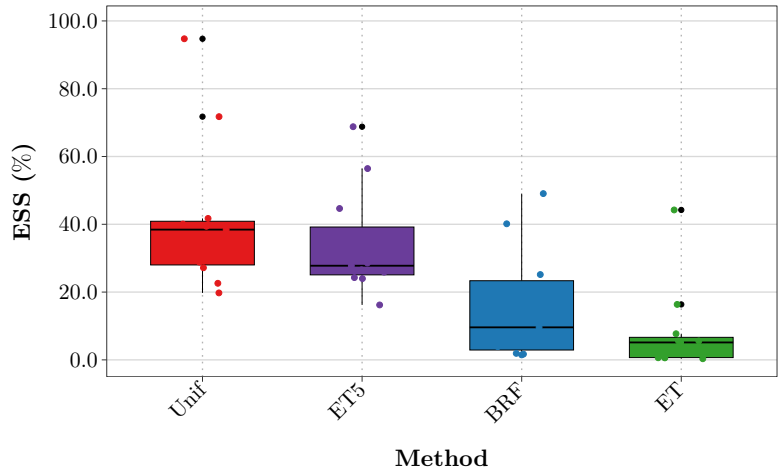}
    }
  \caption{Summary of the effective sample sizes for each Random Forest algorithm over the classification datasets.}
\end{figure}

\begin{figure}[htbp]
  \centering
    \resizebox{0.7\linewidth}{!}{
    \input{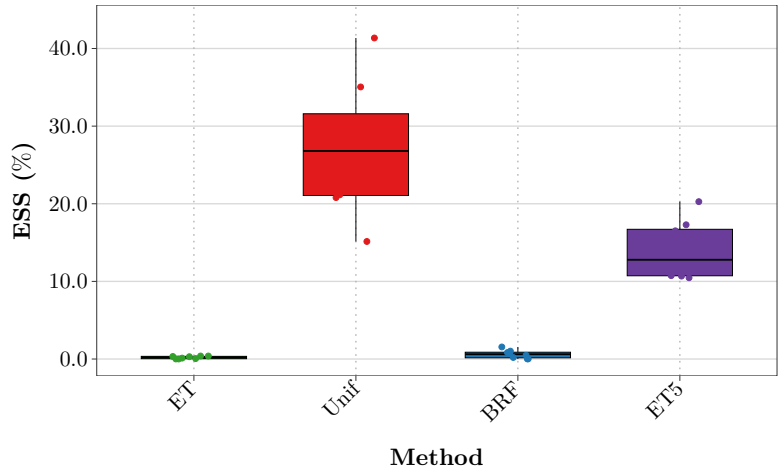}
    }
  \caption{Summary of the effective sample sizes for each Random Forest algorithm over the regression datasets.}
\end{figure}

\subsection{Details on the variable importance simulation study}

\subsection*{Appendix: Detailed Description of Simulation Scenarios}

We now provide a complete specification of each data-generating scenario. In all cases, features not explicitly defined are drawn independently from the distribution stated for that scenario. The signal set $S$ identifies which features truly affect the outcome.

\paragraph{S1: Additive.}
We draw $X \sim \mathcal{N}(0, \Sigma)$ where $\Sigma_{jk} = 0.5^{|j-k|}$. The outcome is
\[
Y = 1.0 \cdot X_1 + 0.8 \cdot 2\sin(\tfrac{\pi}{2} X_2) + 0.6 \cdot \max(0, X_3) + \varepsilon, \quad \varepsilon \sim \mathcal{N}(0, 0.5^2),
\]
with signal set $S = \{1, 2, 3\}$. This tests a simple additive model with mild correlation.

\paragraph{S2: Collinear signals.}
We first draw a latent variable $Z \sim \mathcal{N}(0,1)$ and noise terms $\eta_1, \eta_2 \sim \mathcal{N}(0, 0.1^2)$. The first two features are noisy proxies of $Z$:
\[
X_1 = Z + \eta_1, \quad X_2 = Z + \eta_2,
\]
while the remaining features are independent standard normals. The outcome depends only on $Z$:
\[
Y = \sin(Z) + \varepsilon, \quad \varepsilon \sim \mathcal{N}(0, 0.3^2).
\]
Signal set: $S = \{1, 2\}$. Here both $X_1$ and $X_2$ carry information about $Y$ but are highly collinear.

\paragraph{S3: XOR.}
Features are drawn as $X_j \sim \mathcal{U}([0,1])$ independently. The outcome is a binary XOR of thresholded features:
\[
Y = \mathds{1}_{\{(X_1 > 0.5) \oplus (X_2 > 0.5)\}},
\]
where $\oplus$ denotes the XOR operation. We then add 5\% label noise by randomly flipping the value of $Y$. Signal set: $S = \{1,2\}$. This scenario has a pure interaction effect with no marginal main effects.

\paragraph{S4: Categorical distractor.}
Features $X_1, X_2$ are drawn from $\mathcal{N}(0,1)$, and we include a categorical noise variable $X_4 \in \{0, 1, \dots, 49\}$ drawn uniformly. The outcome is
\[
Y = 1.0 \cdot X_1 + 0.8 \cdot 2\sin(\tfrac{\pi}{2} X_2) + \varepsilon, \quad \varepsilon \sim \mathcal{N}(0, 0.5^2).
\]
Signal set: $S = \{1,2\}$. The high-cardinality categorical variable acts as a distractor with many potential splits but no predictive value.

\paragraph{S5: Local relevance.}
All features are $X_j \sim \mathcal{U}([0,1])$. The outcome depends only on whether $X_1$ exceeds a threshold:
\[
Y = \mathds{1}_{\{X_1 > 0.8\}} + \varepsilon, \quad \varepsilon \sim \mathcal{N}(0, 0.1^2).
\]
Signal set: $S = \{1\}$. This tests whether methods can detect a feature that matters only in a localized region.

\paragraph{S6: Redundant noises.}
Features are drawn from $\mathcal{N}(0,1)$. For $j \in \{11, 12, \dots, 20\}$, we create noisy copies of the first ten features:
\[
X_j = X_{j-10} + \xi_j, \quad \xi_j \sim \mathcal{N}(0, 0.05^2).
\]
The outcome is
\[
Y = 1.0 \cdot X_1 + 0.8 \cdot 2\sin(\tfrac{\pi}{2} X_2) + \varepsilon, \quad \varepsilon \sim \mathcal{N}(0, 0.5^2).
\]
Signal set: $S = \{1,2\}$. The challenge is to distinguish the original signals from their redundant noisy copies.

\paragraph{S7: High-dimensional.}
We use $p=50$ features drawn from $\mathcal{U}([0,1])$. For $i \in \{1,2,3,4,5\}$, we introduce correlation:
\[
X_{i+10} = 0.9 \cdot X_i + 0.436 \cdot X_{i+10}.
\]
The outcome is linear in the first five features:
\[
Y = 3.0 \cdot X_1 + 2.5 \cdot X_2 + 2.0 \cdot X_3 + 1.5 \cdot X_4 + 1.0 \cdot X_5 + \varepsilon, \quad \varepsilon \sim \mathcal{N}(0, 1.5^2).
\]
Signal set: $S = \{1,2,3,4,5\}$. This tests performance in higher dimensions with sparse signals and correlated noise.

\paragraph{S8: Weak signal.}
Features are drawn from $\mathcal{N}(0,1)$. The outcome has a linear structure with weak coefficients and large noise:
\[
Y = 0.5 \cdot X_1 + 0.4 \cdot X_2 + 0.3 \cdot X_3 + \varepsilon, \quad \varepsilon \sim \mathcal{N}(0, 2.0^2).
\]
Signal set: $S = \{1,2,3\}$. The low signal-to-noise ratio makes detection challenging.

\paragraph{S9: Correlated noise.}
Features are drawn from $\mathcal{N}(0,1)$. We then create "confounding correlations" by correlating noise features with the true signals:
\[
X_{i+5} = 0.95 \cdot X_i + 0.312 \cdot X_{i+5}, \quad X_{i+10} = 0.95 \cdot X_i + 0.312 \cdot X_{i+10}, \quad i \in \{1,2,3\}.
\]
The outcome is
\[
Y = 3.0 \cdot X_1 + 2.5 \cdot X_2 + 2.0 \cdot X_3 + \varepsilon, \quad \varepsilon \sim \mathcal{N}(0, 0.5^2).
\]
Signal set: $S = \{1,2,3\}$. The correlated noises are strongly associated with the signals but have no causal effect on $Y$.

\paragraph{S10: Context-dependent.}
Features are drawn from $\mathcal{U}([0,1])$. The effect of $X_2$ and $X_3$ on the outcome depends on the value of $X_1$:
\[
Y = \begin{cases}
2.0 \cdot X_2 + \varepsilon & \text{if } X_1 > 0.5, \\
2.0 \cdot X_3 + \varepsilon & \text{if } X_1 \leq 0.5,
\end{cases}
\quad \varepsilon \sim \mathcal{N}(0, 0.3^2).
\]
Signal set: $S = \{1,2,3\}$. All three features are important, but $X_2$ and $X_3$ have context-dependent effects.

\phantomsection
\addcontentsline{toc}{section}{References}

\vskip 0.2in
\bibliography{refs.bib}

\end{document}